\PassOptionsToPackage{table,xcdraw}{xcolor}
\documentclass[10pt,twocolumn,letterpaper]{article}

\usepackage{cvpr}              

\newcommand{\ours}[0]{{HATIE}}
\newcommand{\affaddr}[1]{#1}

\usepackage{bm}

\definecolor{cvprblue}{rgb}{0.21,0.49,0.74}
\usepackage[accsupp]{axessibility}  
\usepackage[pagebackref,breaklinks,colorlinks,allcolors=cvprblue]{hyperref}
\usepackage{booktabs}
\usepackage{multirow}
\usepackage[table,xcdraw]{xcolor}
\usepackage{caption}

\def\paperID{9125} 
\def\confName{CVPR}
\def\confYear{2025}

\title{Towards Scalable Human-aligned Benchmark for Text-guided Image Editing}

\author{Suho Ryu\thanks{Equal Contribution}, Kihyun Kim\footnotemark[1], Eugene Baek\footnotemark[1], Dongsoo Shin, Joonseok Lee\thanks{Corresponding author}\\
\affaddr{Graduate School of Data Science, Seoul National University}\\
{\tt\small \{jmhera2007,ki5477,eugene0103,dongsoo,joonseok\}@snu.ac.kr}\\
{\tt \url{https://github.com/SuhoRyu/HATIE}}
}

\begin{document}
\maketitle
\begin{abstract}
A variety of text-guided image editing models have been proposed recently. However, there is no widely-accepted standard evaluation method mainly due to the subjective nature of the task, letting researchers rely on manual user study.
To address this, we introduce a novel Human-Aligned benchmark for Text-guided Image Editing (\ours). Providing a large-scale benchmark set covering a wide range of editing tasks, it allows reliable evaluation, not limited to specific easy-to-evaluate cases. Also, \ours\ provides a fully-automated and omnidirectional evaluation pipeline. Particularly, we combine multiple scores measuring various aspects of editing so as to align with human perception.
We empirically verify that the evaluation of \ours\ is indeed human-aligned in various aspects, and provide benchmark results on several state-of-the-art models to provide deeper insights on their performance.
\end{abstract}    
\section{Introduction}
\label{sec:intro}

With recent advances in diffusion-based image generation models~\cite{rombach2022high,saharia2022photorealistic,ramesh2021zero}, text-guided image editing has been prominently developed.
For example, Imagic~\cite{kawar2023imagic}, Prompt-to-Prompt~\cite{hertz2023p2p}, and SDEdit~\cite{meng2022sdedit} have pioneered this direction in early years, and recent models \citep{zhang2024magicbrush, geng2024instructdiffusion, huang2024smartedit} have remarkably improved the performance.

However, there is no widely-accepted standard evaluation methodology to accurately and objectively assess these models.
This partly comes from the nature of the editing task that there can be multiple correct outputs for each instance.
For example, ``adding a table in a living room scene'' can be satisfied with many different shapes of the table.
Even for a more specific query like ``a circular wooden tea table with three legs'', the pixel-level details can be infinitely diverse.
This becomes even worse when the query is more ambiguous.

This inherent nature of text-driven image editing causes difficulties in evaluation.
Due to the lack of a `golden' ground truth, it is crucial to have a proper metric to measure the deviation between the edited image and the ground truth.
Unlike traditional machine learning problems where we can easily measure the deviation using a simple distance metric like $L_2$, it is well-known that image distance cannot be simply measured by pixel-level distances.
Thus, it is essential to have \textit{human-like high-level understanding} to measure the relevance with human perception to properly evaluate the quality of edited images.
For this, model-based metrics such as CLIP \citep{radford2021clip} similarity or LPIPS \citep{zhang2018lpips} distance has been used. 
Also, large language models (LLM) or vision language models (VLM) have been widely adopted recently.
Relying on a particular model, however, sacrifices reliability and reproducibility of evaluation results.

\begin{figure}[t]
  \centering
  \begin{subfigure}{0.3\linewidth}
    \centering
    \includegraphics[width=\linewidth]{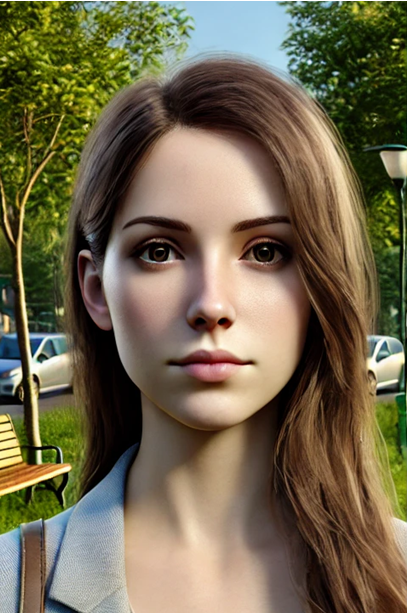}
    \caption{Original image}
    \label{fig:intro_ex_a}
  \end{subfigure}
  \hfill
  \begin{subfigure}{0.6\linewidth}
    \centering
    \includegraphics[width=\linewidth]{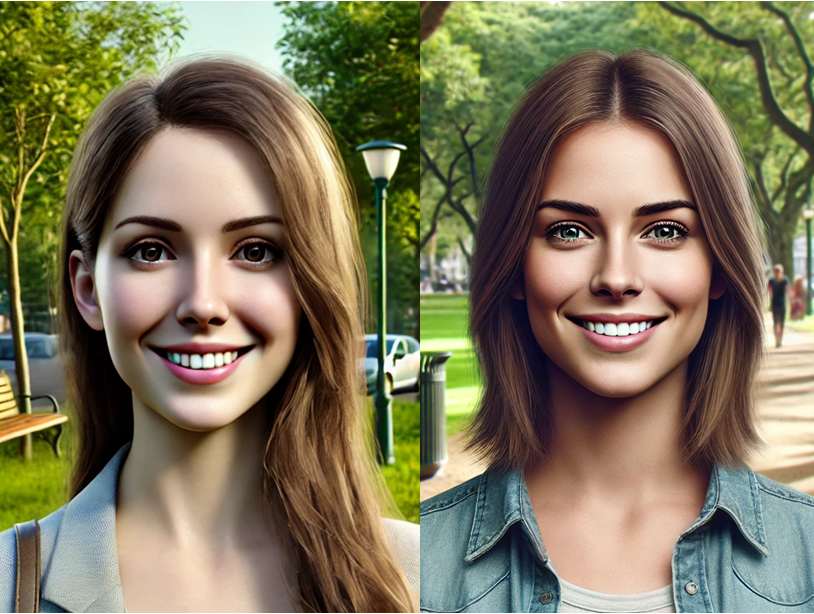}
    \caption{Edited images to ``make her smile''}
    \label{fig:intro_ex_b}
  \end{subfigure}
  \vspace{-0.2cm}
  \caption{
  \textbf{An example highlighting the importance of consistency in image editing.} (a) Original image (b) Edited images for a prompt ``Make her smile''. The left result is more consistent with the input than the right one, better preserving her identity. 
  }
  \label{fig:intro_ex}
\end{figure}

In addition, there are \textit{multiple aspects} to consider when we evaluate the edited images.
First of all, the edited image itself should be realistic and feasible.
The overall configuration and specific details of each object in the image should be consistent with our common knowledge.
Second, the output image should correctly reflect the instructed change.
At the same time, it is expected not to alter unintended aspects in the instruction.
For instance, if an image shows a person and the task is to make her smile, the identity of the person should not change after editing (\cref{fig:intro_ex}).
Due to these multi-faceted aspects of image editing, a crude evaluation would be insufficient to assess detailed quality of the edit.
As it is unclear how to automate this overall evaluation, most previous research have relied on user study, which is fundamentally limited to scale and reproduce.

There have been a few notable attempts to automate evaluation of image editing, listed in \cref{tab:bench_comp}. 
\citet{kawar2023imagic} and \citet{wang2023editbench} presented a benchmark guideline and an image set.
Due to their small \textit{scale of the benchmark set}, however, this work is insufficient to perform robust evaluation, and thus they also have relied on user study.
\citet{shi2020ldie} provided a set of editing requests with ground truth result image for automated evaluation.
However, their editing tasks are quite limited to those where a clear ground truth can be obtained at almost pixel-level, \emph{e.g.}, 
gray scaling an image.
Aiming at a similar goal to ours, \citet{basu2023editval} diversified editing tasks and target objects to cover a wider range of editing.
The evaluation, however, has been done by user study.

\begin{table}
    \centering \renewcommand{\tabcolsep}{2pt}
    \resizebox{\linewidth}{!}{
    \footnotesize
    \begin{tabular}{l|rrccc}
    \toprule
        Method & \parbox{1.3cm}{\centering \# Images} & \parbox{1.3cm}{\centering \# Queries} & \parbox{1.6cm}{\centering \# Obj. class} & \parbox{1.3cm}{\centering Auto-eval} \\
        \toprule
        TEdBench \citep{kawar2023imagic} & 100 & 100 & N/A & No \\
        EditBench \citep{wang2023editbench} & 240 & 720 & N/A & N/A\\
        GIER \citep{shi2020ldie} & 6,179 & 30,895 & N/A & N/A \\
        EditVal \citep{basu2023editval} & 92 & 648 & 19 & Partial \\
        \midrule
        \ours\ (Ours) & \textbf{18,226} & \textbf{49,840} & \textbf{76} & \textbf{Full} \\
        \bottomrule
    \end{tabular}
    }
    \vspace{-2mm}
    \caption{\textbf{Comparison with other image editing benchmarks.} \ours\ provide a fully-automated large-scale evaluation method covering wide range of objects and tasks.}
    \label{tab:bench_comp}
    \vspace{-0.5cm}
\end{table}

In this paper, we propose a scalable \textbf{H}uman-\textbf{A}ligned \textbf{T}ext-guided \textbf{I}mage \textbf{E}diting benchmark (\textbf{\ours}) that addresses the aforementioned limitations.
First, our large-scale benchmark set \textit{covers a wide range of editing cases}, not limited to specific easy-to-evaluate cases.
Second, \ours\ measures the quality of edited images in a fully-automated and \textit{multifaceted} way, \textit{aligned with human perception}.
Our benchmark allows robust, objective, and labor-free evaluation of existing and new image editing models, paving the road for the advancement of this field.

\section{Problem Formulation}
\label{sec:pre}

Given an image $I_0$ and a natural language caption $C$, the text-guided image editing task aims to generate an edited image $I_e$, which faithfully reflects the instruction described in $C$ while minimizing unintended deviations from $I_0$.
A text-guided image editing mode $f$
may take an additional parameter $p$ to adjust the degree of manipulation, focusing more (or less) on reflecting the instruction $C$ compared to preserving the context.
Formally, $I_e = f(I_0, C; p)$.

Depending on the type of the caption $C$, text-guided image editing models can be divided into two categories: description-based and instruction-based.
Description-based models are conditioned by two separate captions, $C = \{C_0, C_e\}$, where $C_0$ and $C_e$ describe the original and desired images, respectively.
For example, given $C_0 =$ ``a photo of a dog'' and $C_e =$ ``a photo of a cat'', the editing task is to replace the dog in $I_0$ with a cat.
Many early approaches \citep{hertz2023p2p, kawar2023imagic, cao2023masactrl, lin2024text, couairon2022diffedit, kwon2022diffusion, wu2024freediff, nichol2021glide, avrahami2022blended} have adopted this scheme, 
as it is straightforward to directly apply an image generation model conditioned on a text caption.
However, this approach has drawbacks.
Certain types of editing queries cannot be easily articulated with two separate captions.
For example, to remove a dog from a photo, $C_0$ can be ``A photo of a dog'' but $C_e$ would be ``A photo without a dog'', which is not a clear description of the desired image.

On the other hand, instruction-based models \citep{brooks2023instructpix2pix, zhang2024magicbrush, geng2024instructdiffusion, wang2023instructedit} take a single instruction $C$, describing the desired delta between the original and the edited image.
Under this setting, with an instruction $C =$ ``remove the dog'', a model is expected to generate an image $I_e$ without the dog from $I_0$.

Our framework \ours\ is applicable to both caption types.
Note that models taking different types of captions are not comparable.
It might be possible to convert one type of captions to the other, but evaluating on this indirect setting would be unfair for one side.

\section{The \ours\ Framework}
\label{sec:method}

\begin{figure*}
    \centering
    \includegraphics[width=\linewidth]{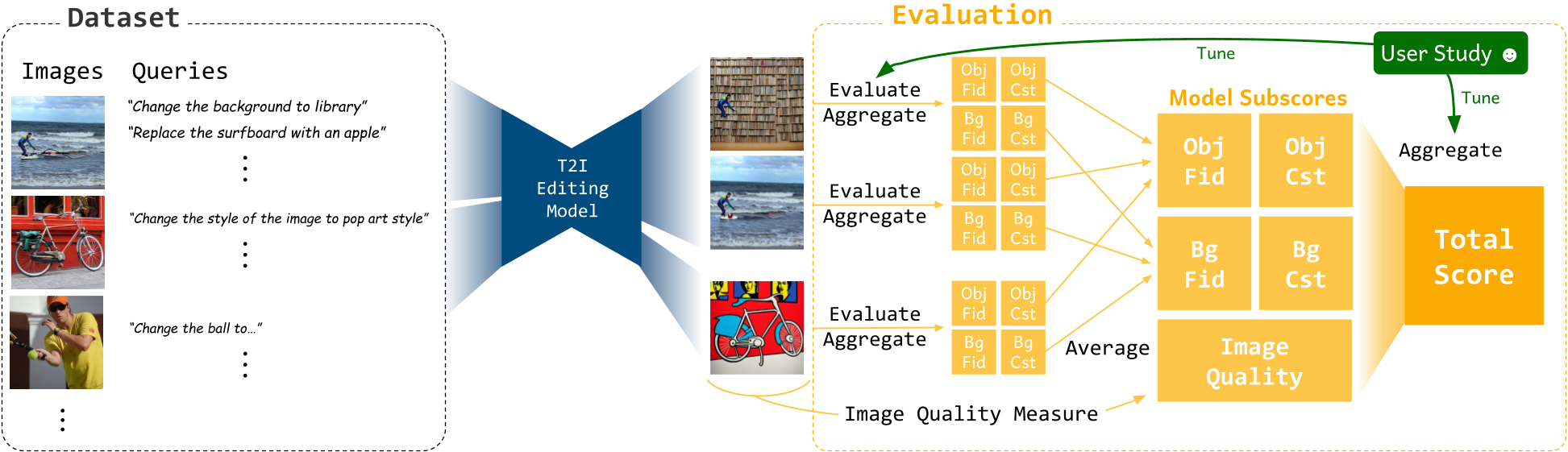}
    \caption{\textbf{Overview of our \ours\ Benchmark.} \ours\ consists of an image and query dataset for editing, along with an automated evaluation pipeline for assessing editing performance. We curate a large-scale comprehensive dataset with images and corresponding editing queries, on which a model would perform text-guided image editing. Then, \ours\ evaluates the edited images from 5 different aspects: \textit{Object Fidelity, Background Fidelity, Object Consistency, Background Consistency,} and \textit{Image Quality}. Finally, these scores are aggregated by a weight fitted to human feedback through our user study, producing the final \textit{Total Score.}}
    \label{fig:workflow}
\end{figure*}

For a fair comparison of editing models, a same set of queries should be given to every model, and the set should be large enough to cover a wide range of possible domains, so that no model is particularly favored.
Also, the results should be evaluated by consistent, objective, and equitable metrics.
Moreover, users should be able to consistently reproduce the evaluation results at any time.
For this, it is discouraged to rely on external services (\emph{e.g.}, ChatGPT) without public access to the actual parameters for evaluation.

In order to satisfy these conditions, our \ours\ provides a large-scale comprehensive dataset, containing images to be edited and annotations of editable objects in them (\cref{sec:method:dataset}), a set of curated editing queries for these editable objects, in both description and instruction forms (\cref{sec:method:query}), and
a set of evaluation metrics designed to evaluate the edited images from various perspectives (\cref{sec:method:eval}).
\cref{fig:workflow} illustrates a diagram of the overall benchmark workflow.

\subsection{Dataset}
\label{sec:method:dataset}

\textbf{Base Dataset.}
We take the GQA dataset \citep{hudson2019gqa}, one of foundational datasets designed for Visual Question and Answering (VQA) tasks as our benchmark base.
It consists of 113K real-world images with in-depth annotations, including the list of objects labeled with their names, bounding boxes, attributes, and relations with other objects (scene graph).
As a precise visual understanding is crucial to figure out what kind of editing jobs are feasible in each image, rich annotations in GQA make it suitable for generating edit queries.

\noindent
\textbf{Data Filtering and Augmentation.}
Not every object in a scene is suitable for editing.
We filter out indistinguishable objects that are too small, cropped, or occluded.
Also, for automated evaluation, it is necessary for an object to be detected by object detector, which is typically trained on a dataset labeled for a fixed set of classes, \emph{e.g.}, MS-COCO~\citep{lin2014microsoft}.
We first match the freeform GQA object names with 80 object classes in COCO, and filter out objects not included in the COCO classes. Then, we remove the objects that are not detected by an instance segmentation model~\cite{wu2019detectron2} used at evaluation (\cref{sec:method:eval}).
We also exclude images containing two or more instances belonging to the same class to avoid any ambiguity in figuring out which one is the target to be edited.
More details on dataset filtering is in \cref{sec:edit_type}.

Then, the remaining images are trimmed and resized to $512 \times 512$ to be compatible with editing models.
Since both editable objects and background are important for editing, we preserve the original image as much as possible, and make sure that editable objects are not cropped.

\begin{figure}
    \centering
    \includegraphics[width=\linewidth]{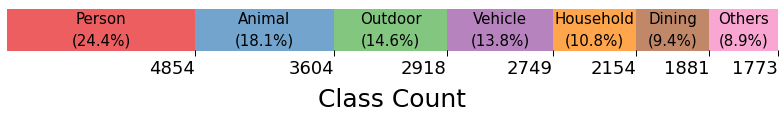}
    \vspace{-0.5cm}
    \caption{\textbf{Object Class Distribution in Our Dataset.} \ours\ evaluates fairly by providing evenly distributed dataset.}
    \label{fig:dataset_dist}
    \vspace{-0.3cm}
\end{figure}

In total, our benchmark set consists of 18,226 images with 19,933 editable objects within 76 COCO classes.
More detailed object class distribution is plotted in \cref{fig:dataset_dist} and \cref{sec:balanced_query}.

\subsection{Editing Query Formulation}
\label{sec:method:query}

\textbf{Query Templates.}
On the collected images and objects, we generate queries to edit them.
To automate this process but still make the queries realistic, we categorize the editing tasks into object-centric and non-object-centric.

An object-centric editing query focuses on a specific object, such as adding an object that is not present in the image (\textsc{Object Addition}), removing an existing object (\textsc{Object Removal}), replacing an object with something else (\textsc{Object Replacement}), and altering some attributes (\textsc{Object Attribute Change}) or size (\textsc{Object Resizing}) of a particular object.
More formally, given a set $\mathcal{C}$ of object classes and an image with editable object $o$ in class $c_o \in \mathcal{C}$ with attributes $\{a_i\} \in \mathcal{A}$, we define templates for these object-centric edits as follows:

\vspace{0.2cm}
\begin{enumerate}[leftmargin=0.7cm]
    \item \textsc{Object Addition}: Create an additional object $o'$ in class $c_{o'} \in \mathcal{C}$ located at $r_{o'} \in \mathcal{R}$ relative to $o$.
    \vspace{0.1cm}
    \item \textsc{Object Removal}: Remove the object $o$ from the image, generating a natural scene just without $o$.
    \vspace{0.1cm}
    \item \textsc{Object Replacement}: Replace $o$ with another object $o'$ in class $c_{o'} \in \mathcal{C}$.
    \vspace{0.1cm}
    \item \textsc{Object Attribute Change}: Change the attribute $a_i$ of $o$ to $a_j \in \mathcal{A}$.
    \vspace{0.1cm}
    \item \textsc{Object Resizing}: Make $o$ larger (or smaller).
\end{enumerate}
\vspace{0.2cm}

\noindent To construct the options, \emph{i.e.}, $\mathcal{R}$ and $\mathcal{A}$, we employ a data-driven approach.
We gather every attribute of each object class $c \in \mathcal{C}$ and every relation between each class pair $(c_1, c_2) \in \mathcal{C} \times \mathcal{C}$ from the annotations in the entire dataset.
We classify and bind the collected attributes into 4 categories (color, state, material, and action), constructing a set of corresponding attributes: $\mathcal{A}_\text{color}$, $\mathcal{A}_\text{state}$, $\mathcal{A}_\text{mat}$, and $\mathcal{A}_\text{act}$.
Also, we consolidate the freeform relation captions into 9 relations: $\mathcal{R}$ = \{under, above, in, on, left, right, next to, front, behind\}.
These attributes and relations cover most cases observed in the dataset.

On the other hand, a non-object-centric query involves background or style changes affecting the entire image: 

\vspace{0.2cm}
\begin{enumerate}[leftmargin=0.7cm]
    \item \textsc{Background Change}: Change the background of the image to $b \in \mathcal{B}$.
    \vspace{0.1cm}
    \item \textsc{Style Change}: Change the image style to $s \in \mathcal{S}$.
\end{enumerate}
\vspace{0.2cm}

\noindent $\mathcal{B}$ and $\mathcal{S}$ are the set of options for background and style changes.
Using a similar approach like above, we collect 27 distinct
backgrounds in $\mathcal{B}$ and 10 styles in $\mathcal{S}$.

\vspace{0.1cm} \noindent
\textbf{Feasibility of Queries.}
\label{sec:query_gen}
A query should be carefully designed, since feasibility of a specific edit depends on many factors in image such as position and size of the objects, relations between the objects or even view point and angle of the image.
For instance, adding a surfboard \textit{inside} a person would not make sense.
Adding the surfboard \textit{under} the person sounds reasonable, but it is impossible if it is out of the view.
While manually defining queries for each image would yield the most accurate and diverse set, it is highly labor-intensive and may introduce subjective bias.

In order to restrict the queries within feasible options, we conduct a statistical analysis on the annotated relations between objects.
We identify other relevant object types and relative locations that are frequently observed for each class, and restrict the options for the target object $o'$ to this commonly observed ones.
In \textsc{Object Addition}, for example, we may add a \textit{laptop} on the table (a frequently observed relation), but not a \textit{car} (a rarely observed relation).
As another example with \textsc{Object Attribute Change}, the original attribute (\textit{e.g.}, $a_i =$ `brown' $\in$ $\mathcal{A}_\text{color}$) is replaced with another one (\textit{e.g.}, $a_j$ = `yellow') from the same category ($\mathcal{A}_\text{color}$).
We detail all task-specific considerations to make the queries feasible in \cref{sec:qgen_rule_detail}.

For non-object-centric tasks, on the other hand, we have relatively less restrictions.
Changing the background to a random place may look odd but it is hardly impossible to imagine.
Thus, we randomly select a background or style from pre-defined candidate sets of 27 backgrounds in $\mathcal{B}$ and 10 styles in $\mathcal{S}$.
The full list of candidate backgrounds and styles are in \cref{sec:qgen_rule_detail}.

\begin{figure}
    \centering
    \begin{subfigure}{0.49\linewidth}
        \centering
        \includegraphics[width=\linewidth]{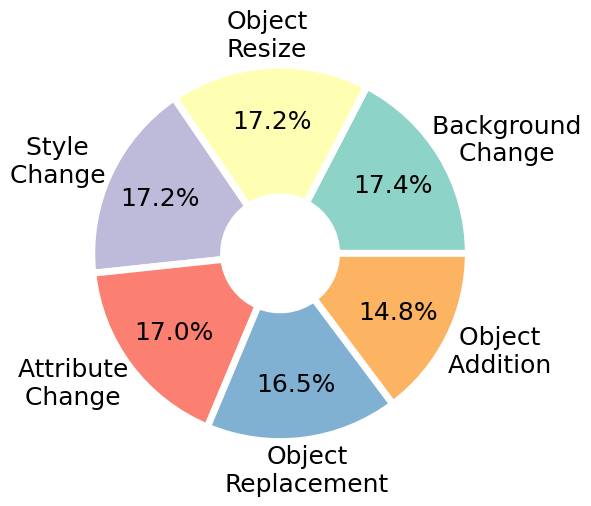}
        \caption{Edit Type Distribution}
    \end{subfigure}
    \hfill
    \begin{subfigure}{0.49\linewidth}
        \centering
        \includegraphics[width=\linewidth]{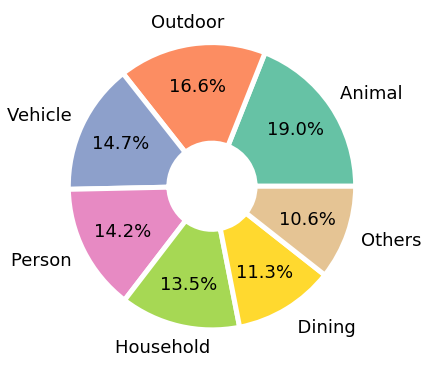}
        \caption{Target Class Distribution}
    \end{subfigure}
    \caption{\textbf{Query Set Distribution.} (a) Distribution of edit types in our query set, (b) Distribution of the object classes designated as the target in object-centric queries.}
    \label{fig:task_dist}
    \vspace{-0.3cm}
\end{figure}

Considering these feasibility restrictions, we have generated 49,840 editing queries on the 18,226 images in total, balanced across various query types and options (\cref{fig:task_dist}, 
\cref{sec:balanced_query}).

\vspace{0.1cm} \noindent
\textbf{Description and Instruction Generation.}
Finally, we formulate the natural language queries into an appropriate form, either description-based or instruction-based, using LLMs \cite{dubey2024llama,wu2024gpt}.
See \cref{sec:capgen_prompt} for the detailed generation flow and prompts.

\begin{figure*}
\centering
\vspace{-0.1cm}
\begin{subfigure}{0.44\linewidth}
  \centering
  \includegraphics[width=\textwidth]{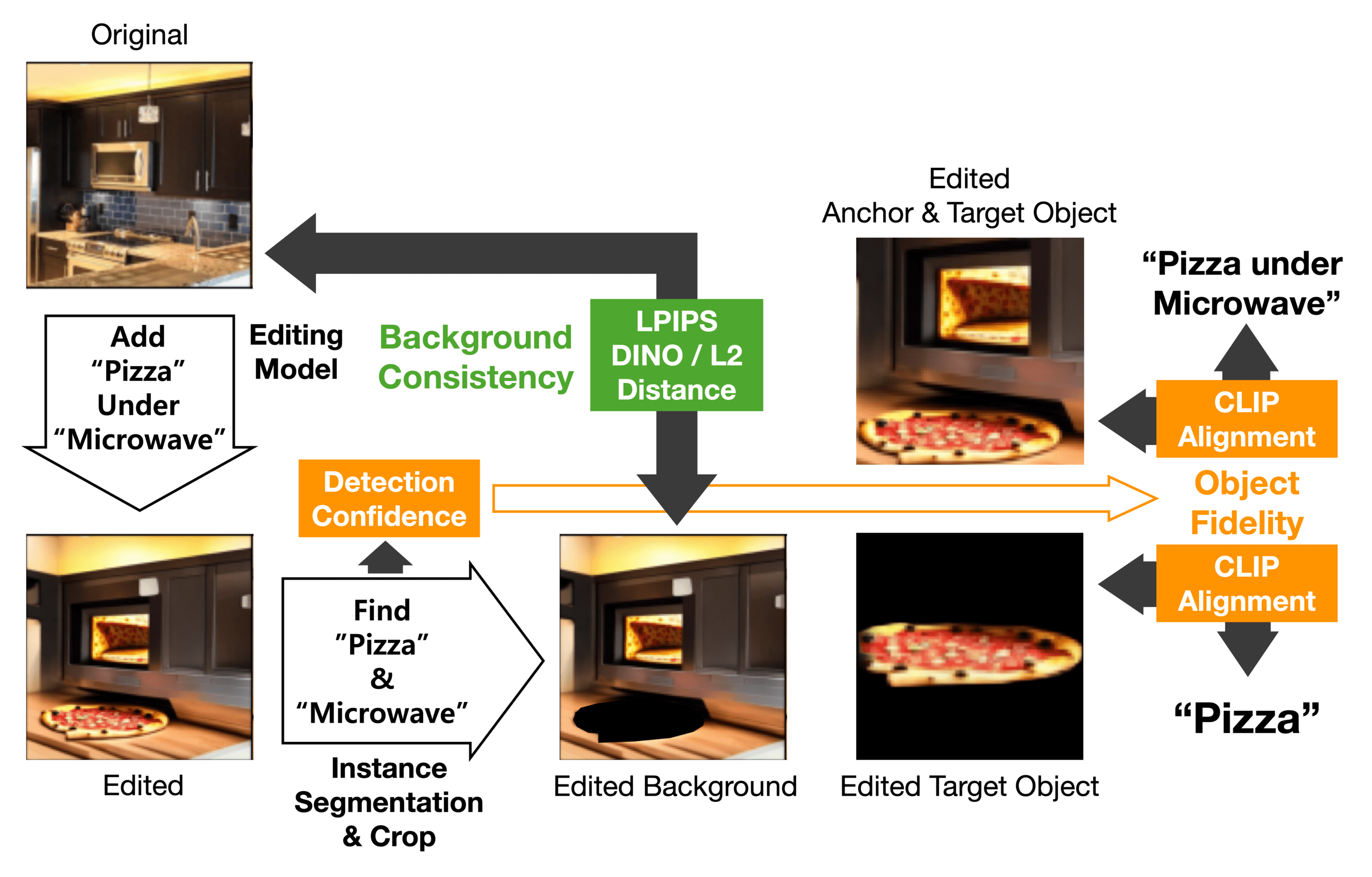}
  \caption{\textsc{Object Addition}}
  \label{fig:eval_add}
\end{subfigure}
\hfill
\begin{subfigure}{0.51\linewidth}
  \centering
  \includegraphics[width=\textwidth]{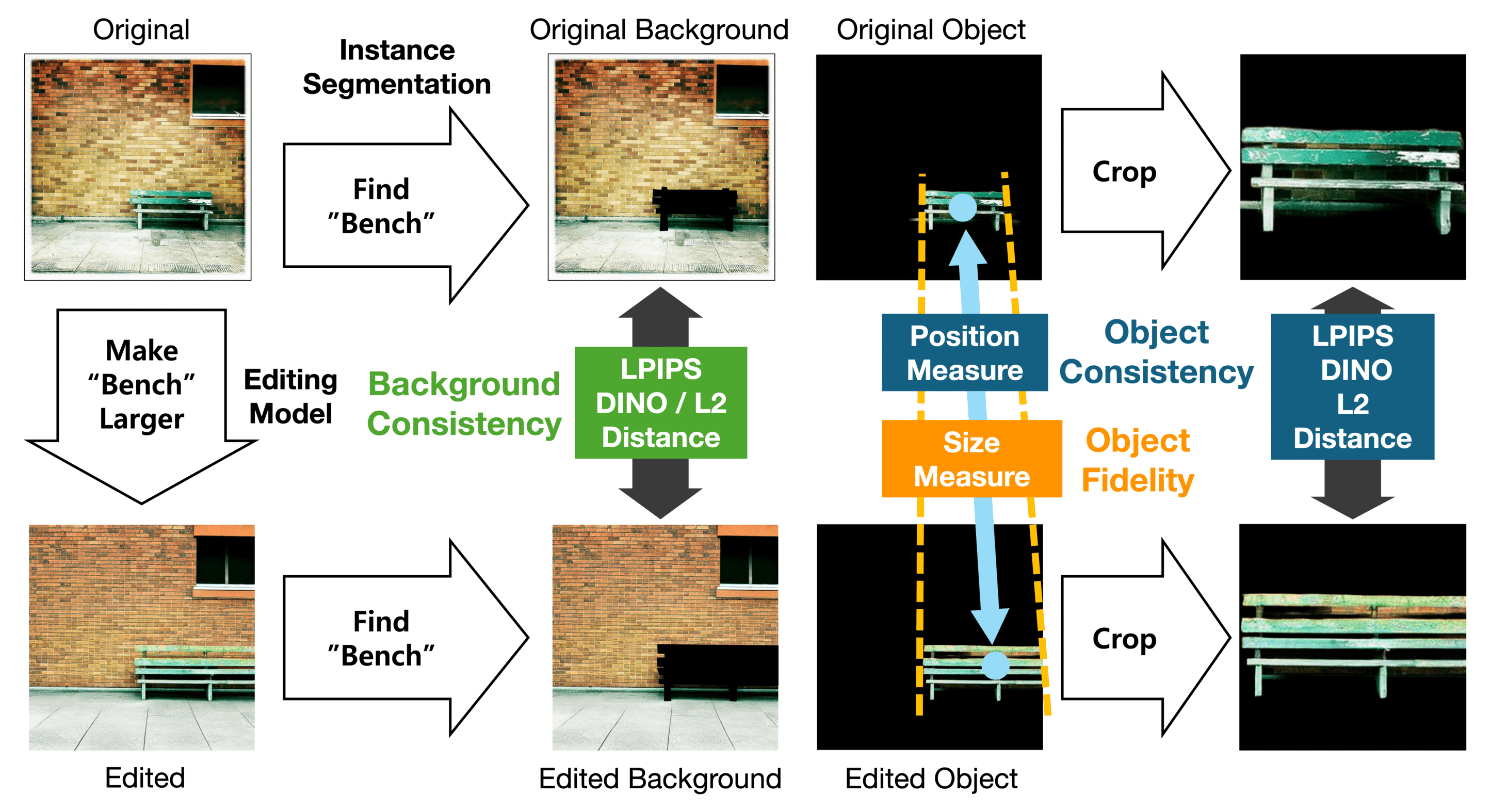}
  \caption{\textsc{Object Resizing}}
  \label{fig:eval_res}
\end{subfigure}
\begin{subfigure}{0.47\linewidth}
  \centering
  \includegraphics[width=\textwidth]{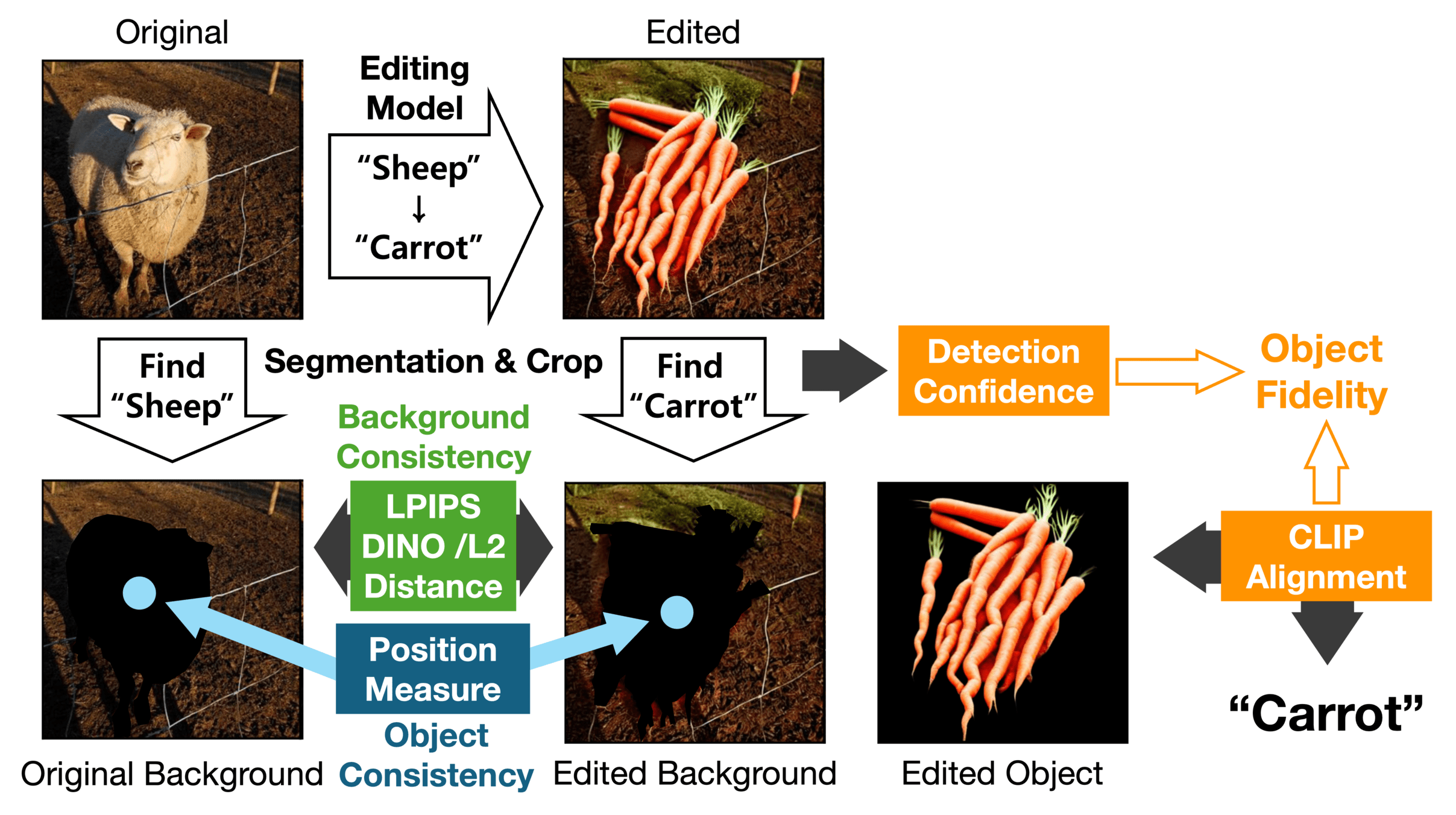}
  \caption{\textsc{Object Replacement}}
  \label{fig:eval_rep}
\end{subfigure}
\hfill
\begin{subfigure}{0.48\linewidth}
  \centering
  \includegraphics[width=\textwidth]{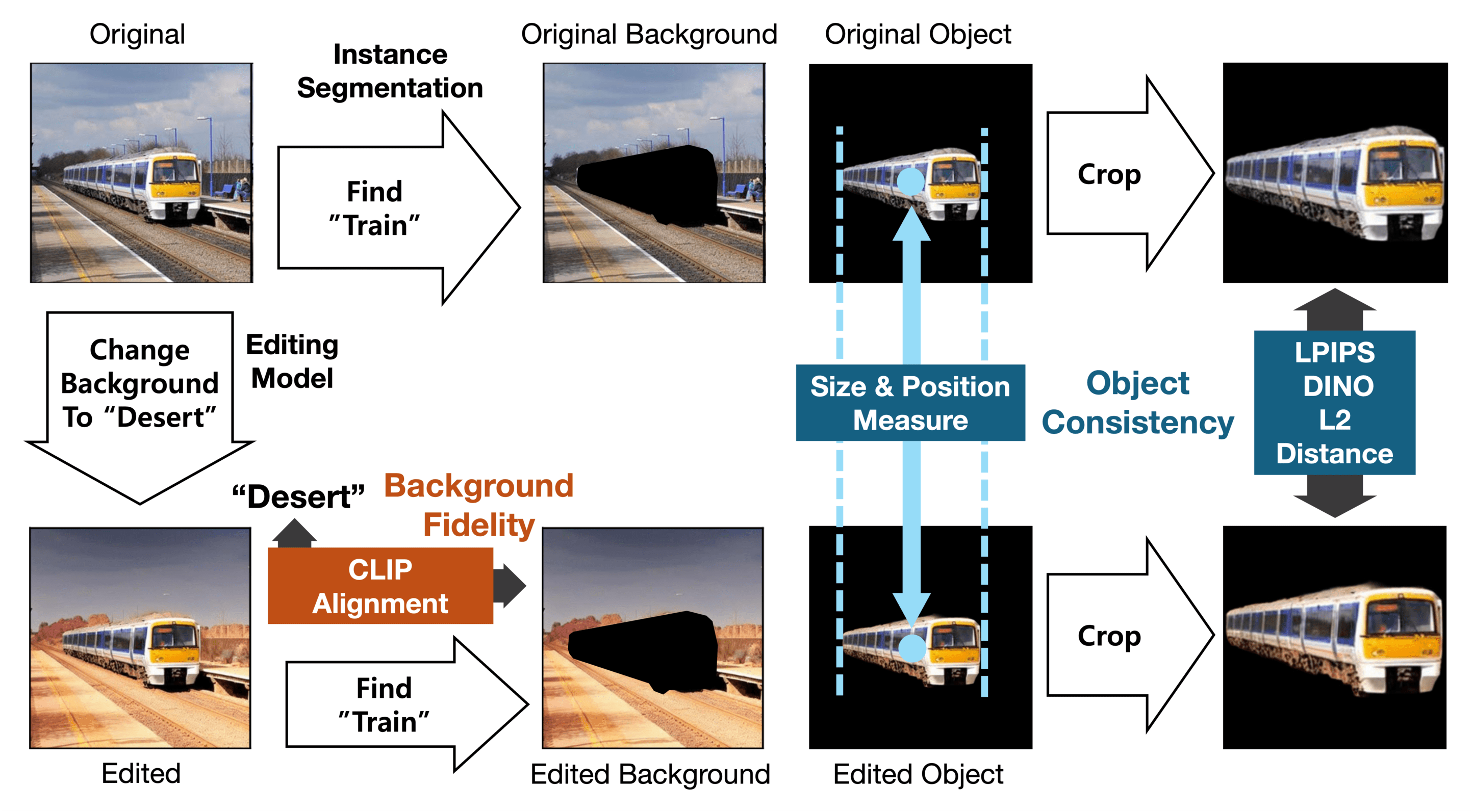}
  \caption{\textsc{Background Change}}
  \label{fig:eval_bg}
\end{subfigure}
\begin{subfigure}{0.60\linewidth}
  \centering
  \includegraphics[width=\textwidth]{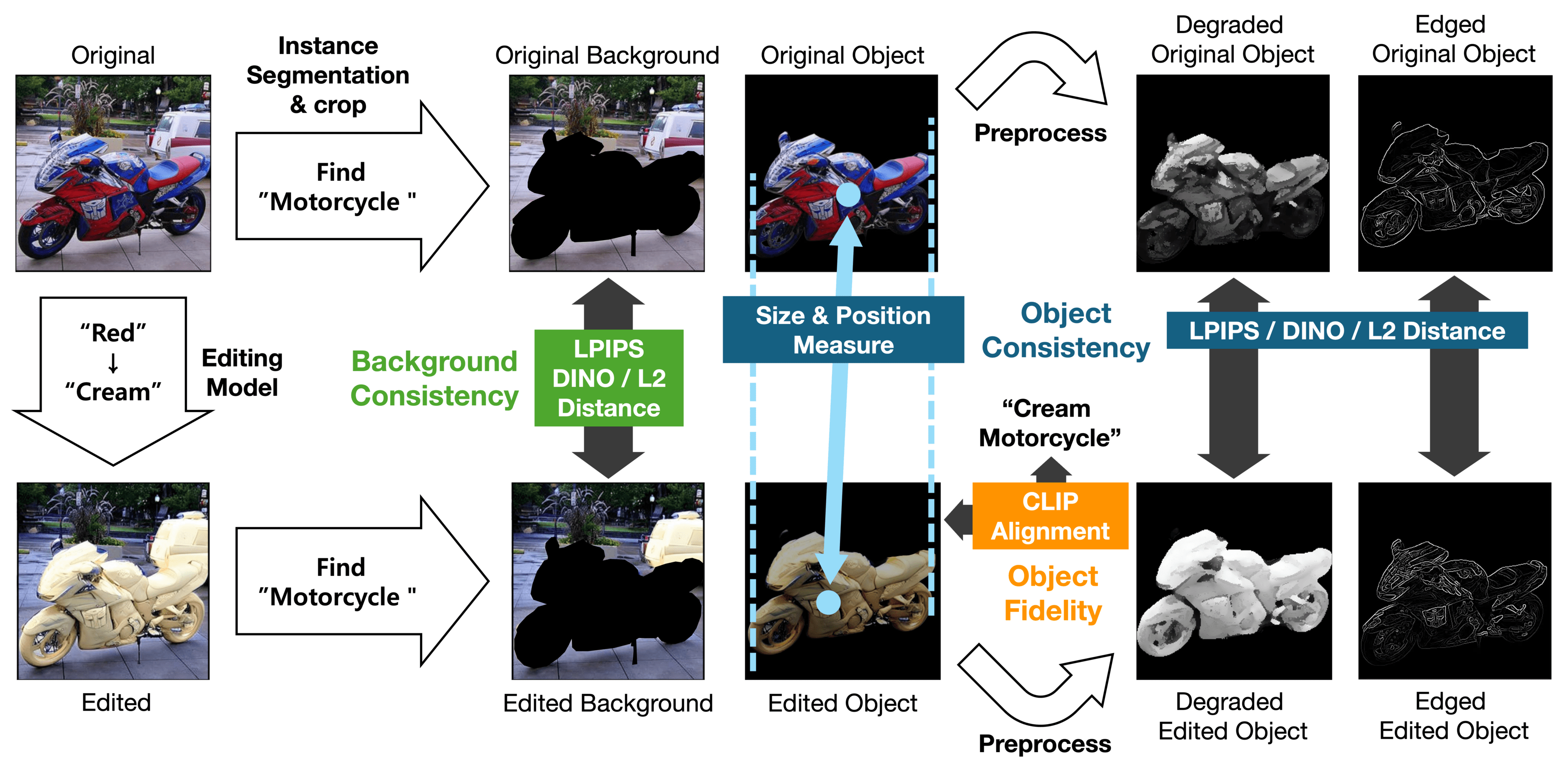}
  \caption{\textsc{Object Attribute Change}}
  \label{fig:eval_attr}
\end{subfigure}
\hfill
\begin{subfigure}{0.35\linewidth}
  \centering
  \includegraphics[width=\textwidth]{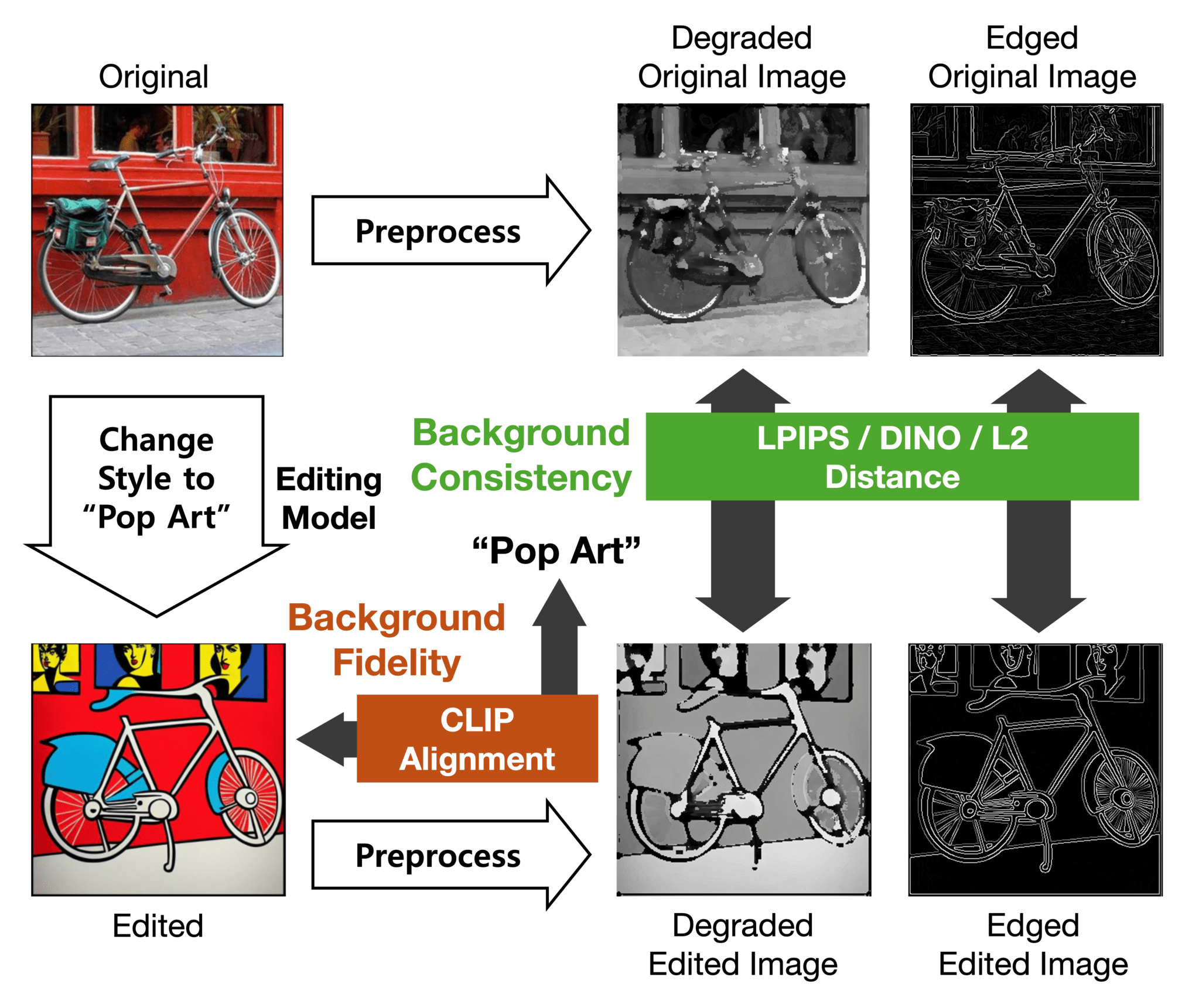}
  \caption{\textsc{Style Change}}
  \label{fig:eval_style}
\end{subfigure}
\vspace{-0.1cm}
\caption{\textbf{Evaluation Workflow Specific to Each Editing Task.} See \cref{sec:eval_workflow} for more details.}
\label{fig:eval_workflow}
\vspace{-0.3cm}
\end{figure*}

\subsection{Evaluation Criteria and Metrics}
\label{sec:method:eval}

A good edit means that the target of the edit is faithfully modified according to the user's intention, while non-target elements 
remain unaltered.
Therefore, we design the evaluation based on three criteria.
First, we assess the general quality of the output images (\textbf{Image Quality}), as an image with poor quality itself is an indicator of suboptimal generation.
Second, it is important to faithfully modify the image according to the instruction (\textbf{Fidelity}).
Lastly, it is also important to verify that the model preserves the context of the original image except for the intended change (\textbf{Consistency}).
With the help of an instance segmentation model \citep{wu2019detectron2},
we further distinguish the latter two for the object and background.
In total, we evaluate in 5 criteria: \textit{Image Quality, Object Fidelity, Background Fidelity, Object Consistency,} and \textit{Background Consistency}.

\vspace{0.1cm} \noindent
\textbf{Image Quality.}
We measure the overall Image Quality (IQ) $\sigma^{IQ}$ through Fr\'{e}chet Inception Distance (FID) \citep{heusel2017gans}.
FID is calculated between input image set and corresponding output image set from the model, and rescale it to range $[0, 1]$ by
\begin{equation}
    \sigma^{IQ} = 1 - \tanh(\mathrm{FID} / 25).
\end{equation}

\noindent The scaling factor 25 is to make the score close to 1 for the state-of-the-art models.

\vspace{0.1cm} \noindent
\textbf{Fidelity.}
For Object Fidelity (OF), we combine the following three metrics.
First, we compute CLIP alignment~\cite{hessel-etal-2021-clipscore} (CLIP embedding similarity between text and image) $\sigma^{OF}_\text{clip}$ between the target object and the desired target text.
Specifically, we detect the target object $o$ 
inside the edited image $I_e$ using an instance segmentation model~\cite{wu2019detectron2}, and zero out the rest (later called the background).
Then, we cut out the target object region with a minimal square enclosing it, denoted by $M_o$.
We compute the CLIP alignment $\sigma^{OF}_\text{clip}$ between $M_o$ and $o$. 
For example, if the query is to recolor a hat to red, we detect and separate the hat ($M_o$) from the output image $I_e$ and calculate CLIP similarity to the target texts ($o$) such as `hat' and `red hat'.

Second, we take the detection confidence $\sigma^{OF}_\text{det}$ from the instance segmentation model for target object, \textit{i.e.}, the hat.
Lastly, we compute size fidelity $\sigma^{OF}_\text{size}$ for the \textsc{Object Resizing} queries, quantifying how much size of the object is appropriately adjusted through a rule-based formula described in \cref{sec:metric_detail}.
The overall Object Fidelity score $\sigma^{OC}$ is a convex combination of the three metrics:

\vspace{-0.1cm}
\begin{equation}
    \sigma^{OF} = w^{OF}_\text{clip} \sigma^{OF}_\text{clip} + w^{OF}_\text{det} \sigma^{OF}_\text{det} + w^{OF}_\text{size} \sigma^{OF}_\text{size},
    \label{eq:score_OF}
\end{equation}

\noindent where the weights $w_*^{OF}$ are fitted so as to align with human annotations, described in \cref{sec:user_align}.

For Background Fidelity (BF), we similarly measure the CLIP alignment $\sigma^{BF}$ between the separated background from $I_e$ and the target text $C'_e$, such as `living room'. 

\vspace{0.1cm} \noindent
\textbf{Consistency.}
Object Consistency (OC) is measured by five scores.
We first segment the target object from the original and edited image through the same process as Object Fidelity, and resize them to the same size.
Then, we compute the LPIPS distance \cite{zhang2018lpips}, DINO embedding similarity \citep{oquab2023dinov2}, and $L_2$ distance, denoted by $\sigma^{OC}_\text{lpips}$, $\sigma^{OC}_\text{dino}$, and $\sigma^{OC}_{\ell 2}$, respectively, between the cropped objects.
We also compute position and size consistency, $\sigma^{OC}_\text{pos}$ and $\sigma^{OC}_\text{size}$, quantifying how much the original position and size of the object is preserved. (See \cref{sec:metric_detail} for details.)
We convert the distance metrics (LPIPS and $L_2$) to similarity by taking inverse, and normalize all metrics to the same range $[0, 1]$.
Similarly, we evaluate Background Consistency (BC) using LPIPS, DINO, and L2 between the original and edited backgrounds, denoted by $\sigma^{BC}_\text{lpips}$, $\sigma^{BC}_\text{dino}$, and $\sigma^{BC}_{\ell 2}$, respectively.

The overall Object and Background Consistency scores, $\sigma^{OC}$ and $\sigma^{BC}$, are a convex combination of the corresponding metrics:
\begin{align}
    \sigma^{OC} &=&& w^{OC}_\text{lpips} \sigma^{OC}_\text{lpips} + w^{OC}_\text{dino} \sigma^{OC}_\text{dino} + w^{OC}_{\ell 2} \sigma^{OC}_{\ell 2} \label{eq:score_OC} \\
    &&&+ w^{OC}_\text{pos} \sigma^{OC}_\text{pos} + w^{OC}_\text{size} \sigma^{OC}_\text{size}, \nonumber \\
    \sigma^{BC} &=&& w^{BC}_\text{lpips} \sigma^{BC}_\text{lpips} + w^{BC}_\text{dino} \sigma^{BC}_\text{dino} + w^{BC}_{\ell 2} \sigma^{BC}_{\ell 2}.
    \label{eq:score_BC}
\end{align}
where the weights $w_*^{OC}$ and $w_*^{BC}$ are learned by user study described in \cref{sec:user_align}.

\vspace{0.1cm} \noindent
\textbf{Total Score.}
We aggregate the five scores into a single metric to derive an overall score of a model.
For the fidelity and consistency scores, which are measured for each single output image, we take the average over all test cases.
Image Quality is computed once from the `set' of the edited outputs.
We aggregate all five scores into a single Total Score $\sigma^\text{Total}$ by convex combination:

\begin{equation}
    \sigma^\text{Total} = \sum_{x \in \mathcal{X}} w^{x} \sigma^{x}
    \label{eq:score_total}
\end{equation}

\noindent where $\mathcal{X} = \{ \text{IQ, OF, BF, OC, BC} \}$, and the weights are fitted to align with human perception in \cref{sec:user_align}.

Some scoring criteria can be inappropriate for specific tasks; \emph{e.g.}, evaluating Object Consistency (OC) for \textsc{Object Replacement} task is a non-sense.
Thus, each task is evaluated only with a pre-defined set of scoring criteria.
\cref{fig:eval_workflow} illustrates properly curated workflow to evaluate each task.

\begin{figure}[t]
    \centering
    \includegraphics[width=\linewidth]{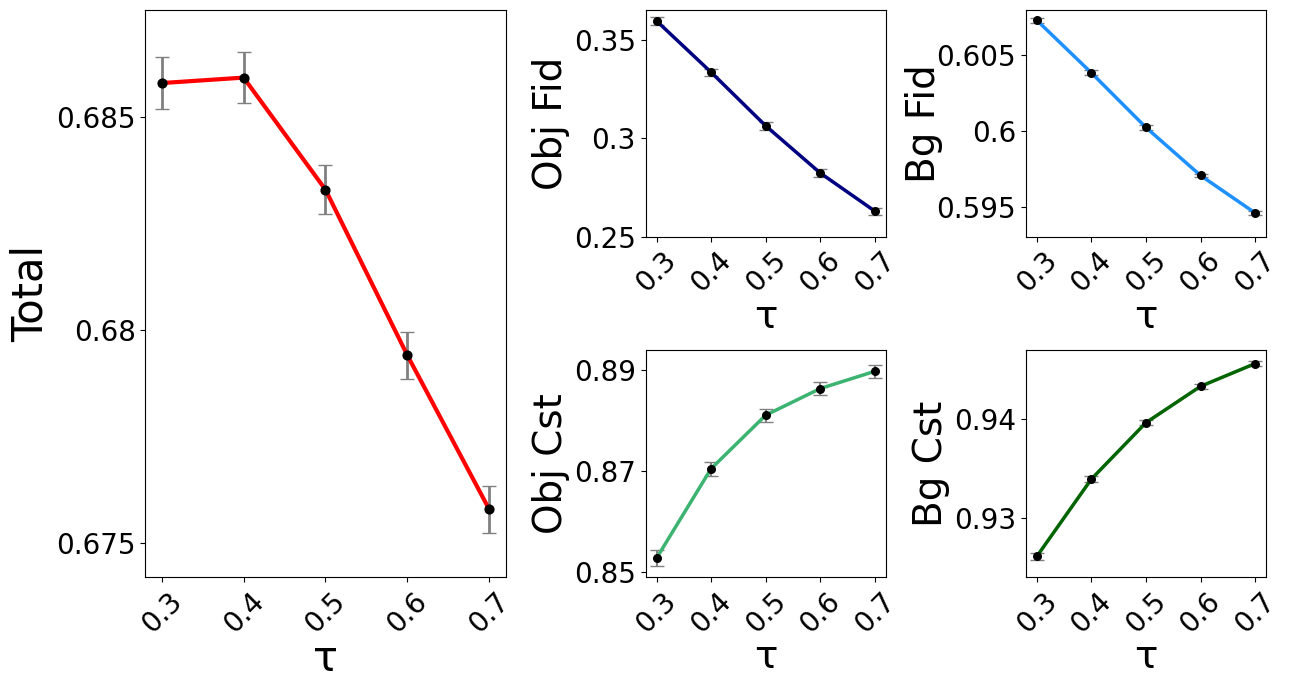}
    \vspace{-0.6cm}
    \caption{\textbf{Aggregated \ours\ scores with Varied Editing Intensity for Prompt-to-Prompt.} Larger $\tau$ means a weaker edit.}
    \label{fig:param_comp}
\vspace{-0.1cm}
\end{figure}

\subsection{Human-aligned Benchmark}
\label{sec:user_align}

In order to align the scores from our evaluation framework with human perception, we conduct a user study to evaluate editing results from several models using the same criteria introduced in \cref{sec:method:eval} (except for Image Quality).
Then, we calibrate the weights to combine metrics so as to agree with this user study results.

\vspace{0.1cm} \noindent
\textbf{User Study Design.}
We sample 4,050 output images edited by 6 description-based models (see \cref{sec:benchmark}) for 2,025 editing queries , $\mathcal{M} = \{m_i\}$, evenly across query types.
Among these, 2,700 images from 1,350 queries are used for this alignment, leaving the rest for test purpose.
These samples are placed into 16 same-sized question sets.

We construct each question to compare edited images by two anonymous models, showing the original image and query as well.
Among the 16 sets, 8 of them ask to judge the overall editing quality, while the rest 8 sets ask one of the four specific criteria (OF, BF, OC, BC; 2 sets per each).
We divide 24 participants into 8 groups, and assign one question set from overall quality assessment and one set from specific criterion per group.
Each question is answered by 3 participants, and we take the majority vote.
\cref{sec:user_study_ex} provides examples of the survey questions.

\vspace{0.1cm} \noindent
\textbf{Fitting the Score Weights.}
For each evaluation criterion $k \in \{OF, BF, OC, BC\}$, we compute the winning rate of each model from the user study result, denoted by $\mathbf{u}^{k} \in [0, 1]^{M}$, where $M$ is the number of competing models.
Suppose now we similarly compute the winning rate using our metrics in Eq.~\eqref{eq:score_OF}-\eqref{eq:score_BC}, denoted by $\mathbf{v}^{k} \in [0, 1]^{M}$.
We choose the weights in Eq.~\eqref{eq:score_OF}-\eqref{eq:score_BC} such that the Pearson correlation between $\mathbf{u}^{k}$ and $\mathbf{v}^{k}$ is maximized, indicating that our evaluation result is best-aligned with the human perception revealed in the user study.
For this, we performed grid search for all possible combinations of weights with step size of 0.01.

The weights to aggregate the five main criteria scores in Eq.~\eqref{eq:score_total} are also determined in a similar way.
That is, we determine the weights such that the winning rate vector made by user study and that by our method are mostly aligned in terms of Pearson correlation.
The degree of fit with user study is plotted in \cref{sec:app_opt}.
We evaluate this alignment using set-aside user study results in \cref{sec:exp:human}.

\section{Empirical Verification}
\label{sec:exp}

We first verify the effectiveness of our proposed benchmark and evaluation framework in two ways: quantitative and qualitative demonstration when the intensity of editing changes (\cref{sec:exp:intensity}) and verification of alignment to the human perception (\cref{sec:exp:human}).

\begin{figure}
    \vspace{-0.1cm}
    \centering
    \includegraphics[width=\linewidth]{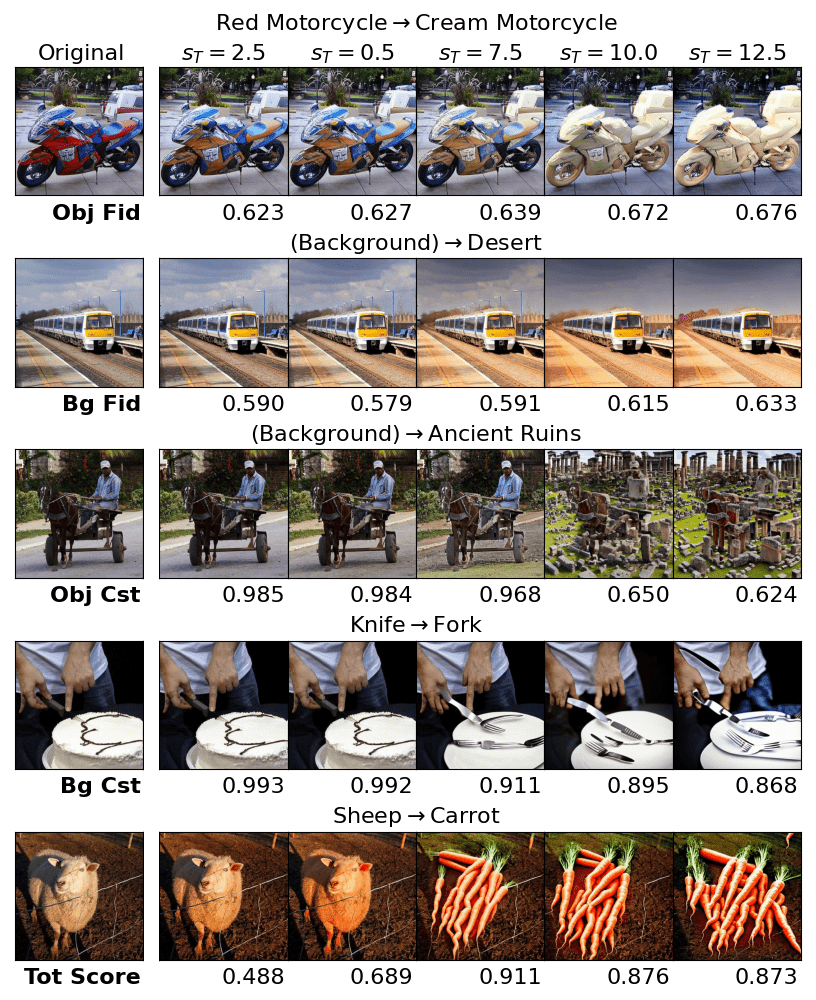}
    \vspace{-0.5cm}
    \caption{\textbf{Demonstration of Our Evaluation Metrics for sample images for each criteria.} Tested model is InstructPix2Pix with $s_T \in \{2.5, 5.0, 7.5, 10.0, 12.5\}$. Higher $s_T$ means stronger edit.}
    \label{fig:ip2pcomp}
    \vspace{-0.2cm}
\end{figure}

\subsection{Benchmark with Varied Edit Intensity}
\label{sec:exp:intensity}

To verify our benchmark accurately responds to the model characteristics and performance changes, we vary the hyperparameters controlling the degree of edits in several models; specifically, the interpolation parameter $\eta$ (Imagic~\cite{kawar2023imagic}), text guidance scale $s_T$ (MasaCtrl~\cite{cao2023masactrl}, FreeDiff~\cite{wu2024freediff}, and InstructPix2Pix (IP2P)~\cite{brooks2023instructpix2pix}), and self-replace step $\tau$ (Prompt-to-Prompt (P2P) with null-text inversion~\citep{mokady2023null}). More details are provided in \cref{sec:app:add:detail_res}.

\vspace{0.1cm} \noindent
\textbf{Quantitative Analysis.}
\cref{fig:param_comp} shows the change of aggregated benchmark scores of P2P model as its parameter $\tau$ changes (lower $\tau$ means stronger edit).
As $\tau$ increases (the intended edit intensity decreases), two fidelity scores decrease while two consistency scores increase.
The total scores show trade-off between fidelity and consistency, and the optimal setting is found where the two aspects are balanced ($\tau$ = 0.4).
From this result, we conclude that our \ours\ precisely captures the model's editing performance comprehensively, as well as across each specific aspect.
It is also notable that the error of each point, measured by bootstrapping, is small, indicating not only that \ours\ benchmark is stable, but also that our benchmark is sensitive enough to distinguish the performance gap between the models.
This implies that our benchmark is suitable for fine-grained comparison of the models.
A similar trend is observed with other models, reported in Appendix \ref{sec:app:add:detail_res}.

More practically, as our framework can analyze throughout hyperparameter space, it would be also more reliable for finding the optimal hyperparameters of model than manual exploration.

\begin{figure}[t]
    \vspace{-0.1cm}
    \centering
    \includegraphics[width=\linewidth]{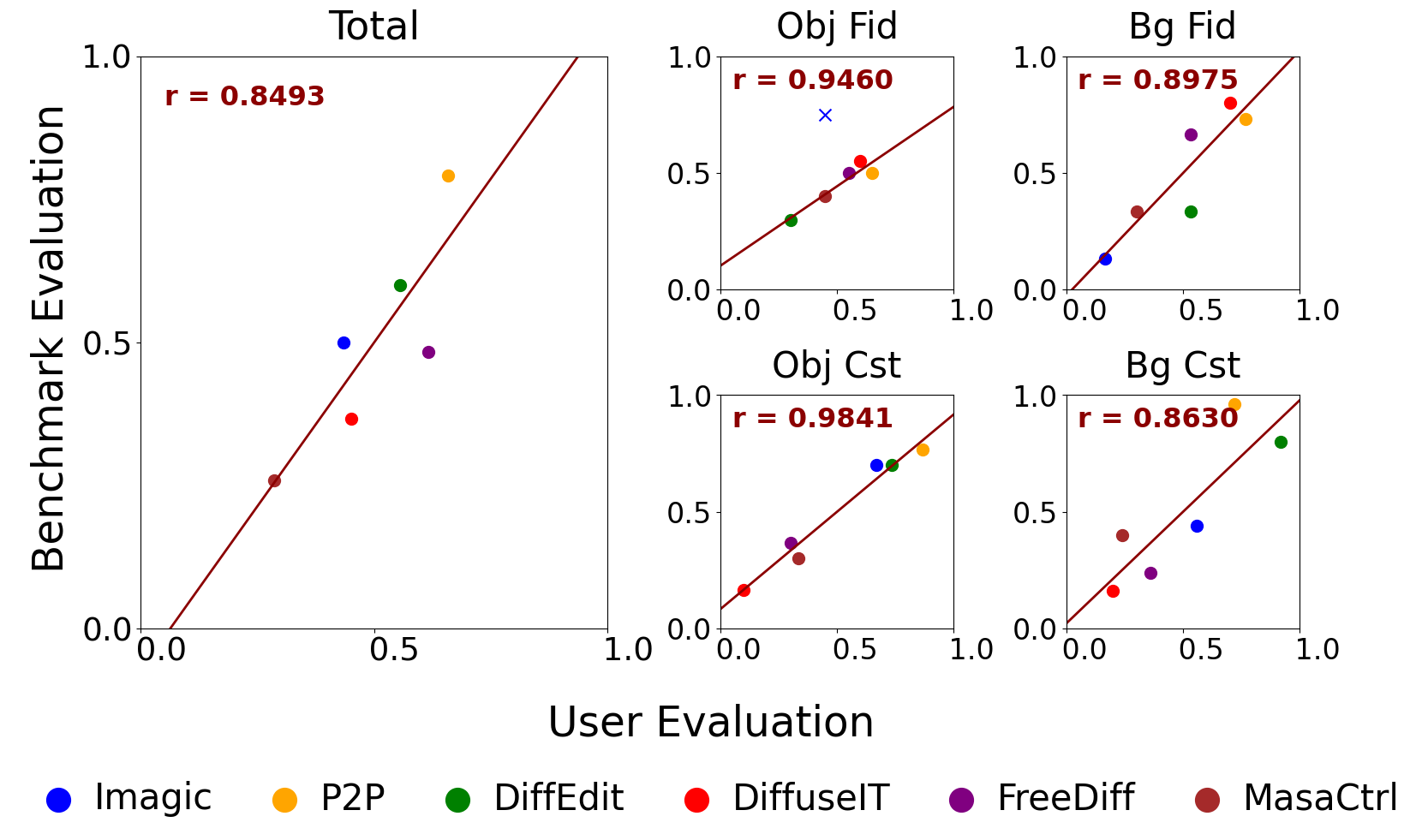}
    \vspace{-0.7cm}
    \caption{\textbf{Relation between winning rates by users and \ours,} measured on the user study test set. The least square linear fit (red line) and Pearson's correlation coefficient ($r$) are reported.
    The cross mark in object fidelity figure exhibits an outlier which has been excluded from correlation calculation.}
    \vspace{0.3cm}
    \label{fig:user_alignment_test}
\end{figure}

\vspace{0.1cm} \noindent
\textbf{Qualitative Demonstration.}
\cref{fig:ip2pcomp} illustrates evaluation samples with IP2P across different text guidance scales $s_T$.
As the edit intensity gets stronger (higher $s_T$), model well reflects the intended edit more clearly and lose its original context to some extent.
Our fidelity and consistency scores successfully reflect this trend, and the aggregated Total score clearly distinguishes successful and failed edits.
This demonstrates that our \ours\ cognitively follows the trend of editing quality in each field.

\begin{table}[t]
    \centering \renewcommand{\tabcolsep}{2pt}
    \resizebox{\linewidth}{!}{
    \small
    \begin{tabular}{cccccc}
        \toprule
        Correlation & \parbox{1.5cm}{\centering Object\\Fidelity} & \parbox{1.5cm}{\centering Background\\Fidelity} & \parbox{1.5cm}{\centering Object\\Consistency} & \parbox{1.5cm}{\centering Background\\Consistency} & \parbox{1.5cm}{\centering Total\\Score} \\
        \midrule
        $\rho$ & 0.8208 & 0.8971 & 0.9276 & 0.8857 & 0.7143 \\
        $\tau$ & 0.7379 & 0.7857 & 0.8281 & 0.7333 & 0.6000 \\
        \bottomrule
    \end{tabular}
    }
    \vspace{-0.2cm}
    \caption{\textbf{Correlation coefficients between model winning rates from user study and our metrics} on the user study test set.}
    \label{tab:corr_test}
    \vspace{0.3cm}
\end{table}

\begin{table}
  \centering
  \resizebox{0.8\linewidth}{!}{
  \small
  \begin{tabular}{lccc}
    \toprule
    Metric & $r$ & $\rho$ & $\tau$ \\
    \midrule
    CLIP Alignment & -0.3410 & -0.2000 & -0.2000 \\
    Detection Rate & 0.4293 & 0.5218 & 0.4140 \\
    \midrule
    LPIPS & 0.5468 & 0.4857 & 0.3333 \\
    DINO & 0.3367 & 0.0286 & 0.0667 \\
    $L_2$ & 0.4599 & 0.2000 & 0.0667 \\
    Position Score & -0.1497 & -0.0580 & 0.0000 \\
    Size Score & -0.3900 & -0.3714 & -0.2000 \\
    Detection Confidence & 0.0287 & -0.0857 & -0.0667 \\
    FID & 0.5058 & 0.3714 & 0.3333 \\
    \bottomrule
  \end{tabular}
  }
  \vspace{-0.1cm}
  \caption{\textbf{Correlation coefficients between model winning rates from user study and conventional metrics} on overall edit quality assessments in test set survey.}
  \label{tab:corr_conven}
  \vspace{-0.1cm}
\end{table}

\subsection{Alignment to Human Perception}
\label{sec:exp:human}

We evaluate the correlation between our benchmark scores and human evaluation on a set-aside user study results which are not used to fit the weights in \cref{sec:user_align}.
Among the collected user annotations for 4,050 images from 2,025 edit queries in user study, 1,350 images for 675 queries were left for test purpose, as mentioned in \cref{sec:user_align}.
These are split into 4 question sets for overall quality assessment and 4 sets for the specific criteria (1 set for each).

\begin{table*}[t]
  \vspace{-0.1cm}
  \centering
  \begin{minipage}{\textwidth}
    \centering
    \small
    \resizebox{\textwidth}{!}{
    {\footnotesize
    \begin{tabular}{ll p{2cm}<{\centering} p{2cm}<{\centering} p{2cm}<{\centering} p{2cm}<{\centering} p{2cm}<{\centering} p{2cm}<{\centering}}
    \toprule[1.5pt]
    \multicolumn{2}{c}{\textbf{Models}} & \textbf{\parbox{2cm}{\centering Object\\Fidelity}} & \textbf{\parbox{2cm}{\centering Background\\Fidelity}} & \textbf{\parbox{2cm}{\centering Object\\Consistency}} & \textbf{\parbox{2cm}{\centering Background\\Consistency}} & \textbf{\parbox{2cm}{\centering Image\\Quality}} & \textbf{\parbox{2cm}{\centering Total\\Score}} \\ \midrule
    \multirow{6}{*}{\parbox{2cm}{\centering \textbf{Description\\-based}}} & DiffEdit & 0.2277 $\pm$ 0.0018 & 0.5910 $\pm$ 0.0001 & 0.8338 $\pm$ 0.0018 & \textbf{0.9608 $\pm$ 0.0002} & 0.7477 $\pm$ 0.0024 & 0.6552 $\pm$ 0.0006
     \\
     & DiffuseIT & 0.3202 $\pm$ 0.0019 & 0.6045 $\pm$ 0.0001 & 0.8616 $\pm$ 0.0012 & 0.8958 $\pm$ 0.0002 & 0.5569 $\pm$ 0.0021 & 0.6682 $\pm$ 0.0006
     \\ 
     & FreeDiff & 0.3174 $\pm$ 0.0020 & 0.5964 $\pm$ 0.0001 & 0.8631 $\pm$ 0.0013 & 0.9150 $\pm$ 0.0002 & 0.7180 $\pm$ 0.0014 & 0.6739 $\pm$ 0.0006
     \\ 
     & P2P & \textbf{0.3595 $\pm$ 0.0020} & \textbf{0.6073 $\pm$ 0.0002} & 0.8528 $\pm$ 0.0015 & 0.9261 $\pm$ 0.0003 & 0.6551 $\pm$ 0.0023 & \textbf{0.6858 $\pm$ 0.0006}
     \\ 
     & Imagic & 0.2271 $\pm$ 0.0017 & 0.5878 $\pm$ 0.0001 & \textbf{0.9237 $\pm$ 0.0009} & 0.9201 $\pm$ 0.0003 & \textbf{0.8406 $\pm$ 0.0014} & 0.6682 $\pm$ 0.0005
     \\ 
     & MasaCtrl & 0.2332 $\pm$ 0.0019 & 0.5967 $\pm$ 0.0001 & 0.6754 $\pm$ 0.0023 & 0.8954 $\pm$ 0.0003 & 0.2685 $\pm$ 0.0022 & 0.5936 $\pm$ 0.0007
     \\ \midrule\midrule
    \multirow{3}{*}{\parbox{2cm}{\centering \textbf{Instruction\\-based}}} & MagicBrush & \textbf{0.5378 $\pm$ 0.0020} & 0.6196 $\pm$ 0.0002 & \textbf{0.8259 $\pm$ 0.0018} & \textbf{0.9513 $\pm$ 0.0003} & \textbf{0.6977 $\pm$ 0.0032} & \textbf{0.7329 $\pm$ 0.0007}
     \\ 
     & InstDiff & 0.4596 $\pm$ 0.0022 & \textbf{0.6205 $\pm$ 0.0001} & 0.6870 $\pm$ 0.0022 & 0.9090 $\pm$ 0.0005 & 0.4148 $\pm$ 0.0039 & 0.6639 $\pm$ 0.0008
     \\ 
     & IP2P & 0.4474 $\pm$ 0.0021 & 0.6169 $\pm$ 0.0002 & 0.7903 $\pm$ 0.0021 & 0.9319 $\pm$ 0.0003 & 0.6658 $\pm$ 0.0031 & 0.6960 $\pm$ 0.0007
     \\ \bottomrule[1.5pt]
    \end{tabular}
    }
    }
    \vspace{-0.3cm}
    \caption{\ours\ benchmark results across score criteria}
    \label{tab:score}
  \end{minipage} \\[0.1cm] 
  
  \begin{minipage}{\textwidth}
    \centering
    \small
    \resizebox{\textwidth}{!}{
    {\footnotesize
    \begin{tabular}{ll p{2cm}<{\centering} p{2cm}<{\centering} p{2cm}<{\centering} p{2cm}<{\centering} p{2cm}<{\centering} p{2cm}<{\centering}}
    \toprule[1.5pt]
    \multicolumn{2}{c}{\textbf{Models}} & \textbf{\parbox{2cm}{\centering Object\\Addition}} & \textbf{\parbox{2cm}{\centering Object\\Replacement}} & \textbf{\parbox{2cm}{\centering Object\\Resizing}} & \textbf{\parbox{2cm}{\centering Attribute\\Change}} & \textbf{\parbox{2cm}{\centering Background\\Change}} & \textbf{\parbox{2cm}{\centering Style\\Change}} \\ \midrule
    \multirow{6}{*}{\parbox{2cm}{\centering \textbf{Description\\-based}}} & DiffEdit & 0.5512 $\pm$ 0.0013 & 0.5190 $\pm$ 0.0012 & 0.6666 $\pm$ 0.0016 & 0.7687 $\pm$ 0.0014 & 0.7382 $\pm$ 0.0012 & \textbf{0.7517 $\pm$ 0.0003}
     \\ 
     & DiffuseIT & 0.5452 $\pm$ 0.0016 & 0.5691 $\pm$ 0.0020 & 0.6780 $\pm$ 0.0015 & 0.7919 $\pm$ 0.0009 & 0.7319 $\pm$ 0.0010 & 0.7208 $\pm$ 0.0003
     \\ 
     & FreeDiff & 0.5101 $\pm$ 0.0013 & 0.5759 $\pm$ 0.0022 & \textbf{0.7054 $\pm$ 0.0016} & 0.7896 $\pm$ 0.0010 & 0.7327 $\pm$ 0.0011 & 0.7411 $\pm$ 0.0003
     \\ 
     & P2P & \bf{0.6246 $\pm$ 0.0020} & \bf{0.6088 $\pm$ 0.0023} & 0.6899 $\pm$ 0.0015 & 0.8031 $\pm$ 0.0010 & 0.7111 $\pm$ 0.0015 & 0.7374 $\pm$ 0.0004
     \\ 
     & Imagic & 0.4964 $\pm$ 0.0010 & 0.4689 $\pm$ 0.0010 & 0.6863 $\pm$ 0.0012 & \textbf{0.8244 $\pm$ 0.0006} & \bf{0.7585 $\pm$ 0.0008} & 0.7304 $\pm$ 0.0003
     \\ 
     & MasaCtrl & 0.4829 $\pm$ 0.0011 & 0.4804 $\pm$ 0.0012 & 0.6388 $\pm$ 0.0020 & 0.6739 $\pm$ 0.0017 & 0.6376 $\pm$ 0.0020 & 0.7351 $\pm$ 0.0003
     \\ \midrule\midrule
    \multirow{3}{*}{\parbox{2cm}{\centering \textbf{Instruction\\-based}}} & MagicBrush & \textbf{0.7461 $\pm$ 0.0018} & \textbf{0.8189 $\pm$ 0.0023} & 0.7279 $\pm$ 0.0016 & \textbf{0.8407 $\pm$ 0.0005} & \textbf{0.6536 $\pm$ 0.0023} & \textbf{0.7471 $\pm$ 0.0004}
     \\ 
     & InstDiff & 0.6260 $\pm$ 0.0020 & 0.6656 $\pm$ 0.0025 & \textbf{0.7488 $\pm$ 0.0019} & 0.7982 $\pm$ 0.0009 & 0.5206 $\pm$ 0.0024 & 0.7066 $\pm$ 0.0005
     \\ 
     & IP2P & 0.6771 $\pm$ 0.0021 & 0.7511 $\pm$ 0.0025 & 0.6929 $\pm$ 0.0012 & 0.8295 $\pm$ 0.0007 & 0.5800 $\pm$ 0.0024 & 0.7407 $\pm$ 0.0004
     \\ \bottomrule[1.5pt]
    \end{tabular}
    }
    }
    \vspace{-0.3cm}
    \caption{\ours\ benchmark results across various edit types}
    \label{tab:per_task_scores}
  \end{minipage} \\[0.1cm]

  \begin{minipage}{\textwidth}
    \centering
    \small
    \resizebox{\textwidth}{!}{
    {\footnotesize \renewcommand{\tabcolsep}{2pt}
    \begin{tabular}{ll p{2cm}<{\centering} p{2cm}<{\centering} p{2cm}<{\centering} p{2cm}<{\centering} p{2cm}<{\centering} p{2cm}<{\centering} p{2cm}<{\centering} p{2cm}<{\centering}}
    \toprule[1.5pt]
    \multicolumn{2}{c}{\textbf{Models}} & \textbf{\parbox{2cm}{\centering Whole Image}} & \textbf{\parbox{2cm}{\centering Person}} & \textbf{\parbox{2cm}{\centering Animal}} & \textbf{\parbox{2cm}{\centering Vehicle}} & \textbf{\parbox{2cm}{\centering Household}} & \textbf{\parbox{2cm}{\centering Dining}} & \textbf{\parbox{2cm}{\centering Outdoor}} & \textbf{\parbox{2cm}{\centering Other}} \\ \midrule
    \multirow{6}{*}{\parbox{2cm}{\centering \textbf{Description\\-based}}} & DiffEdit & 0.7963 $\pm$ 0.0008 & 0.7087 $\pm$ 0.0026 & 0.6506 $\pm$ 0.0026 & 0.6980 $\pm$ 0.0022 & 0.6936 $\pm$ 0.0021 & 0.6344 $\pm$ 0.0031 & 0.6556 $\pm$ 0.0026 & 0.6618 $\pm$ 0.0029 \\
     & DiffuseIT & 0.7669 $\pm$ 0.0007 & 0.7090 $\pm$ 0.0024 & 0.7293 $\pm$ 0.0017 & 0.7141 $\pm$ 0.0019 & 0.6734 $\pm$ 0.0021 & 0.6735 $\pm$ 0.0024 & 0.6909 $\pm$ 0.0022 & 0.6817 $\pm$ 0.0024 \\ 
     & FreeDiff & 0.7837 $\pm$ 0.0007 & \textbf{0.7208 $\pm$ 0.0026} & 0.7268 $\pm$ 0.0019 & 0.7175 $\pm$ 0.0021 & 0.6894 $\pm$ 0.0022 & 0.6534 $\pm$ 0.0028 & 0.6951 $\pm$ 0.0022 & 0.6880 $\pm$ 0.0026 \\ 
     & P2P & 0.7632 $\pm$ 0.0010 & 0.7179 $\pm$ 0.0024 & \textbf{0.7428 $\pm$ 0.0021} & \textbf{0.7448 $\pm$ 0.0020} & \textbf{0.7154 $\pm$ 0.0022} & \textbf{0.6922 $\pm$ 0.0029} & \textbf{0.7224 $\pm$ 0.0021} & \textbf{0.7164 $\pm$ 0.0026} \\ 
     & Imagic & \textbf{0.7967 $\pm$ 0.0006} & 0.7085 $\pm$ 0.0021 & 0.7003 $\pm$ 0.0016 & 0.7043 $\pm$ 0.0017 & 0.6977 $\pm$ 0.0017 & 0.6619 $\pm$ 0.0021 & 0.6984 $\pm$ 0.0018 & 0.6819 $\pm$ 0.0021 \\ 
     & MasaCtrl & 0.7163 $\pm$ 0.0013 & 0.6819 $\pm$ 0.0032 & 0.6197 $\pm$ 0.0026 & 0.6527 $\pm$ 0.0027 & 0.5527 $\pm$ 0.0032 & 0.5507 $\pm$ 0.0036 & 0.5925 $\pm$ 0.0029 & 0.5771 $\pm$ 0.0036 \\ \midrule\midrule
    \multirow{3}{*}{\parbox{2cm}{\centering \textbf{Instruction\\-based}}} & MagicBrush & \textbf{0.7273 $\pm$ 0.0015} & \textbf{0.7613 $\pm$ 0.0027} & \textbf{0.8254 $\pm$ 0.0016} & \textbf{0.8162 $\pm$ 0.0018} & \textbf{0.7841 $\pm$ 0.0019} & \textbf{0.8084 $\pm$ 0.0020} & \textbf{0.8064 $\pm$ 0.0017} & \textbf{0.7935 $\pm$ 0.0022} \\ 
     & InstDiff & 0.6114 $\pm$ 0.0016 & 0.7520 $\pm$ 0.0031 & 0.7770 $\pm$ 0.0020 & 0.7488 $\pm$ 0.0022 & 0.6976 $\pm$ 0.0023 & 0.7206 $\pm$ 0.0027 & 0.7560 $\pm$ 0.0020 & 0.7252 $\pm$ 0.0025 \\ 
     & IP2P & 0.6749 $\pm$ 0.0016 & 0.7214 $\pm$ 0.0019 & 0.7786 $\pm$ 0.0017 & 0.7726 $\pm$ 0.0019 & 0.7635 $\pm$ 0.0021 & 0.7808 $\pm$ 0.0022 & 0.7829 $\pm$ 0.0019 & 0.7689 $\pm$ 0.0023 \\ 
    \bottomrule[1.5pt]
    \end{tabular}
    }
    }
    \vspace{-0.3cm}
    \caption{\ours\ benchmark results across target object classes}
    \label{tab:per_class_scores}
  \end{minipage}
  \vspace{-0.4cm}
\end{table*}

We obtain the correlation between the model winning rates by user study and those by \ours\ metrics, in a similar manner to \cref{sec:user_align}, plotted in \cref{fig:user_alignment_test}.
Results from ours are successfully aligned with human evaluation results with strong correlation to each other (Pearson's coefficient $r$ over 0.8) throughout every criterion.
Two other types of correlation coefficients (Spearman's $\rho$ and Kendall's $\tau$) are reported in \cref{tab:corr_test}, showing a similar trend.
Overall, our benchmark is well-aligned with user survey results, providing reliable evaluations close to human perception. We also provide alignment result tested on unseen new dataset in \cref{sec:app:add:imagenet}

\noindent \textbf{Comparison with Conventional Metrics.}
We compare our benchmark's human alignment results with two conventional metrics used in previous studies: (1) CLIP alignment, which measures the similarity between the target caption ($C_e$) and the edited image ($I_e$), as used in \cite{kawar2023imagic,hertz2023p2p} 
, and (2) detection rate used in \cite{basu2023editval} which evaluates the presence of objects generated by the queries, 1 if object is detected and 0 if not. 
We also compare our results to each metrics used in \ours. Within \ours\ framework, instead of aggregating multiple metric scores, we use each metric alone and compare the results with the user evaluation.

\cref{tab:corr_conven} shows the correlation coefficients between user evaluation and conventional metrics on the overall edit quality in user study test set in the upper section, and same data obtained with single metrics in the lower section.
This clearly shows much inferior alignment with human evaluation results, compared to ours in \cref{tab:corr_test}.
Our structured and combined use of conventional metrics enables us to successfully adjust benchmarks to better align with human perceptions.

\section{\ours\ Benchmark Results and Discussion}
\label{sec:benchmark}

In this section, we evaluate image editing models using the proposed framework.
We compare 6 description-based editing models (Imagic, P2P, DiffEdit~\cite{couairon2022diffedit}, DiffuseIT~\cite{kwon2022diffusion}, FreeDiff, and MasaCtrl) and 3 instruction-based models (IP2P, InstructDiffusion (InstDiff)~\cite{geng2024instructdiffusion}, and MagicBrush~\cite{zhang2024magicbrush}).
For models with adjustable parameters, we choose the default value found in the officially released code ($s_T = 7.5$, $s_I = 1.5$ and $\tau = 0.3$). 
For Imagic, we choose $\eta = 0.6$, which we achieve the best result.

\cref{tab:score} shows the benchmark performance of the competing models, both for description-based and instruction-based.
Among the description-based models, Imagic and DiffEdit tend to be more competitive in consistency, while DiffuseIT and P2P are more competent in fidelity.
P2P turns out to be the best performing all-round, showing outstanding results both for fidelity and consistency.
Among the instruction-based models, MagicBrush performs the best in most areas, leading to the highest total score.
It is also notable that models with high consistency scores tend to show great image quality score in both description and instruction-based group.
(We emphasize again that models using different caption types are not fairly comparable.)

\cref{tab:per_task_scores} and \ref{tab:per_class_scores} show the benchmark scores aggregated separately by edit types and edited object classes. ``Whole Image'' in \cref{tab:per_class_scores} stands for targets of non-object centric edits.
From the tables, we can see that each model often has its strong and weak points.
For example, looking at \cref{tab:per_task_scores}, Imagic tends to produce high quality image, performing \textsc{Attribute Change} and \textsc{Background Change} relatively better, while showing weakness on \textsc{Object Addition} or \textsc{Replacement}.
InstDiff underperforms in most edit tasks compared to other instruction-based models, but is particularly strong for \textsc{Object Resizing}.
In \cref{tab:per_class_scores}, P2P performs well on most of the objects, while it has weakness in whole image scale edits.
Oppositely, Imagic underperforms in every target classes except for edits on whole image.

\section{Summary and Limitations}
\label{sec:con}

We present a novel benchmark \ours, which includes a vast dataset covering a wide range of editing types and object classes.
Our evaluation scheme provides accurate assessment through a structured use of multiple metrics in categorized criteria, quantifying the performance of an editing model in a manner similar to human perception. 
\ours\ allows precise evaluation and analysis of text-guided image editing model, and we hope it will guide the future direction of image editing research.

\vspace{0.1cm} \noindent
\textbf{Limitations.}
\ours\ aims to cover wide range of editing tasks, but some are excluded. Object removal for description-based models and movement/rotation are omitted due to challenges in generating captions and defining reference points. Certain attribute types are excluded due to difficulties in localization. See \cref{sec:limit} for details.


\subsection*{Acknowledgments}
This work was supported by Samsung Electronics (IO240512-09881-01), Youlchon Foundation, 
by SOFT Foundry Institute at SNU, and NRF grants (RS-2021-NR05515, RS-2024-00336576, RS-2023-00222663) and IITP grants (RS-2024-00353131, RS-2022-II220264), funded by the government of Korea.


\begin{thebibliography}{35}
\providecommand{\natexlab}[1]{#1}
\providecommand{\url}[1]{\texttt{#1}}
\expandafter\ifx\csname urlstyle\endcsname\relax
  \providecommand{\doi}[1]{doi: #1}\else
  \providecommand{\doi}{doi: \begingroup \urlstyle{rm}\Url}\fi

\bibitem[Avrahami et~al.(2022)Avrahami, Lischinski, and Fried]{avrahami2022blended}
Omri Avrahami, Dani Lischinski, and Ohad Fried.
\newblock Blended diffusion for text-driven editing of natural images.
\newblock In \emph{CVPR}, 2022.

\bibitem[Basu et~al.(2023)Basu, Saberi, Bhardwaj, Chegini, Massiceti, Sanjabi, Hu, and Feizi]{basu2023editval}
Samyadeep Basu, Mehrdad Saberi, Shweta Bhardwaj, Atoosa~Malemir Chegini, Daniela Massiceti, Maziar Sanjabi, Shell~Xu Hu, and Soheil Feizi.
\newblock Edit{V}al: Benchmarking diffusion based text-guided image editing methods.
\newblock \emph{arXiv:2310.02426}, 2023.

\bibitem[Brooks et~al.(2023)Brooks, Holynski, and Efros]{brooks2023instructpix2pix}
Tim Brooks, Aleksander Holynski, and Alexei~A Efros.
\newblock Instruct{P}ix2{P}ix: Learning to follow image editing instructions.
\newblock In \emph{CVPR}, 2023.

\bibitem[Canny(1986)]{canny1986computational}
John Canny.
\newblock A computational approach to edge detection.
\newblock \emph{IEEE Transactions on pattern analysis and machine intelligence}, 8\penalty0 (6):\penalty0 679--698, 1986.

\bibitem[Cao et~al.(2023)Cao, Wang, Qi, Shan, Qie, and Zheng]{cao2023masactrl}
Mingdeng Cao, Xintao Wang, Zhongang Qi, Ying Shan, Xiaohu Qie, and Yinqiang Zheng.
\newblock Masa{C}trl: Tuning-free mutual self-attention control for consistent image synthesis and editing.
\newblock In \emph{ICCV}, 2023.

\bibitem[Couairon et~al.(2023)Couairon, Verbeek, Schwenk, and Cord]{couairon2022diffedit}
Guillaume Couairon, Jakob Verbeek, Holger Schwenk, and Matthieu Cord.
\newblock Diff{E}dit: Diffusion-based semantic image editing with mask guidance.
\newblock In \emph{ICLR}, 2023.

\bibitem[Deng et~al.(2009)Deng, Dong, Socher, Li, Li, and Fei-Fei]{deng2009imagenet}
Jia Deng, Wei Dong, Richard Socher, Li-Jia Li, Kai Li, and Li Fei-Fei.
\newblock Imagenet: A large-scale hierarchical image database.
\newblock In \emph{CVPR}, 2009.

\bibitem[Dubey et~al.(2024)Dubey, Jauhri, Pandey, Kadian, Al-Dahle, Letman, Mathur, Schelten, Yang, Fan, et~al.]{dubey2024llama}
Abhimanyu Dubey, Abhinav Jauhri, Abhinav Pandey, Abhishek Kadian, Ahmad Al-Dahle, Aiesha Letman, Akhil Mathur, Alan Schelten, Amy Yang, Angela Fan, et~al.
\newblock The llama 3 herd of models.
\newblock \emph{arXiv:2407.21783}, 2024.

\bibitem[Geng et~al.(2024)Geng, Yang, Hang, Li, Gu, Zhang, Bao, Zhang, Li, Hu, et~al.]{geng2024instructdiffusion}
Zigang Geng, Binxin Yang, Tiankai Hang, Chen Li, Shuyang Gu, Ting Zhang, Jianmin Bao, Zheng Zhang, Houqiang Li, Han Hu, et~al.
\newblock Instruct{D}iffusion: A generalist modeling interface for vision tasks.
\newblock In \emph{CVPR}, 2024.

\bibitem[Hertz et~al.(2023)Hertz, Mokady, Tenenbaum, Aberman, Pritch, and Cohen-Or]{hertz2023p2p}
Amir Hertz, Ron Mokady, Jay Tenenbaum, Kfir Aberman, Yael Pritch, and Daniel Cohen-Or.
\newblock Prompt-to-prompt image editing with cross attention control.
\newblock In \emph{ICLR}, 2023.

\bibitem[Hessel et~al.(2021)Hessel, Holtzman, Forbes, Le~Bras, and Choi]{hessel-etal-2021-clipscore}
Jack Hessel, Ari Holtzman, Maxwell Forbes, Ronan Le~Bras, and Yejin Choi.
\newblock {CLIPS}core: A reference-free evaluation metric for image captioning".
\newblock In \emph{EMNLP}, 2021.

\bibitem[Heusel et~al.(2017)Heusel, Ramsauer, Unterthiner, Nessler, and Hochreiter]{heusel2017gans}
Martin Heusel, Hubert Ramsauer, Thomas Unterthiner, Bernhard Nessler, and Sepp Hochreiter.
\newblock {GAN}s trained by a two time-scale update rule converge to a local nash equilibrium.
\newblock \emph{NIPS}, 2017.

\bibitem[Huang et~al.(2024)Huang, Xie, Wang, Yuan, Cun, Ge, Zhou, Dong, Huang, Zhang, et~al.]{huang2024smartedit}
Yuzhou Huang, Liangbin Xie, Xintao Wang, Ziyang Yuan, Xiaodong Cun, Yixiao Ge, Jiantao Zhou, Chao Dong, Rui Huang, Ruimao Zhang, et~al.
\newblock Smart{E}dit: Exploring complex instruction-based image editing with multimodal large language models.
\newblock In \emph{CVPR}, 2024.

\bibitem[Hudson and Manning(2019)]{hudson2019gqa}
Drew~A Hudson and Christopher~D Manning.
\newblock {GQA}: A new dataset for real-world visual reasoning and compositional question answering.
\newblock In \emph{CVPR}, 2019.

\bibitem[Kawar et~al.(2023)Kawar, Zada, Lang, Tov, Chang, Dekel, Mosseri, and Irani]{kawar2023imagic}
Bahjat Kawar, Shiran Zada, Oran Lang, Omer Tov, Huiwen Chang, Tali Dekel, Inbar Mosseri, and Michal Irani.
\newblock Imagic: Text-based real image editing with diffusion models.
\newblock In \emph{CVPR}, 2023.

\bibitem[Kwon and Ye(2023)]{kwon2022diffusion}
Gihyun Kwon and Jong~Chul Ye.
\newblock Diffusion-based image translation using disentangled style and content representation.
\newblock In \emph{ICLR}, 2023.

\bibitem[Li et~al.(2022)Li, Xu, Tian, Wang, Yan, Bi, Ye, Chen, Xu, Cao, et~al.]{li2022mplug}
Chenliang Li, Haiyang Xu, Junfeng Tian, Wei Wang, Ming Yan, Bin Bi, Jiabo Ye, Hehong Chen, Guohai Xu, Zheng Cao, et~al.
\newblock m{PLUG}: Effective and efficient vision-language learning by cross-modal skip-connections.
\newblock In \emph{EMNLP}, 2022.

\bibitem[Lin et~al.(2014)Lin, Maire, Belongie, Hays, Perona, Ramanan, Doll{\'a}r, and Zitnick]{lin2014microsoft}
Tsung-Yi Lin, Michael Maire, Serge Belongie, James Hays, Pietro Perona, Deva Ramanan, Piotr Doll{\'a}r, and C~Lawrence Zitnick.
\newblock Microsoft {COCO}: Common objects in context.
\newblock In \emph{ECCV}, 2014.

\bibitem[Lin et~al.(2024)Lin, Chen, Tsai, Jiang, and Yang]{lin2024text}
Yuanze Lin, Yi-Wen Chen, Yi-Hsuan Tsai, Lu Jiang, and Ming-Hsuan Yang.
\newblock Text-driven image editing via learnable regions.
\newblock In \emph{CVPR}, 2024.

\bibitem[Meng et~al.(2022)Meng, He, Song, Song, Wu, Zhu, and Ermon]{meng2022sdedit}
Chenlin Meng, Yutong He, Yang Song, Jiaming Song, Jiajun Wu, Jun-Yan Zhu, and Stefano Ermon.
\newblock {SDE}dit: Guided image synthesis and editing with stochastic differential equations.
\newblock In \emph{ICLR}, 2022.

\bibitem[Mokady et~al.(2023)Mokady, Hertz, Aberman, Pritch, and Cohen-Or]{mokady2023null}
Ron Mokady, Amir Hertz, Kfir Aberman, Yael Pritch, and Daniel Cohen-Or.
\newblock Null-text inversion for editing real images using guided diffusion models.
\newblock In \emph{CVPR}, 2023.

\bibitem[Nichol et~al.(2022)Nichol, Dhariwal, Ramesh, Shyam, Mishkin, McGrew, Sutskever, and Chen]{nichol2021glide}
Alex Nichol, Prafulla Dhariwal, Aditya Ramesh, Pranav Shyam, Pamela Mishkin, Bob McGrew, Ilya Sutskever, and Mark Chen.
\newblock {GLIDE}: Towards photorealistic image generation and editing with text-guided diffusion models.
\newblock In \emph{ICML}, 2022.

\bibitem[Oquab et~al.(2023)Oquab, Darcet, Moutakanni, Vo, Szafraniec, Khalidov, Fernandez, Haziza, Massa, El-Nouby, et~al.]{oquab2023dinov2}
Maxime Oquab, Timoth{\'e}e Darcet, Th{\'e}o Moutakanni, Huy Vo, Marc Szafraniec, Vasil Khalidov, Pierre Fernandez, Daniel Haziza, Francisco Massa, Alaaeldin El-Nouby, et~al.
\newblock {DINO}v2: Learning robust visual features without supervision.
\newblock In \emph{TMLR}, 2023.

\bibitem[Radford et~al.(2021)Radford, Kim, Hallacy, Ramesh, Goh, Agarwal, Sastry, Askell, Mishkin, Clark, et~al.]{radford2021clip}
Alec Radford, Jong~Wook Kim, Chris Hallacy, Aditya Ramesh, Gabriel Goh, Sandhini Agarwal, Girish Sastry, Amanda Askell, Pamela Mishkin, Jack Clark, et~al.
\newblock Learning transferable visual models from natural language supervision.
\newblock In \emph{ICML}, 2021.

\bibitem[Ramesh et~al.(2021)Ramesh, Pavlov, Goh, Gray, Voss, Radford, Chen, and Sutskever]{ramesh2021zero}
Aditya Ramesh, Mikhail Pavlov, Gabriel Goh, Scott Gray, Chelsea Voss, Alec Radford, Mark Chen, and Ilya Sutskever.
\newblock Zero-shot text-to-image generation.
\newblock In \emph{ICML}, 2021.

\bibitem[Rombach et~al.(2022)Rombach, Blattmann, Lorenz, Esser, and Ommer]{rombach2022high}
Robin Rombach, Andreas Blattmann, Dominik Lorenz, Patrick Esser, and Bj{\"o}rn Ommer.
\newblock High-resolution image synthesis with latent diffusion models.
\newblock In \emph{CVPR}, 2022.

\bibitem[Saharia et~al.(2022)Saharia, Chan, Saxena, Li, Whang, Denton, Ghasemipour, Gontijo~Lopes, Karagol~Ayan, Salimans, et~al.]{saharia2022photorealistic}
Chitwan Saharia, William Chan, Saurabh Saxena, Lala Li, Jay Whang, Emily~L Denton, Kamyar Ghasemipour, Raphael Gontijo~Lopes, Burcu Karagol~Ayan, Tim Salimans, et~al.
\newblock Photorealistic text-to-image diffusion models with deep language understanding.
\newblock \emph{NeurIPS}, 2022.

\bibitem[Shi et~al.(2020)Shi, Xu, Bui, Dernoncourt, Wen, and Xu]{shi2020ldie}
Jing Shi, Ning Xu, Trung Bui, Franck Dernoncourt, Zheng Wen, and Chenliang Xu.
\newblock A benchmark and baseline for language-driven image editing.
\newblock In \emph{ACCV}, 2020.

\bibitem[Wang et~al.(2023{\natexlab{a}})Wang, Zhang, Birsak, and Wonka]{wang2023instructedit}
Qian Wang, Biao Zhang, Michael Birsak, and Peter Wonka.
\newblock Instruct{E}dit: Improving automatic masks for diffusion-based image editing with user instructions.
\newblock \emph{arXiv:2305.18047}, 2023{\natexlab{a}}.

\bibitem[Wang et~al.(2023{\natexlab{b}})Wang, Saharia, Montgomery, Pont-Tuset, Noy, Pellegrini, Onoe, Laszlo, Fleet, Soricut, et~al.]{wang2023editbench}
Su Wang, Chitwan Saharia, Ceslee Montgomery, Jordi Pont-Tuset, Shai Noy, Stefano Pellegrini, Yasumasa Onoe, Sarah Laszlo, David~J Fleet, Radu Soricut, et~al.
\newblock {I}magen editor and {E}dit{B}ench: Advancing and evaluating text-guided image inpainting.
\newblock In \emph{CVPR}, 2023{\natexlab{b}}.

\bibitem[Wu et~al.(2024{\natexlab{a}})Wu, Fan, Qin, Gu, Zhao, and Chan]{wu2024freediff}
Wei Wu, Qingnan Fan, Shuai Qin, Hong Gu, Ruoyu Zhao, and Antoni~B Chan.
\newblock Free{D}iff: Progressive frequency truncation for image editing with diffusion models.
\newblock In \emph{ECCV}, 2024{\natexlab{a}}.

\bibitem[Wu et~al.(2019)Wu, Kirillov, Massa, Lo, and Girshick]{wu2019detectron2}
Yuxin Wu, Alexander Kirillov, Francisco Massa, Wan-Yen Lo, and Ross Girshick.
\newblock Detectron2.
\newblock \url{https://github.com/facebookresearch/detectron2}, 2019.

\bibitem[Wu et~al.(2024{\natexlab{b}})Wu, Hu, Fu, Zhou, and Li]{wu2024gpt}
Yiqi Wu, Xiaodan Hu, Ziming Fu, Siling Zhou, and Jiangong Li.
\newblock {GPT}-4o: Visual perception performance of multimodal large language models in piglet activity understanding.
\newblock \emph{arXiv:2406.09781}, 2024{\natexlab{b}}.

\bibitem[Zhang et~al.(2024)Zhang, Mo, Chen, Sun, and Su]{zhang2024magicbrush}
Kai Zhang, Lingbo Mo, Wenhu Chen, Huan Sun, and Yu Su.
\newblock Magic{B}rush: A manually annotated dataset for instruction-guided image editing.
\newblock \emph{NeurIPS}, 2024.

\bibitem[Zhang et~al.(2018)Zhang, Isola, Efros, Shechtman, and Wang]{zhang2018lpips}
Richard Zhang, Phillip Isola, Alexei~A Efros, Eli Shechtman, and Oliver Wang.
\newblock The unreasonable effectiveness of deep features as a perceptual metric.
\newblock In \emph{CVPR}, 2018.

\end{thebibliography}
\newpage
\onecolumn
\appendix

\pagenumbering{roman}
\renewcommand\thetable{\Roman{table}}
\renewcommand\thefigure{\Roman{figure}}
\setcounter{section}{0}
\setcounter{table}{0}
\setcounter{figure}{0}

\noindent\textbf{\Large Appendix}

\section{More Details on Dataset Filtering}
\label{sec:edit_type}

In order to decide the candidate objects to be potentially edited, we first check if the object is sufficiently large.
This is because the object needs to be at least distinguishable in the image and to have enough pixels to make any meaningful edits.
For this reason, we filter out objects whose bounding box size is smaller than 0.5\% of the whole image.
Also, in order to avoid a situation that a query asks to edit the unseen area of an object, we require the whole body of the target object to be shown in the image; so we eliminate objects that have bounding box adjacent to the edge of the image, regarding that those objects may have been cropped.
We further eliminate occluded objects with help from a pre-trained Vision Question and Answering (VQA) model \citep{li2022mplug}.
Given an image with an object $A$, we ask the VQA model to answer the following six yes-or-no questions, and only the objects that receive 4 or more
desired answers are kept.

\vspace{2mm}
\begin{itemize}[noitemsep]
    \item \textit{Is the $A$ hidden behind another object?}
    \item \textit{Is part of the $A$ covered by another object?} 
    \item \textit{Is the $A$ partially outside the image frame?} 
    \item \textit{Is part of the $A$ blocked by something else in the scene?}
    \item \textit{Are parts of the $A$ visible?} 
    \item \textit{Is the $A$ fully in view without anything blocking it?} 
\end{itemize}

\vspace{2mm}
\noindent
Lastly, an off-the-shelf instance segmentation model should be able to detect the objects to automate the evaluation process using our approach.
Thus, we filter out objects that have its IoU lower than 0.5 between the annotated bounding box in GQA dataset and the detected segment.

\section{More Details on Editing Query}
\label{sec:app_qgen}

\subsection{Details on Generation Rules}
\label{sec:qgen_rule_detail}

\textbf{Object-centric Queries.}
For \textsc{Object Addition}, the target object class $c_{o'}$ and its desired placement $r_{o'}$ is required.
To select feasible target object type $c_{o'}$, we conduct a statistical analysis of the relative position information from the base dataset to identify relevant objects that frequently have relationships with the anchor object $o$ in the specified positions.
To decide all feasible location of target object $o'$ relative to $o$, we utilize again the relative position information annotated in the base dataset to accurately determine a placement that accounts for depth.
Additionally, in order to make the addition feasible, we restrict the target location to some space that is unoccupied by other objects in the image based on scene graph annotation in the base dataset. We also make sure if there is large enough margin in the image to generate any additional object. These ensure that we generate a query only with a plausible target object at a feasible location.
We also make sure not to ask the model to generate any object class that already exists in the image. This is to avoid having two or more same class objects in the image that might cause confusion during the segmentation process at evaluation.

The \textsc{Object Removal} task involves removing an editable object from an image.
There is little feasibility restriction for this task, as long as the target object $o$ has been detected and belongs to one of the covered classes in $\mathcal{C}$.
While this task appears simple, removing the main object from the image can significantly impact its description.
As maintaining the original image description is crucial for editing consistency, we exclude object removal tasks when we evaluate description-based methods.

The \textsc{Object Replacement} task involves altering the core identity of an object, keeping the original position unchanged.
The replacement target class $c_{o'}$ is selected similarly to the \textsc{Object Addition} task, by choosing an object from the dataset that has a realistic and contextually appropriate relationship with all other objects in the image.
This ensures that the edited image maintains plausible object positioning.
Again, we avoid any object class that already exists in the image to prevent confusion at evaluation.

For \textsc{Object Resizing}, when the target object is too small (or too large) relative to the image size, we only generate a query to make the object larger (smaller).

For \textsc{Object Attribute Change}, it is crucial to preserve core identity of the object $o$ while editing the specific characteristics mentioned in the query.
We similarly curate a list of attribute words from the base dataset.
As we do not know what kind of attributes an object can have other than the annotated ones, we restrict changes only to the annotated attributes.
In addition, when choosing a replacement attribute, the option set is structured slightly different for each attribute type (color, state, material, action).
A typical option set of certain object class will look like as follows:

\vspace{2mm}
\begin{itemize}[leftmargin=0.7cm]
    \item Color: `black', `green', `brown', ...
    \item State: `wet', `dry', `new', `old', `rusted', ...
    \item Material: `wood', `metal', ...
    \item Action: `standing', `running', ...
\end{itemize}

\vspace{2mm}
\noindent
If we are to change an attribute other than those in the State group, we simply choose a replacement from the same attribute group (\textit{i.e.}, `green' from Color to replace `black').
On the other hand, attributes in the State group usually cannot be replaced with any random state attributes.
For example, a state `wet' should not be replaced with `rusted'.
Thus, we define feasible alternatives for each state individually, based on the statistics from the base dataset annotations (\textit{i.e.}, couple `wet' and `dry').

\vspace{0.1cm} \noindent
\textbf{Non-object-centric Queries.} The full list of elements on 
\textsc{Background Change} and \textsc{Style Change}
options are as follows:

\vspace{2mm}
\noindent
$\mathcal{B}$ = \{`beach',
    `pine forest',
    `urban city',
    `desert',
    `snow field',
    `country side farm',
    `tropical jungle',
    `vineyard',
    `lake side',
    `mountain top',
    `living room',
    `cave',
    `art gallery',
    `ancient ruins',
    `space station',
    `grass field',
    `train station',
    `library',
    `restaurant',
    `airport',
    `hospital',
    `gym',
    `zoo',
    `aquarium',
    `museum',
    `concert hall',
    `stadium'\},

\vspace{2mm}
\noindent
$\mathcal{S}$ = \{`watercolor painting',
    `Van Gogh art',
    `oil painting',
    `cartoon',
    `gray scale',
    `pencil sketch',
    `mosaic art',
    `pop art',
    `graffiti art',
    `ancient Egyptian art'\}.

\vspace{2mm}
\noindent
To prevent editing the background to the one that is already the original background, we select target background ($b$) from half of the options with the lowest CLIP alignment with the original image ($I_0$).

\subsection{Detailed Flow for Edit Description and Instruction Generation}
\label{sec:capgen_prompt}

For description-based editing, we first generate a caption for the original image ($C_0$).
As the base dataset does not provide a natural language description of the images, we use Llama3~\cite{dubey2024llama} to generate $C_0$.
To generate a caption focusing on the editable objects, we directly demand it to use the exact object names and repeat generation until the caption contains the exact names.
We also require the model to answer within 60 words to avoid long captions beyond the maximum input length of the editing models. 

Then, we refine the generated captions to insert available attributes of the editable objects using GPT4~\cite{wu2024gpt}.
Specifically, we ask it to first remove any attribute descriptions of the object in the caption to prevent collision, and then to insert desired attributes of the objects using the exact words we provide;
\textit{e.g.}, ``a photo of a crimson cat'' (first draft) $\rightarrow$ ``a photo of a cat'' (original attributes removed) $\rightarrow$ ``a photo of a wet red cat'' (desired attributes inserted, $C_0$).

We generate a target caption ($C_e$) specific to each task.
For \textsc{Object Replacement} queries, we ask GPT to find and replace the object with the desired one.
Then, we ask it to correct any grammatical errors (\textit{e.g.}, ``a photo of a person with his cat'' ($C_0$) $\rightarrow$ ``a photo of a phone with his cat'' (original object replaced) $\rightarrow$ ``a photo of a phone with its cat'' (grammar fixed, $C_e$)).
For \textsc{Object Addition}, \textsc{Object Resizing}, and \textsc{Background Change}, we ask GPT to add a desired information to the original caption.
For \textsc{Object Attribute Change}, we simply replace the original attribute word with a desired one.
For \textsc{Style Change} queries, we attach a short prefix sentence describing about the style of the image, and ask GPT to combine the two sentences into one; \textit{e.g.}, ``A cat is sitting on a couch.'' ($C_0$) $\rightarrow$ ``An oil painting art. The art contains following: A cat is sitting on a couch.'' (prefix attached) $\rightarrow$ ``An oil painting art of cat sitting on a couch.'' (Combined sentence, $C_e$).

To finalize, we test run generated original and target captions ($I_0$ and $I_e$), through the text encoders of the editing models to check their compatibility. We manually fix captions with any problems oddness. The full list of LLM prompts used to generate captions are provided in \cref{sec:capgen_prompt_list}.

For instruction-based editing, we manually craft the instruction templates for each edit type,
\textit{e.g.}, ``change the color of the $\{object\}$ from $\{a\}$ to $\{b\}$'', and fill them with desired words to finalize.
The full list of templates are provided in \cref{sec:inst_temp}.

\subsection{Full List of LLM Prompts and Instruction Templates}
\label{sec:promptNtemp}

Below, we provide the full list of LLM prompts and templates we used to generate each caption and instructions.

\subsubsection{LLM Prompts}
\label{sec:capgen_prompt_list}
\begin{figure*}[hb]
    \vspace{-0.5cm}
    \includegraphics[width=\linewidth]{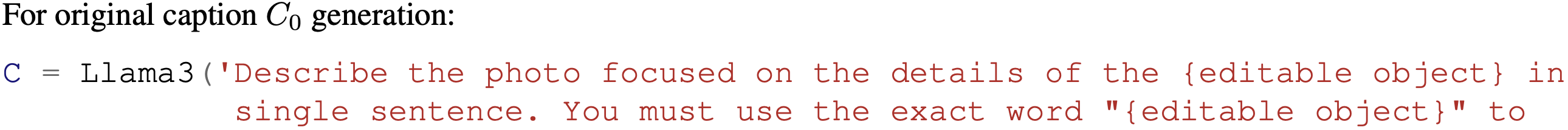}
    \vspace{-1cm}
\end{figure*}

\begin{figure*}[ht]
    \vspace{-1cm}
    \includegraphics[width=\linewidth]{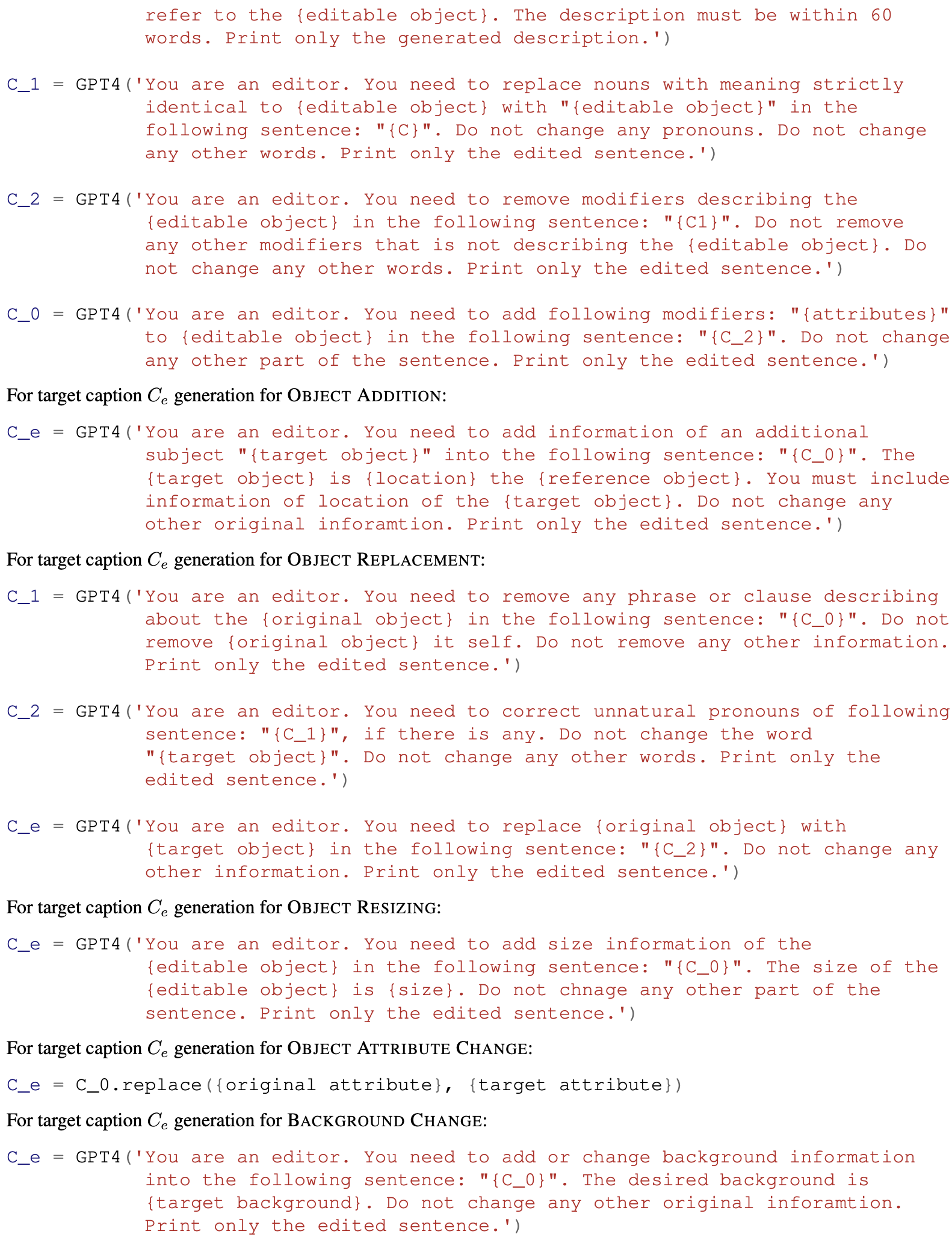}
    \vspace{-1cm}
\end{figure*}

\begin{figure*}[ht]
    \vspace{-1cm}
    \includegraphics[width=\linewidth]{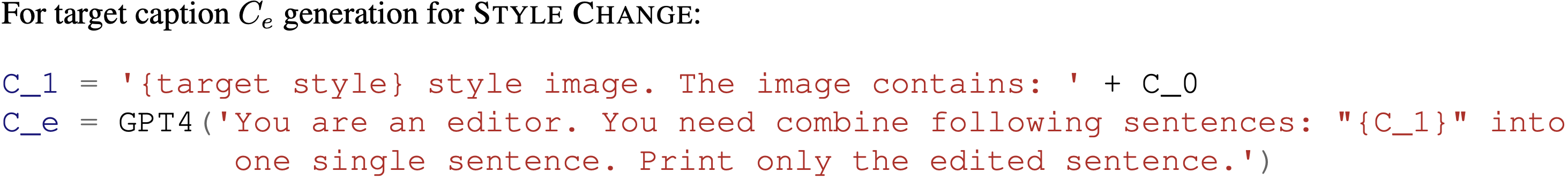}
    \vspace{-1cm}
\end{figure*}

\subsubsection{List of Caption Generation Prompts}
\label{sec:inst_temp}
For instruction $C$ generation for \textsc{Object Addition}:

\begin{figure*}[h]
    \vspace{-0.3cm}
    \includegraphics[width=0.88\linewidth]{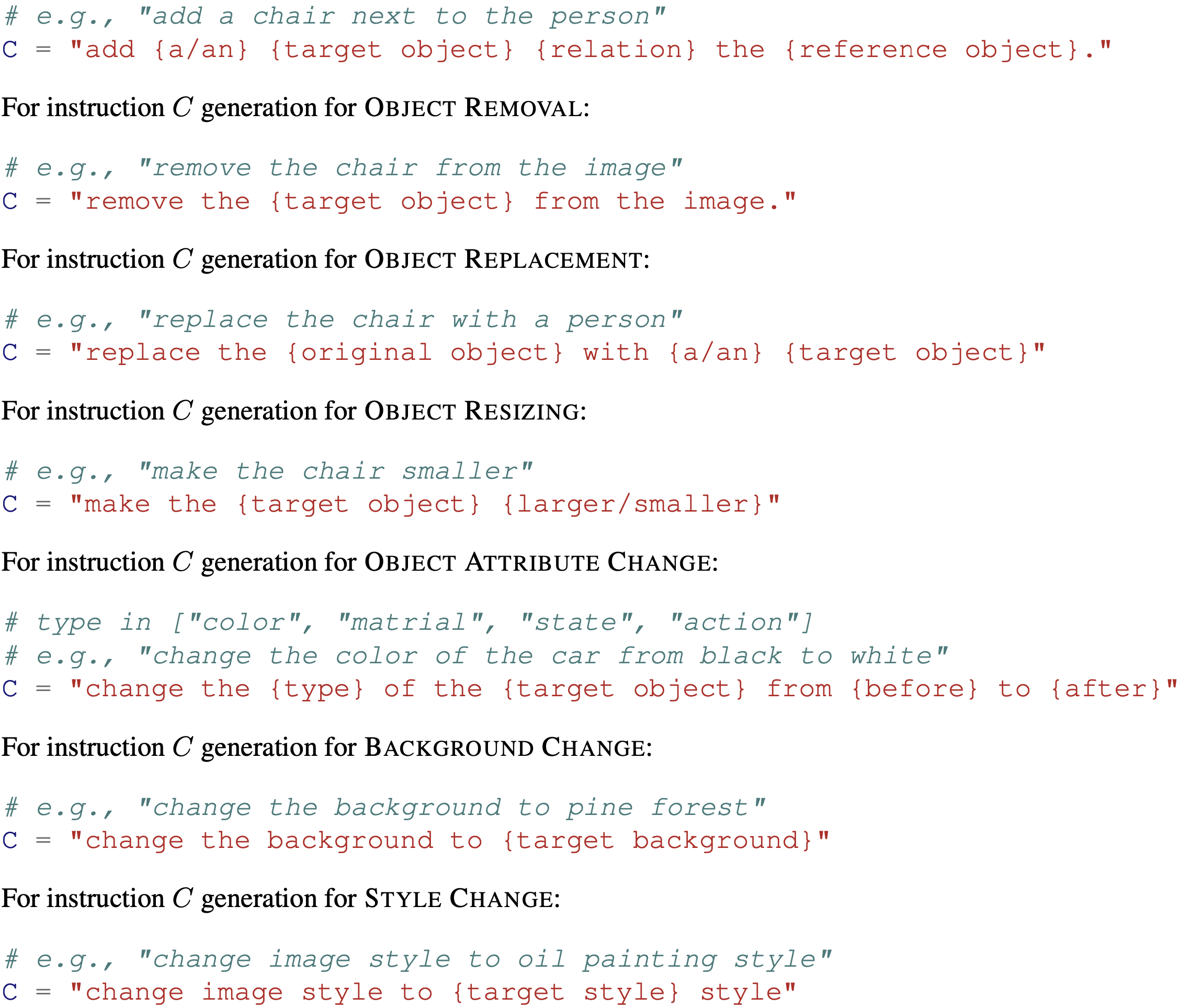}
    \vspace{-0.5cm}
\end{figure*}

\subsection{Balanced Query Types and Options}
\label{sec:balanced_query}

\begin{figure}[h]
    \centering
    \begin{subfigure}{0.48\textwidth}
        \centering
        \includegraphics[height=6cm]{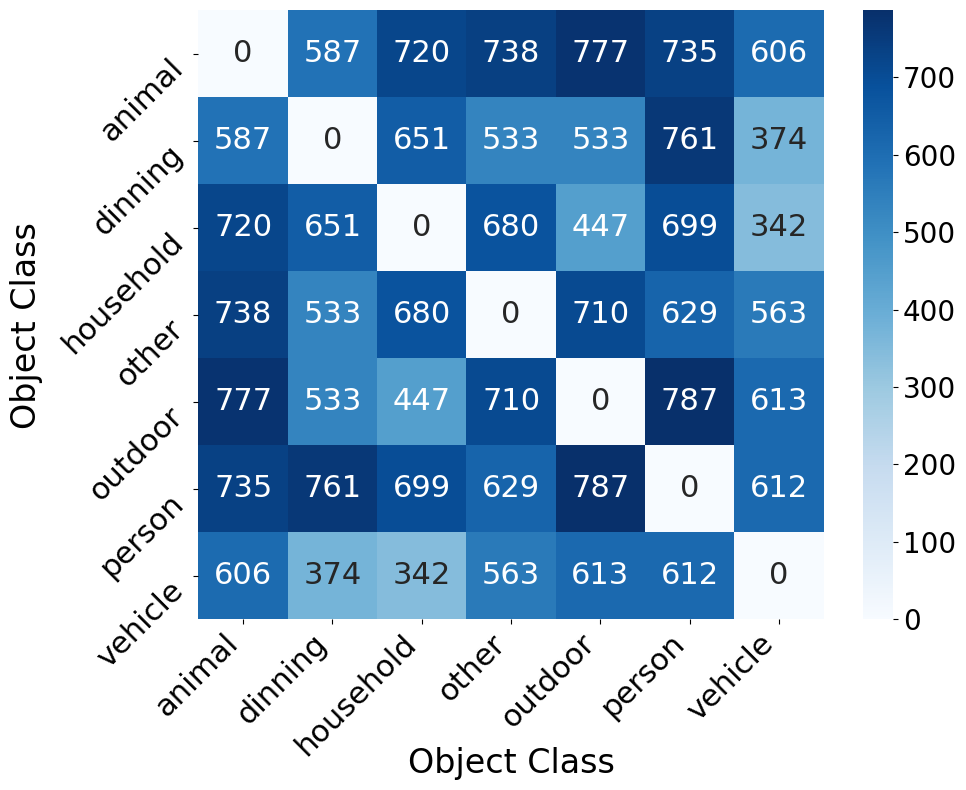}
        \caption{Pairwise object class distribution}
        \label{fig:classpair_heatmap}
    \end{subfigure}
    \begin{subfigure}{0.4\textwidth}
        \centering
        \begin{subfigure}{\linewidth}
            \centering
            \includegraphics[height=4.8cm]{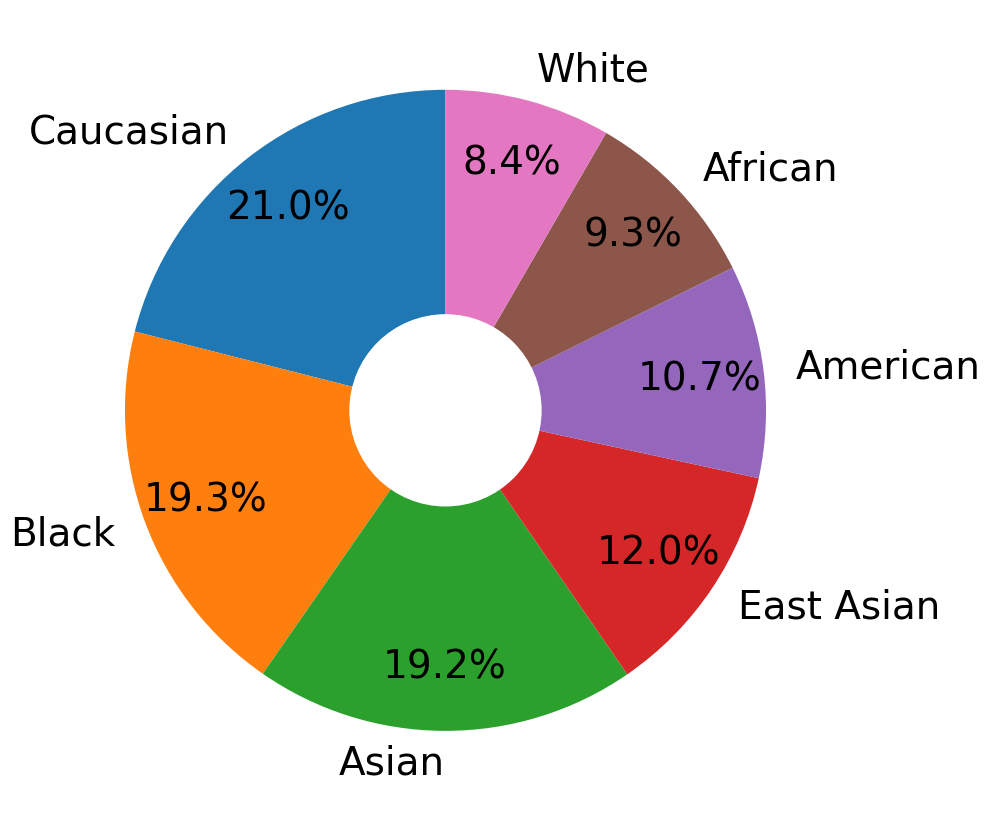}
            \caption{Race distribution}
            \label{fig:race_distribution}
        \end{subfigure}
        \vspace{0.2cm}
        \begin{subfigure}{\linewidth}
            \centering
            \includegraphics[height=1cm]{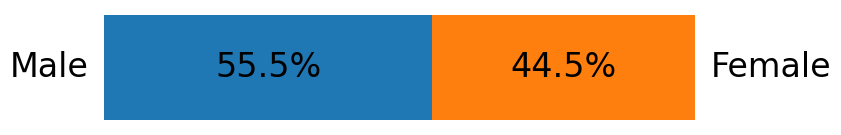}
            \caption{Gender distribution}
            \label{fig:gender_distribution}
        \end{subfigure}
    \end{subfigure}
    \caption{\textbf{Data distribution in editing queries.} (a) Pairwise object class counts (b) Race distribution within the \textit{person} class. (c) Gender distribution within the \textit{person} class.}
\end{figure}

To mitigate bias in edit queries, we extracted valid relationships from large-scale dataset while ensuring a balanced distribution of object types and relations.
The pairwise object type counts in \cref{fig:classpair_heatmap} and \cref{fig:task_dist} indicate that our dataset and queries well balanced throughout object classes.
Additionally, we assess potential gender and racial biases.
Since direct labels are unavailable, we estimate the most probable class by computing CLIP similarity scores between 'person' instances and various gender- and race-related terms.
The results presented in \cref{fig:race_distribution} and \cref{fig:gender_distribution} suggest that our dataset and queries retains balance between gender and racial types.

\section{More Details on Evaluation Pipeline}
\label{sec:eval_detail}

\subsection{Task-specific Evaluation Workflow}
\label{sec:eval_workflow}
We provide detailed evaluation workflow described in \cref{fig:eval_workflow} and \cref{fig:eval_rm}.

\subsubsection{\textsc{Object Addition}}
\label{sec:app:eval_add}

For a satisfactory \textsc{Object Addition}, the newly generated object $o'$ should be 1) within the desired class ($c_{o'}$) and 2) located at the desired location ($r_{o'}$), while 3) retaining the background.
The first point is quantified by $\sigma^{OF}_\text{det}$ and $\sigma^{OF}_\text{clip,c}$, where $\sigma^{OF}_\text{clip,c}$ denotes $\sigma^{OF}_\text{clip}$ computed between the generated object segment $M_o$ and the object class name $o$ (\textit{e.g.}, CLIP alignment between segmented pizza and word ``pizza'' in \cref{fig:eval_add}).
To quantify the second point, we first square crop the edited image ($I_e$) to be minimal in size while enclosing both the reference object ($o$) and the generated one $o'$ (\textit{e.g.}, cropped image containing pizza and microwave in \cref{fig:eval_add}).
Then, we compute the CLIP alignment score between the cropped image and segment of instruction ($C$) that describes $r_{o'}$ (\textit{e.g.}, ``pizza under microwave'' in \cref{fig:eval_add}). We define this as $\sigma^{OF}_\text{clip,r}$.
These two points make up $\sigma^{OF}$ as follows:
\begin{equation}
    \sigma^{OF} = \frac{w^{OF}_\text{clip,c} \sigma^{OF}_\text{clip,c} + w^{OF}_\text{clip,r} \sigma^{OF}_\text{clip,r} + w^{OF}_\text{det} \sigma^{OF}_\text{det}}{w^{OF}_\text{clip,c} + w^{OF}_\text{clip,r} + w^{OF}_\text{det}}.
\end{equation}

The third point is evaluated with $\sigma^{BC}$. The original and edited background for $\sigma^{BC}$ are obtained by masking the $o'$ region from both original ($I_0$) and edited ($I_e$) image.
We mask out $o'$ region from $I_0$ as well (although it may not contain any meaningful features) to remove any discrepancy due to the masked region in the edited background.
When $o'$ is not detected from $I_e$, $\sigma^{OF}$ is set to 0, and $\sigma^{BC}$ is computed with the entire $I_0$ and $I_e$.
To sum up, the \textsc{Object Addition} task is evaluated with two scores, the Object Fidelity $\sigma^{OF}$ and Background Consistency $\sigma^{BC}$.

\subsubsection{\textsc{Object Removal}}
For \textsc{Object Removal} task to be successful, target object $o$ should be 1) undetected from the edited image ($I_e$), while 2) retaining the background.
The first point is quantified by $1-\sigma^{OF}_\text{det}$ and $1-\sigma^{OF}_\text{clip,c}$, explained in \cref{sec:app:eval_add}. We modify the scores to $1-\sigma^{OF}_\text{*}$ format to suit the task's character (the more unlike the edited object is to the original object class, the better the edit is). 
These two scores make up $\sigma^{OF}$ as follows:
\begin{equation}
    \sigma^{OF} = \frac{w^{OF}_\text{clip,c} (1-\sigma^{OF}_\text{clip,c}) + w^{OF}_\text{det} (1-\sigma^{OF}_\text{det})}{w^{OF}_\text{clip,c} + w^{OF}_\text{det}}.
\end{equation}

The second point is evaluated with $\sigma^{BC}$. The original and edited background are obtained by masking the union of original and edited region of the object $o$ from $I_0$ and $I_e$, respectively.
We mask out the same region form $I_0$ and $I_e$ for the same reason in \cref{sec:app:eval_add}.
When $o$ is not detected from $I_e$, $\sigma^{OF}$ is set to 1, and $\sigma^{BC}$ is computed with backgrounds with only the original object regions masked.
To sum up, the \textsc{Object Removal} task is evaluated with two scores, the Object Fidelity $\sigma^{OF}$ and Background Consistency $\sigma^{BC}$.

\subsubsection{\textsc{Object Replacement}}

For a successful \textsc{Object Replacement}, the target object ($o$) should be replaced with a new object ($o'$) 1) within desired class ($c_{o'}$), while 2) retaining the location of $o$, 3) without affecting the background.
The first point is quantified by $\sigma^{OF}_\text{det}$ and $\sigma^{OF}_\text{clip,c}$ explained in \cref{sec:app:eval_add}, which makes up $\sigma^{OF}$ as:
\begin{equation}
    \sigma^{OF} = \frac{w^{OF}_\text{clip,c} \sigma^{OF}_\text{clip,c} + w^{OF}_\text{det} \sigma^{OF}_\text{det}}{w^{OF}_\text{clip,c} + w^{OF}_\text{det}}.
\end{equation}
The second point is quantified by $\sigma^{OC}_\text{pos}$, and solely makes up $\sigma^{OC} = \sigma^{OC}_\text{pos}$.
The third point is again quantified with $\sigma^{BC}$. The original and edited backgrounds are obtained by masking the union of $o$ and $o'$ region from $I_0$ and $I_e$.
We mask out the same region form $I_0$ and $I_e$ for the same reason in \cref{sec:app:eval_add}.
When $o'$ is not detected from $I_e$, $\sigma^{OF}$ and $\sigma^{OC}$ is set to 0, and $\sigma^{BC}$ is computed with backgrounds with only the original object regions masked.
In sum, the \textsc{Object Replacement} score is composed of three scores, Object Fidelity $\sigma^{OF}$, Object Consistency $\sigma^{OC}$ and Background Consistency $\sigma^{BC}$.

\subsubsection{\textsc{Object Resizing}}
For \textsc{Object Resizing}, we define four conditions to satisfy. The target object $o$ should be 1) correctly resized, retaining its 2) shape and 3) position, 4) without affecting the background.
The first point is quantified by $\sigma^{OF}_\text{size}$, which solely makes up $\sigma^{OF} = \sigma^{OF}_\text{size}$.
The second point is quantified by $\sigma^{OC}_\text{lpips}$, $\sigma^{OC}_\text{dino}$, and $\sigma^{OC}_{\ell 2}$, and the third point measured by $\sigma^{OC}_\text{pos}$.
These make up $\sigma^{OC}$ as follows:
\begin{equation}
    \sigma^{OC} = \frac{w^{OC}_\text{lpips} \sigma^{OC}_\text{lpips} + w^{OC}_\text{dino} \sigma^{OC}_\text{dino} + w^{OC}_{\ell 2} \sigma^{OC}_{\ell 2} + w^{OC}_\text{pos} \sigma^{OC}_\text{pos}}{w^{OC}_\text{lpips} + w^{OC}_\text{dino} + w^{OC}_{\ell 2} + w^{OC}_\text{pos}}.
\end{equation}
The fourth point is evaluated again with $\sigma^{BC}$.
The original and edited background are obtained by masking the union of original and resized region of the object $o$ from $I_0$ and $I_e$, respectively.
We mask out the same region form $I_0$ and $I_e$ for the same reason as in \cref{sec:app:eval_add}.
When $o$ is not detected from $I_e$, $\sigma^{OF}$ and $\sigma^{OC}$ is set to 0, and $\sigma^{BC}$ is computed with backgrounds with only the original object regions masked.
In sum, the \textsc{Object Resizing} task is measured with three scores, Object Fidelity $\sigma^{OF}$, Object Consistency $\sigma^{OC}$ and Background Consistency $\sigma^{BC}$.

\subsubsection{\textsc{Object Attribute Change}}
\label{sec:app:eval_attr}

We define five points to measure for the \textsc{Object Attribute Change} task: 1) whether object's ($o$) attribute ($a_i$) is changed to the desired attribute ($a_j$), 2) retaining fundamental morphological characteristics, 3) position and 4) size of $o$, 5) without affecting the background.
The first point is quantified by $\sigma^{OF}_\text{clip,a}$, which is $\sigma^{OF}_\text{clip}$ computed between the edited object segment $M_o$ and the word of $o$ and $a$ combined (\textit{e.g.}, ``cream motorcycle '' in \cref{fig:eval_attr}).
$\sigma^{OF}_\text{clip,a}$ solely makes up $\sigma^{OF} = \sigma^{OF}_\text{clip,a}$

To quantify the second point, we dull out the details of object segments from $I_0$ and $I_e$ by degradation (gray scaling and down-scaling) and Canny-edge detection \citep{canny1986computational}.
This is to remove any inconsistencies due to the attribute edit and remain only the fundamental morphological characters of the object.
Then, we compute $\sigma^{OC}_\text{deg}$ and $\sigma^{OC}_\text{edge}$, which are $\sigma^{OC}$ in \cref{eq:score_OC} computed with the degraded object segment and its detected edges, respectively.
The third and fourth points are quantified with $\sigma^{OC}_\text{pos}$ and $\sigma^{OC}_\text{size}$. These three key-points make up $\sigma^{OC}$ as:
\begin{align}
    \sigma^{OC} &= w^{OC}_\text{deg} \sigma^{OC}_\text{deg} + w^{OC}_\text{edge} \sigma^{OC}_\text{edge} + w'^{OC}_\text{pos} \sigma^{OC}_\text{pos} + w'^{OC}_\text{size} \sigma^{OC}_\text{size}, \\
    \sigma^{OC}_\text{deg} &= w'^{OC}_\text{lpips}\sigma^{OC}_\text{lpips,deg} + w'^{OC}_\text{dino} \sigma^{OC}_\text{dino,deg} + w'^{OC}_{\ell 2} \sigma^{OC}_{\ell 2,\text{deg}}, \\
    \sigma^{OC}_\text{edge} &= w'^{OC}_\text{lpips}\sigma^{OC}_\text{lpips,edge} + w'^{OC}_\text{dino} \sigma^{OC}_\text{dino,edge} + w'^{OC}_{\ell 2} \sigma^{OC}_{\ell 2,\text{edge}}.
\end{align}

\noindent
The weights with apostrophes ($w'^{OC}_\text{*}$) are optimized only for \textsc{Object Attribute Change}, independently from the other weights from other tasks.
The fifth point is measured with $\sigma^{BC}$. 
The original and edited backgrounds are obtained by masking the union of the original and the edited object region from $I_0$ and $I_e$.
We mask out the same regions form $I_0$ and $I_e$ for the same reason as in \cref{sec:app:eval_add}.
When any foreground object $o$ is not detected from $I_e$, $\sigma^{OF}$ and $\sigma^{OC}$ is set to 0, and $\sigma^{BC}$ is computed with backgrounds with only the original object regions masked.
In sum, the \textsc{Object Attribute Change} is evaluated with three scores, Object Fidelity $\sigma^{OF}$, Object Consistency $\sigma^{OC}$, and Background Consistency $\sigma^{BC}$.

\subsubsection{\textsc{Background Change}}

We define three key points to satisfy for a good \textsc{Background Change} editing. 1) The background should be changed to the desired background ($b$), while 2) retaining all the foreground objects.
The first point is quantified by $\sigma^{BF}$.
To obtain the original and edited background, we mask out the union of all the foreground objects detected in $I_0$ and $I_e$, where we indicate the foreground objects as every editable object in the image, since those are what should be unchanged from $C_0$ to $C_e$.

The second point is measured by the average of $\sigma^{OC}_{o \in \mathcal{O}}$, where $\sigma^{OC}_{o}$ is $\sigma^{OC}$ in \cref{eq:score_OC} of an object $o$ in set of all foreground objects $\mathcal{O}$. $\sigma^{OC}$ can be formulated as:
\begin{equation}
    \sigma^{OC} = \frac{1}{|\mathcal{O}|} \sum_{o \in \mathcal{O}}(w^{OC}_\text{lpips} \sigma^{OC}_\text{lpips,$o$} + w^{OC}_\text{dino} \sigma^{OC}_\text{dino,$o$} + w^{OC}_{\ell 2} \sigma^{OC}_{\ell 2,o} + w^{OC}_\text{pos} \sigma^{OC}_\text{pos,$o$} + w^{OC}_\text{size} \sigma^{OC}_\text{size,$o$}).
\end{equation}
When any foreground object $o$ is not detected from $I_e$, $\sigma^{OC}_o$ is set to 0, and $\sigma^{BC}$ is computed without the mask of $o$.
In sum, the \textsc{Background Change} task is evaluated by two scores, Background Fidelity $\sigma^{BF}$ and Object Consistency $\sigma^{OC}$.

\subsubsection{\textsc{Style Change}}
\label{sec:app:style_chg}
Two main key-points that defines a good \textsc{Style Change} is whether the style of the image ($I_0$) is changed 1) to desired style ($s$), while 2) retaining fundamental morphological character of the image. The first point is quantified by $\sigma^{BF}$. Here, we use whole image $I_0$ and $I_e$ for backgrounds.
The second point is quantified by $\sigma^{BC}_\text{deg}$ and $\sigma^{BC}_\text{deg}$, which are $\sigma^{BC}$ computed with degraded and edged (explained in \cref{sec:app:eval_attr}) $I_0$ and $I_e$ respectively. $\sigma^{BC}$ for \textsc{Style Change} is formulated as:

\begin{align}
    \sigma^{BC} &= w^{BC}_\text{deg} \sigma^{BC}_\text{deg} + w^{BC}_\text{edge} \sigma^{BC}_\text{edge}, \\
    \sigma^{BC}_\text{deg} &= w'^{BC}_\text{lpips}\sigma^{BC}_\text{lpips,deg} + w'^{BC}_\text{dino} \sigma^{OC}_\text{dino,deg} + w'^{BC}_{\ell 2} \sigma^{BC}_{\ell 2,\text{deg}}, \\
    \sigma^{BC}_\text{edge} &= w'^{BC}_\text{lpips}\sigma^{BC}_\text{lpips,edge} + w'^{BC}_\text{dino} \sigma^{BC}_\text{dino,edge} + w'^{BC}_{\ell 2} \sigma^{BC}_{\ell 2,\text{edge}}.
\end{align}

\noindent
Scores with subscript ``deg'' and ``edge'' are corresponding scores computed with degraded and edged object segments respectively and weights with apostrophe are weights optimized to \textsc{Style Change} evaluation independently from the other weights from other task evaluations. In sum, \textsc{Style Change} edited image is evaluated by two scores, background fidelity $\sigma^{BF}$ and background consistency $\sigma^{BC}$.

\subsection{Evaluation Metrics}
\label{sec:metric_detail}

\begin{figure}
\centering
\begin{subfigure}{0.35\linewidth}
    \centering
    \includegraphics[width=\textwidth]{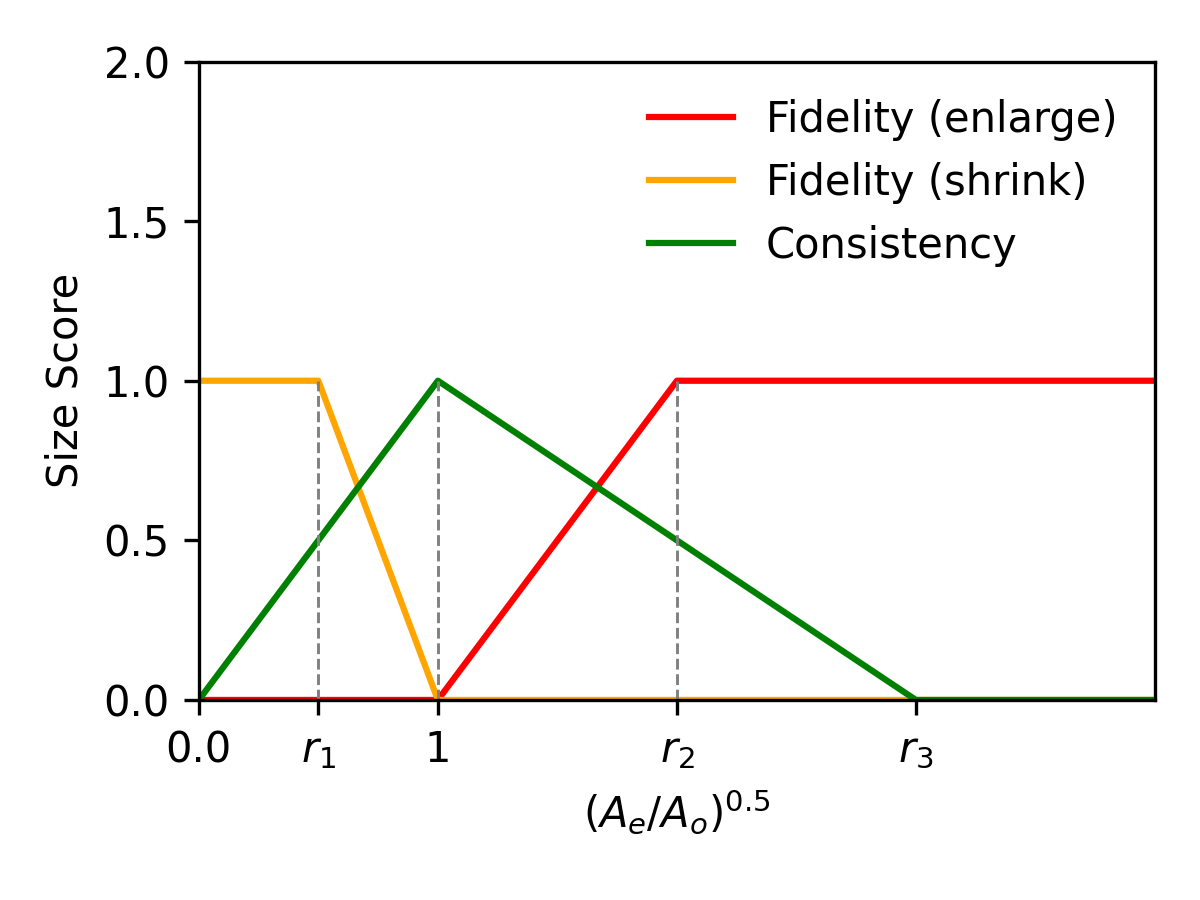}
    \caption{Size score}
    \label{fig:size_score}
\end{subfigure}
\hspace{1cm}
\begin{subfigure}{0.35\linewidth}
    \centering
    \includegraphics[width=\textwidth]{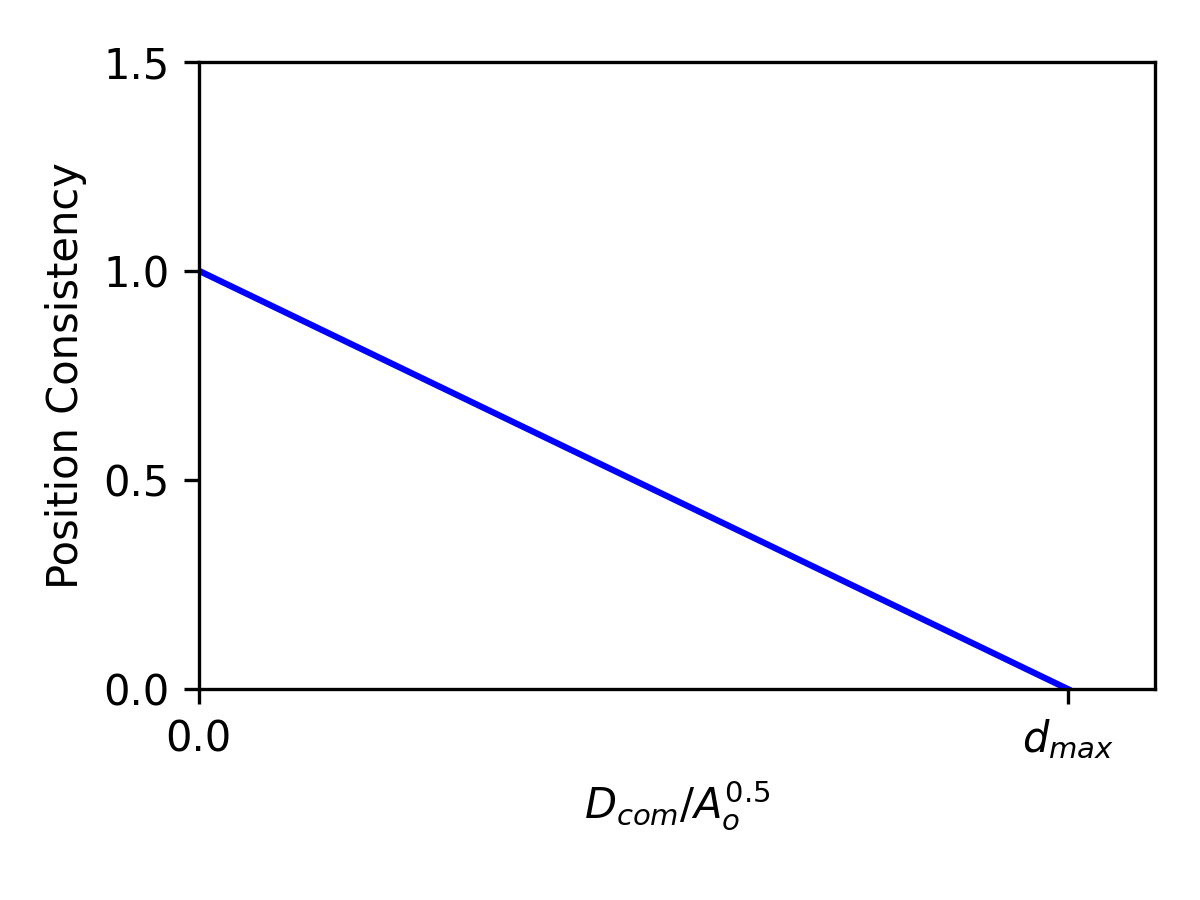}
    \caption{Position score}
    \label{fig:pos_score}
\end{subfigure}
\caption{\textbf{Our (a) size scores and (b) a position score} plotted for relative size and position changes, respectively.}
\vspace{-2mm}
\label{fig:sizepos_score}
\end{figure}

\cref{fig:sizepos_score} describes our mathematic size score and position consistency score. 
For the size score, we first calculate the relative size change $(A_e/A_0)^{0.5}$, where $A_0$ is area of the object mask in the original image, and $A_e$ is the area of the object mask in edited image.
When we calculate the size score for edit fidelity, we give full score for changes greater than thresholds ($r_1$ and $r_2$ in \cref{fig:size_score}) that are empirically set to be the boundary where it shows noticeable size change. The size score is zero for changes opposite to the intended direction, and linearly scaled results are in between.
When we compute the size score for consistency, we give the full score (1.0) for perfect size preservation. The size score is linearly scaled down to 0 as the relative size change changes from 1 to 0 and maximum possible change rate ($r_3$ in \cref{fig:size_score}), which is defined as $((H\times W)/A_0)^{0.5}$, where $H$ and $W$ indicate the height and width of the image, respectively.
Position consistency is linearly scaled down from 1 to 0 as deviation of the object mask's center of mass relative to square root of original mask size increase from 0 to maximum possible value; that is, $= \sqrt{H^2 + W^2}/A_0^{0.5}$.

\subsection{Fitting with Human Evaluation}
\label{sec:app_opt}

\begin{figure}[ht]
  \centering
   \includegraphics[width=0.6\textwidth]{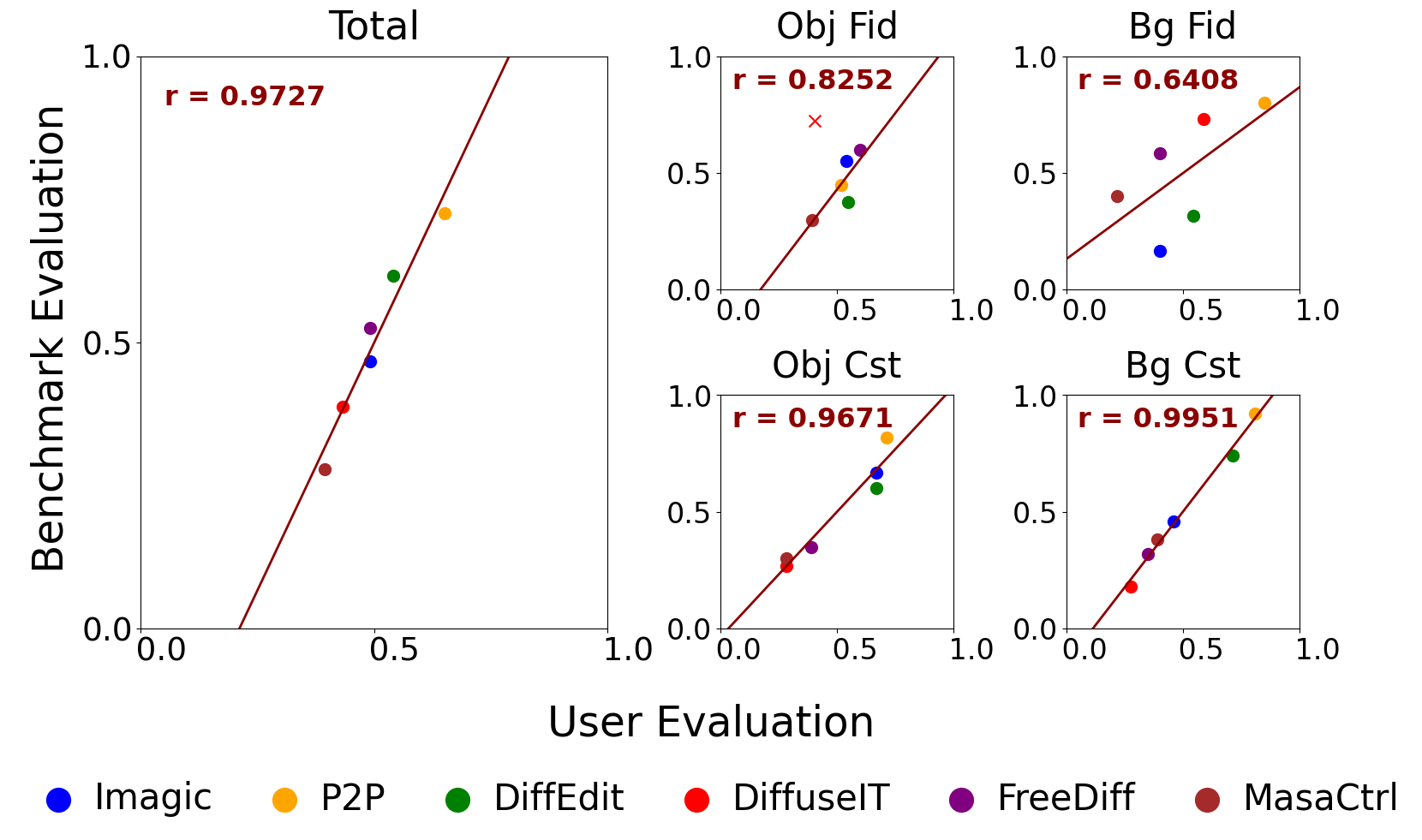}
   \caption{
   \textbf{Relation between winning rates by users and \ours, measured on the user study training set.}
   The least square linear fit (red line) and Pearson’s correlation coefficient ($r$) are reported. (See \cref{fig:user_alignment_test} for the results on the test set.)}
   \label{fig:user_alignment_train}
\end{figure}

\begin{table}
    \centering \small
    \begin{tabular}{cccccc}
        \toprule
        Correlation & \parbox{2cm}{\centering Object\\Fidelity} & \parbox{2cm}{\centering Background\\Fidelity} & \parbox{2cm}{\centering Object\\Consistency} & \parbox{2cm}{\centering Background\\Consistency} & \parbox{2cm}{\centering Total\\Score} \\
        \midrule
        $\rho$ & 0.7000 & 0.6377 & 0.9710 & 1.0000 & 1.0000 \\
        $\tau$ & 0.6000 & 0.5520 & 0.9309 & 1.0000 & 1.0000 \\
        \bottomrule
    \end{tabular}
    \caption{\textbf{Correlation coefficients between model winning rates from user study and our metrics on the user study training set.} (See \cref{tab:corr_test} for the result on the test set.)}
    \label{tab:corr_train}
\end{table}

\cref{fig:user_alignment_train} plots the same figure as \cref{fig:user_alignment_test}  but for the \textit{training} results for our benchmark weight optimization. 
Each figure shows that our optimization process is close to ideal.
However, the two fidelity scores are fitted slightly sub-optimally.
This is because both object size fidelity and background fidelity are based on a single metric.
This makes it impossible to attempt any weight optimization and restricts adjustability of our benchmark.
This indirectly proves again the importance of diversity in scoring metrics to ensure robustness of the benchmark. 
The other two correlation coefficients, Spearman's $\rho$ and Kendall's $\tau$, are also presented in \cref{tab:corr_train}.

\begin{figure}
\centering

\begin{subfigure}{\linewidth}
  \centering
  \includegraphics[width=0.8\textwidth]{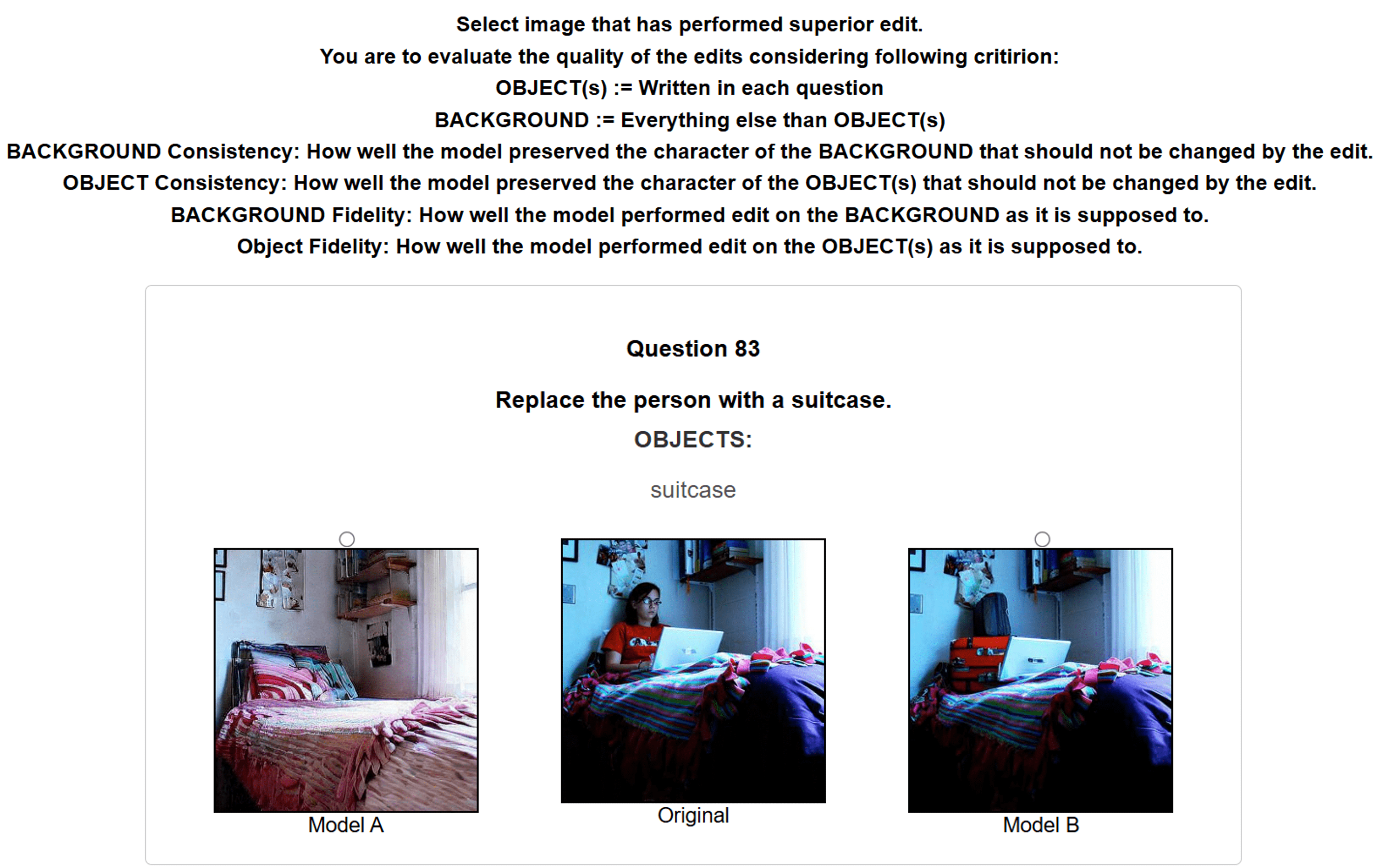}
  \caption{Total Score (Overall Edit Quality)}
  \label{fig:user_study_ex_tot}
\end{subfigure}
\vspace{2mm}

\begin{subfigure}{0.49\linewidth}
  \centering
  \includegraphics[width=\textwidth]{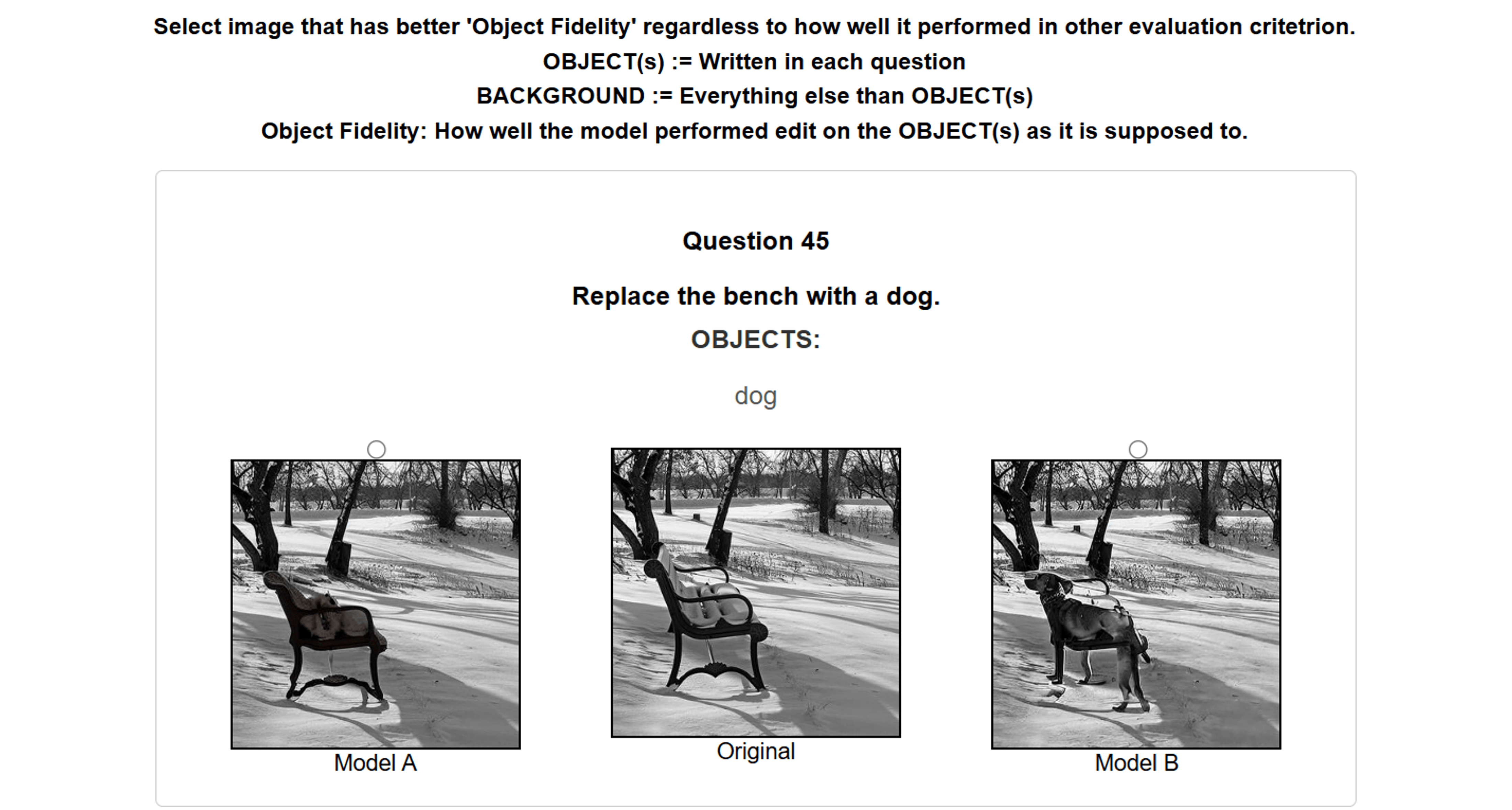}
  \caption{Object Fidelity}
  \label{fig:user_study_ex_of}
\end{subfigure}
\hfill
\begin{subfigure}{0.49\linewidth}
  \centering
  \includegraphics[width=\textwidth]{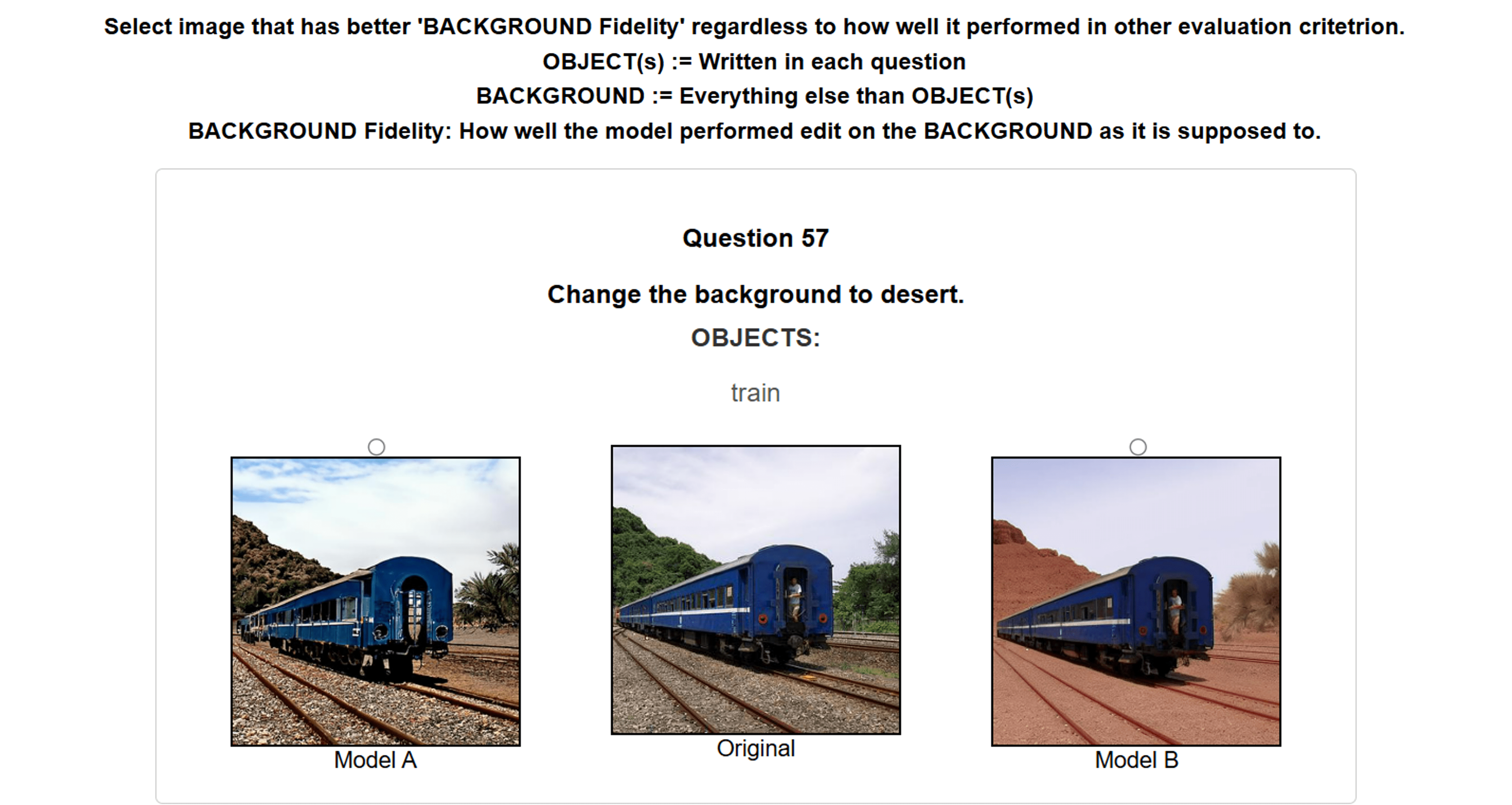}
  \caption{Background Fidelity}
  \label{fig:user_study_bf}
\end{subfigure}
\vspace{3mm}

\begin{subfigure}{0.49\linewidth}
  \centering
  \includegraphics[width=\textwidth]{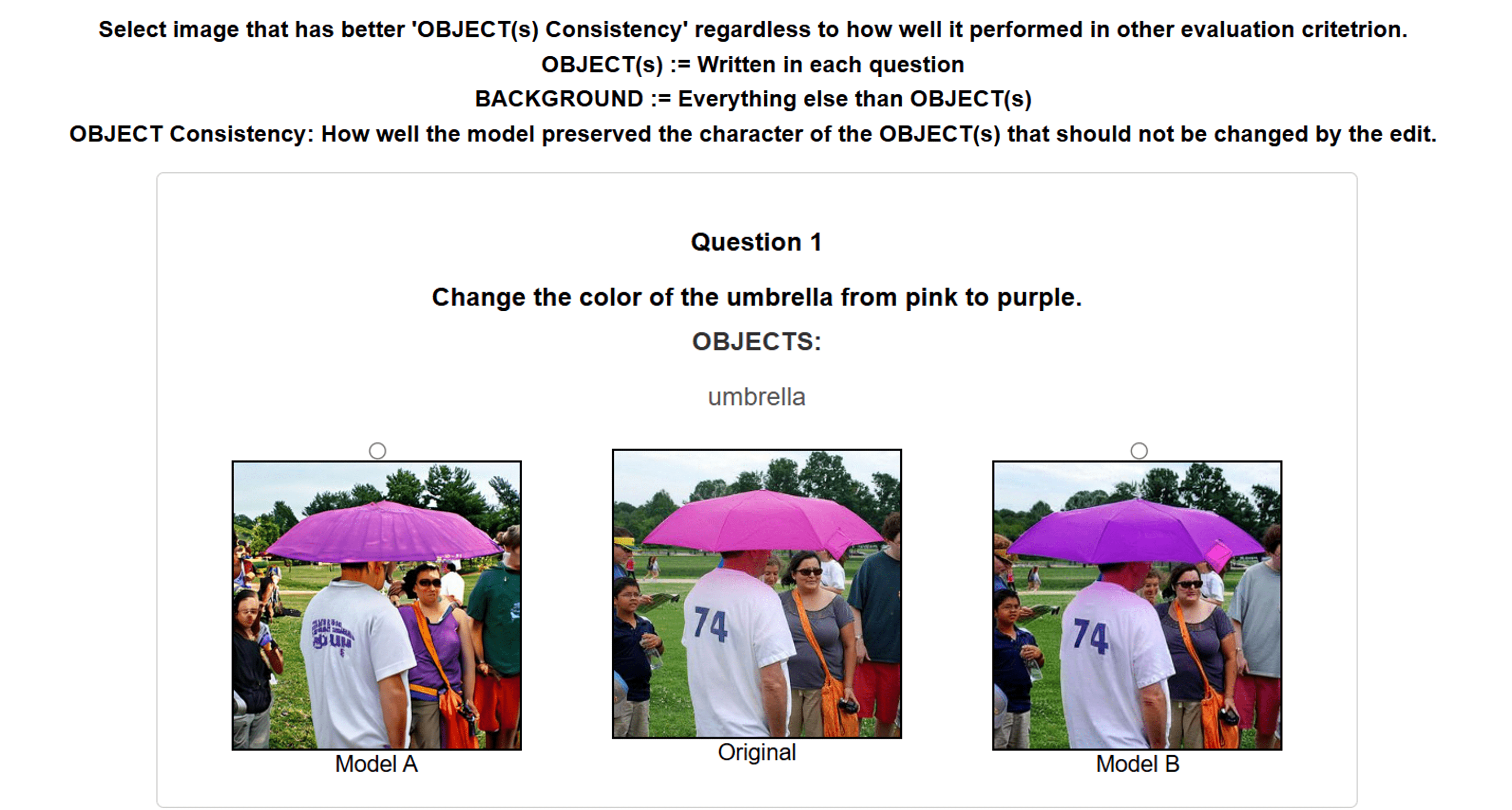}
  \caption{Object Consistency}
  \label{fig:user_study_ex_oc}
\end{subfigure}
\hfill
\begin{subfigure}{0.49\linewidth}
  \centering
  \includegraphics[width=\textwidth]{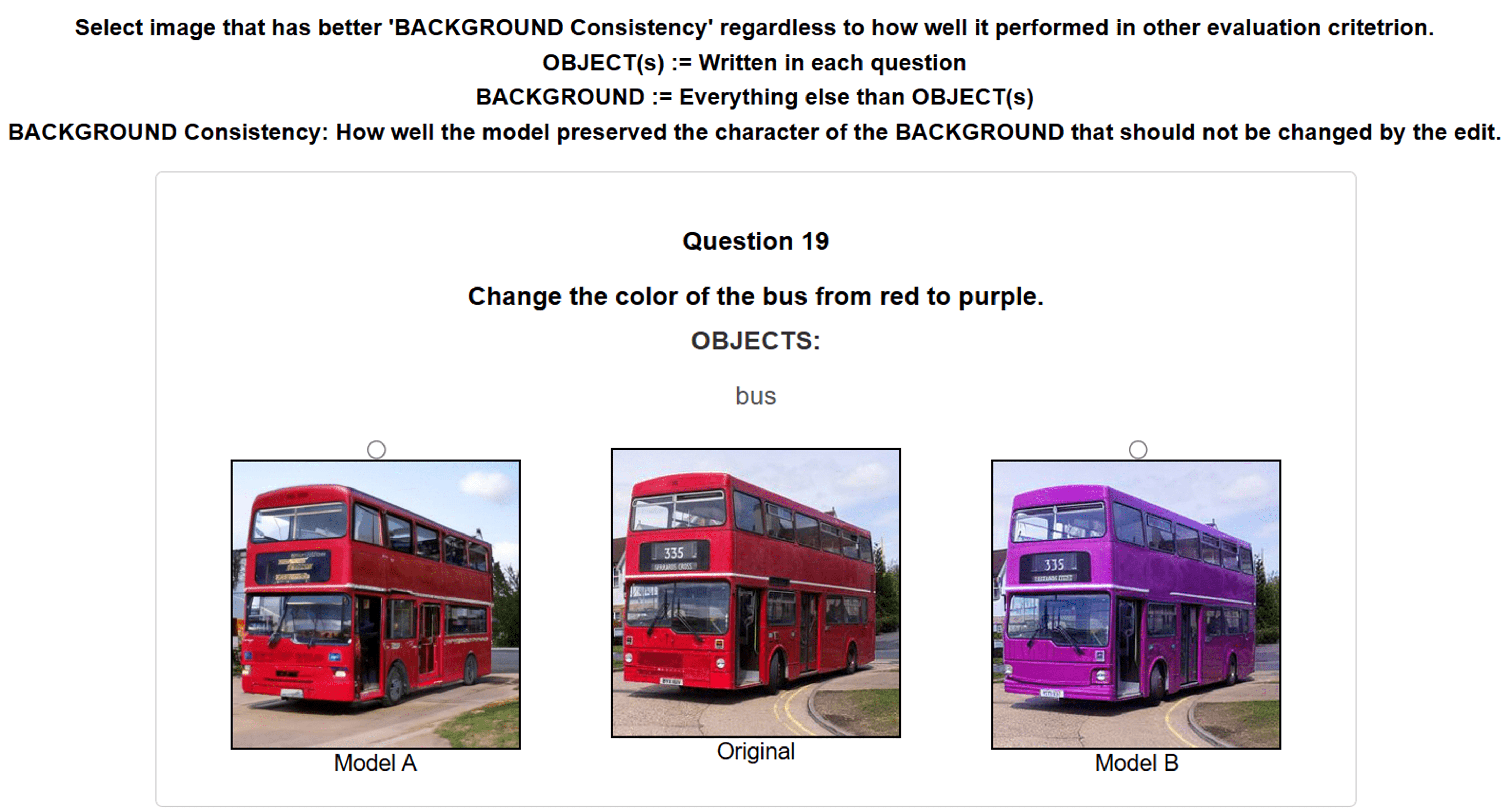}
  \caption{Background Consistency}
  \label{fig:user_study_ex_bc}
\end{subfigure}

\caption{\textbf{User study question examples.}
Each question is binary, asking to choose the better edited result for each score criterion. The target object in each image is also given for the user to help evaluate.
(a) is one of overall edit quality (corresponding to the \textsc{Total Score}) assessment question, selecting the better edited image considering all evaluation criteria.
(b) is a question for Object Fidelity (OF), which is focused on selecting an image that has better quality of the editable object.
(c) is for measuring Background Fidelity (BF), selecting an image with better quality of background along instruction.
(d) is for Object Consistency (OC), selecting an image that preserves the original object better.
(e) is for Background Consistency (BC), which is focused on selecting image that preserves the original background better.}
\label{fig:user_study_ex}
\end{figure}

\subsection{Examples of User Study Questions}
\label{sec:user_study_ex}

Sample guidelines and questions of our user study are provided in \cref{fig:user_study_ex}.
Participants were given with a brief instruction of the questions and evaluation criteria that they were supposed to take into account.
Each question informs the participants what kind of edit has been performed on which object.
Given all these information, participants were asked to choose a superior result out of two options provided with an original image.

\begin{figure}[t]
  \centering
  \includegraphics[width=\linewidth]{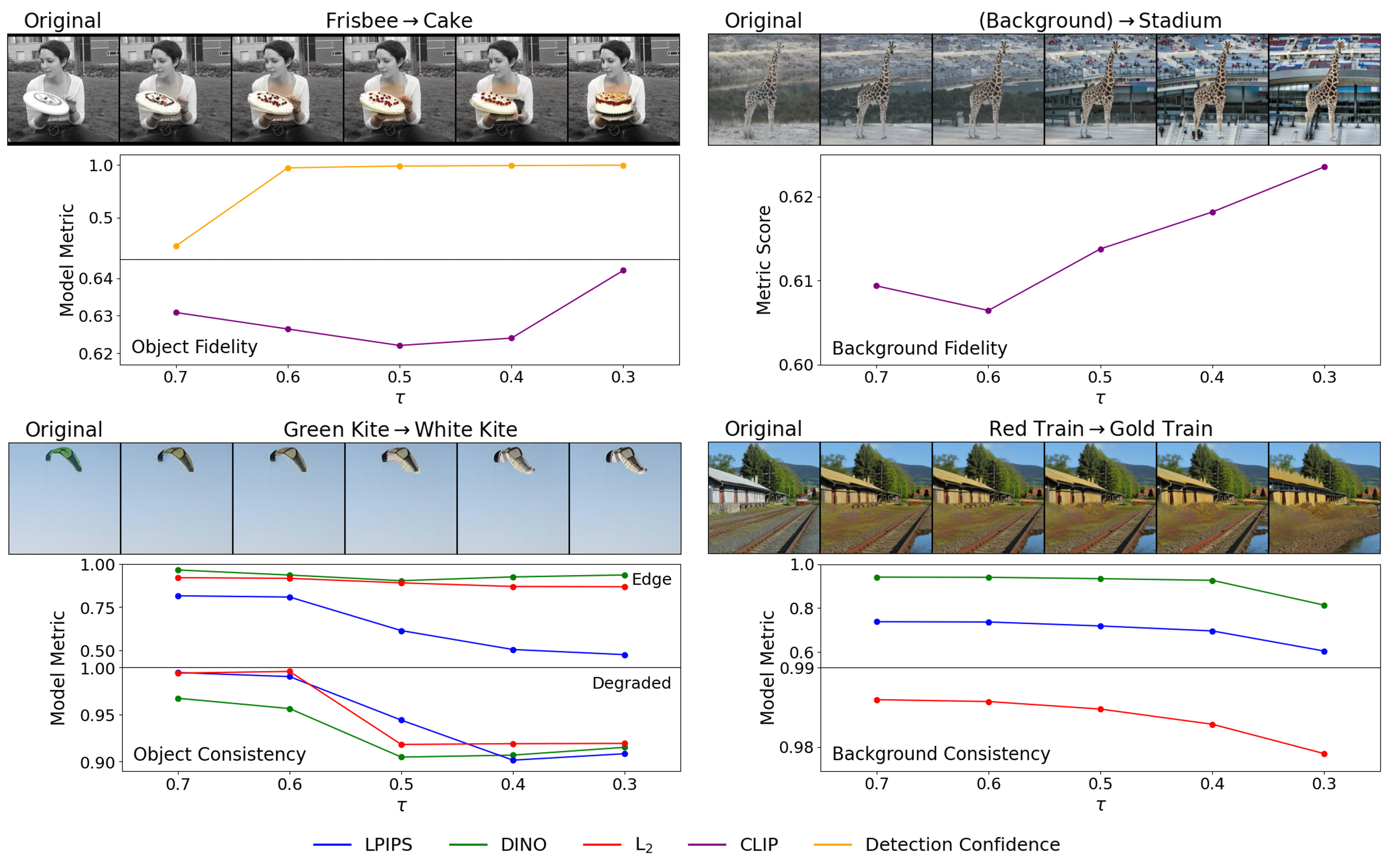}  
  ~
  \vspace{-7mm}
  \caption{\textbf{Evaluations by each metric for edited images with different edit strength.}
  The editing has been done using the P2P model with various hyper-parameter $\tau$, where a higher $\tau$ means weaker editing, and the paired graph illustrates the specific metric scores of our method.
  In each example, the left-most image is the original image and the rest shows the edited images for the query indicated above, using different $\tau$ values indicated below.
  On the top-left, the Object Fidelity (OF) scores are plotted, and the top-right quadrant shows the Background Fidelity (BF) metrics.
  At the bottom-left, the Object Consistency (OC) scores measured for a \textsc{Object Attribute Change} task is shown. The two figures are computed from detected Canny-edge of the objects and degraded object images, respectively.
  Lastly, the bottom-right graph shows Background Consistencies (BC) measured with different metric models.}
  \vspace{-3mm}
  \label{fig:hyperparam}
\end{figure}

\section{Additional Experimental Results}
\label{sec:add_exp}

\subsection{Suitability of Metric Models}
\label{sec:add_exp:suit_metric}

\cref{fig:hyperparam} shows samples of edited images by P2P across a range of $\tau$, along with the trend of our evaluation metric scores.
There are differences in the sensitivity and amplitude of changes, but overall, we observe that all metrics agree with our design intention. As edits are getting more faithful, all the metric scores involved in fidelity are increasing. 
Also, as the edit result gradually fails to retain the original character, the two consistency scores in the second row clearly drop.
These results show that our metrics capture the characteristics and quality of the edited images well, and therefore, are suitable for editing benchmark.

\subsection{Reliability of Instance Segmentation Model}

\begin{figure}[t]
  \centering
  \vspace{-0.2cm}
  \includegraphics[width=0.6\linewidth]{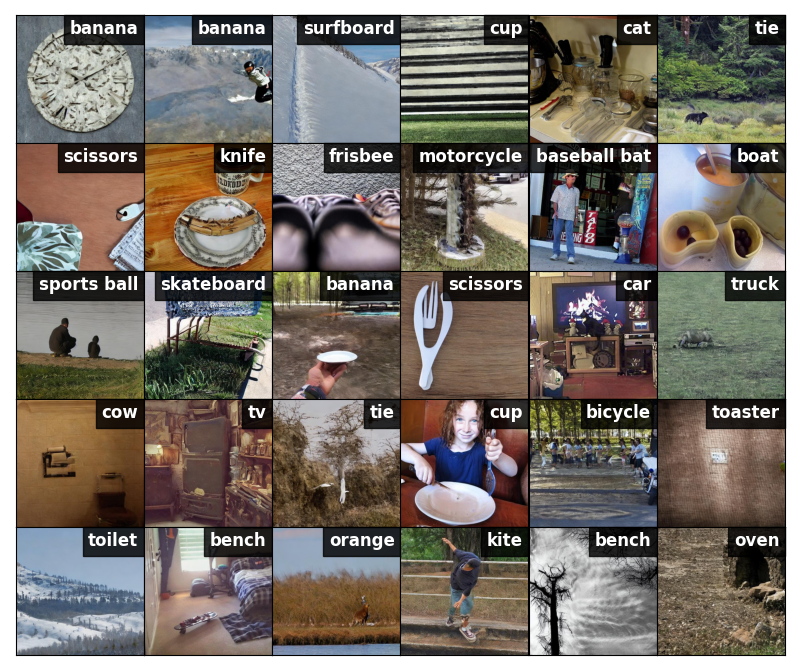}
  \vspace{-0.2cm}
  \caption{\textbf{Illustration of segmentation failure cases.} Images of instance segmentation failures are plotted together with the object names the detector tried to locate.}
  \vspace{-0.3cm}
  \label{fig:segnotfound}
\end{figure}

The credibility of our evaluation method is strongly dependent on that of the instance segmentation model we use.
To avoid detection failures, during the evaluation process, we gather all the detection results, even those with extremely low confidence.
However, this does not guarantee a perfect detection.
The overall failure ratio of the instance segmentation model, where it could not find the desired object in the image, was 16.01\%.
Also, in many cases, these failures were not because of the limitation of the segmentation model. We demonstrate 30 randomly sampled images in \cref{fig:segnotfound} with the object names that the detector had tried to locate.
From this figure, we see that the detection failure is usually due to the absence of desired object, not because of the model's limitation. 
Thus, we claim that the instance segmentation used in our experiment is sufficiently credible to conduct the evaluation.

\begin{figure}
    \centering
    \includegraphics[width=0.6\linewidth]{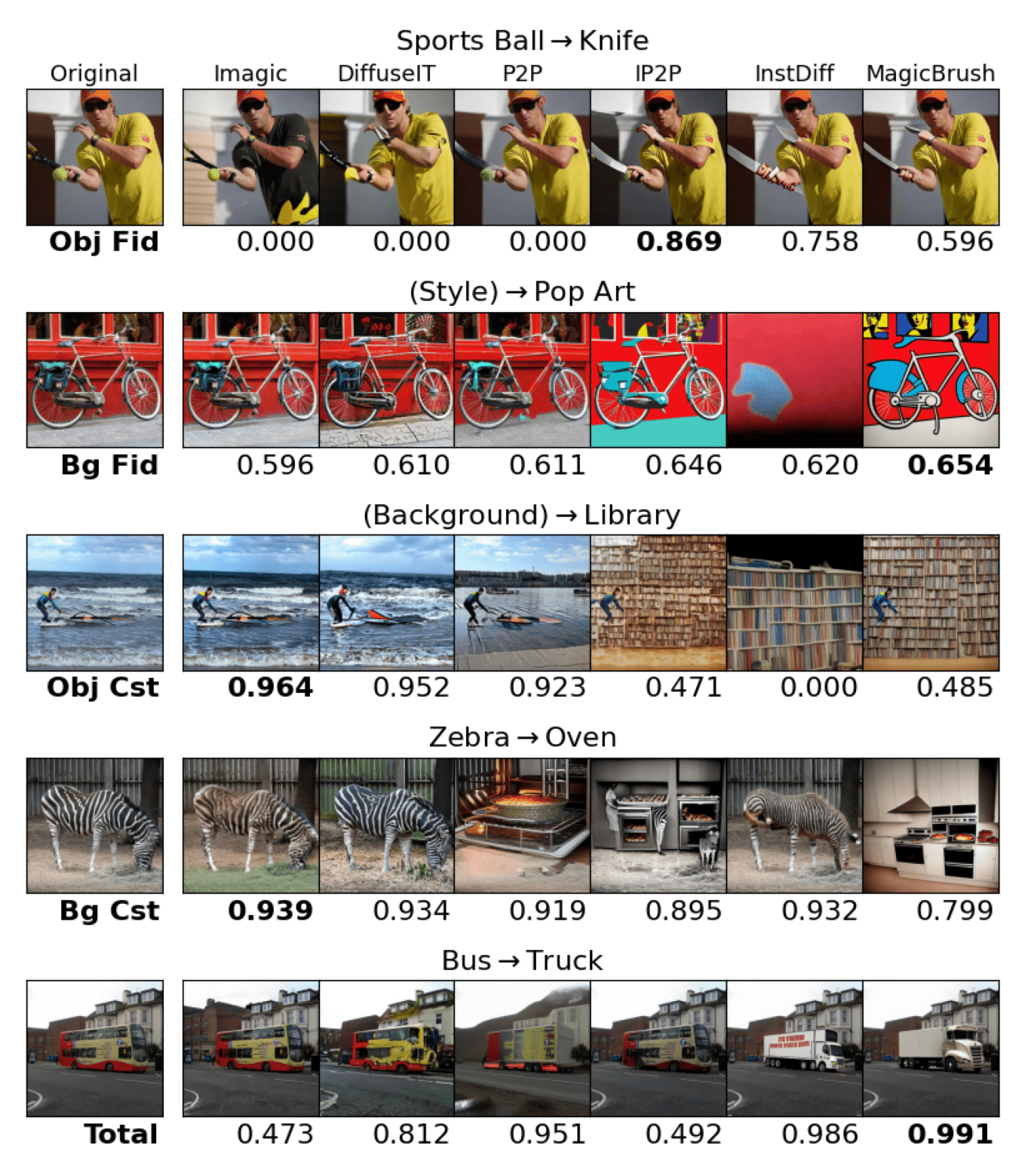}
    \caption{\textbf{Examples of our evaluation results.} The left-most image is the original input image. The next three shows the output image and evaluated metrics for select description-based models, while the last three are those for instruction-based models, according to the query indicated above each example.}
    \label{fig:modelcomp_qual}
\end{figure}

\begin{figure}
    \centering
    \includegraphics[width=0.6\linewidth]{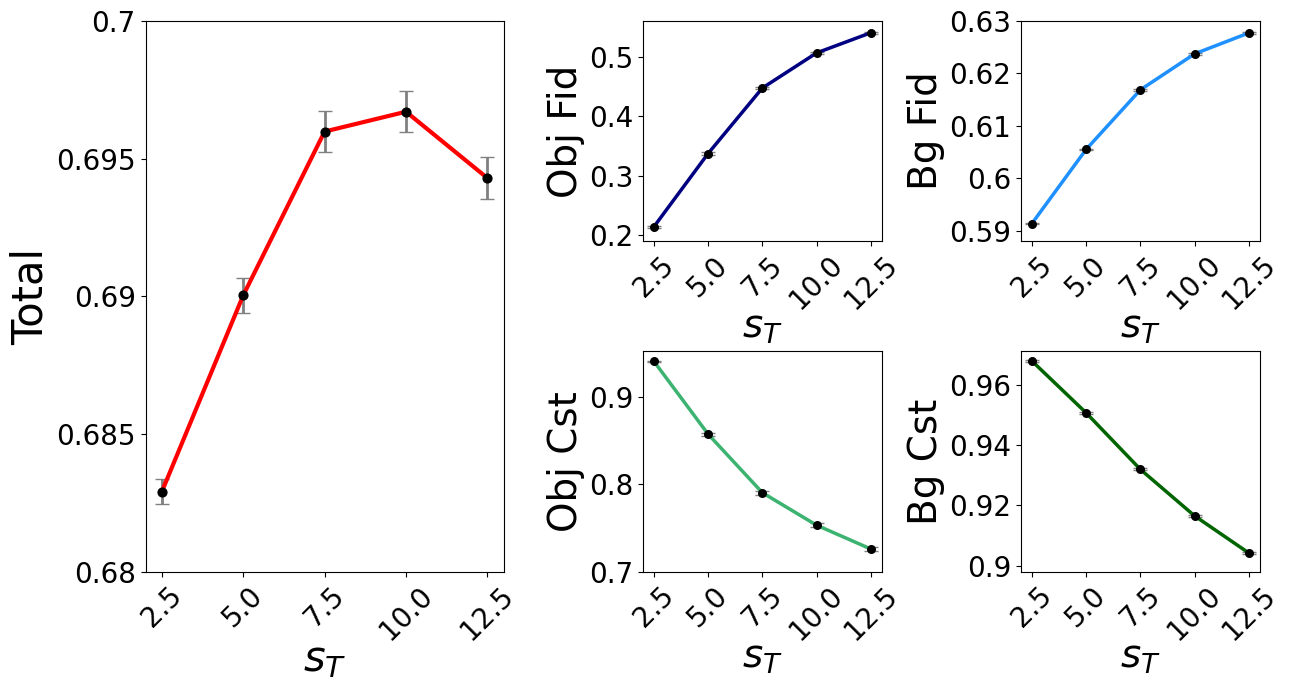}
    \caption{\textbf{Benchmark scores with varied editing intensity for Instruct-Pix2Pix.} Each figure plots each metric and the Total Score for a range of hyper-parameter values.}
    \label{fig:param_comp_ip2p}
\end{figure}

\begin{figure*}
  \centering
  \vspace{7mm}
  \begin{subfigure}{0.49\textwidth}
    \centering
    \includegraphics[width=\linewidth]{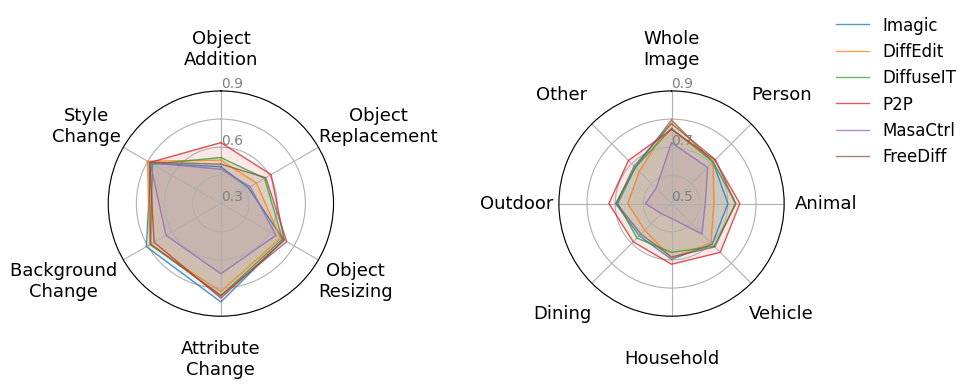}
    \caption{Description-based Models}
    \label{fig:polygon_caption}
  \end{subfigure}
  \hspace{0.5em} 
  \begin{subfigure}{0.49\textwidth}
    \centering
    \includegraphics[width=\linewidth]{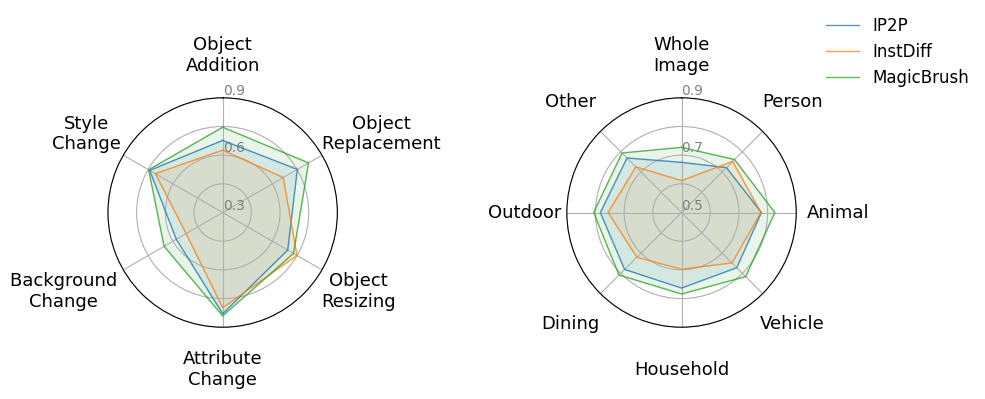}
    \caption{Instruction-based Models}
    \label{fig:polygon_instruction}
  \end{subfigure}

  \caption{\textbf{Evaluation scores for each edit type and edit target object class for each model.} (a) plots for the six description-based models and (b) plots for the three instruction-based models.}
  \label{fig:polygon_results}
  \vspace{15mm}
\end{figure*}

\begin{table*}[ht]
\centering
\resizebox{\textwidth}{!}{
{\footnotesize
\begin{tabular}{lll p{2cm}<{\centering} p{2cm}<{\centering} p{2cm}<{\centering} p{2cm}<{\centering} p{2cm}<{\centering}  p{2cm}<{\centering}}
\toprule[1.5pt]
\multicolumn{3}{c}{\textbf{Models}} & \textbf{\parbox{2cm}{\centering Object\\Fidelity}} & \textbf{\parbox{2cm}{\centering Background\\Fidelity}} & \textbf{\parbox{2cm}{\centering Object\\Consistency}} & \textbf{\parbox{2cm}{\centering Background\\Consistency}} & \textbf{\parbox{2cm}{\centering Image\\Quality}} & \textbf{\parbox{2cm}{\centering Total\\Score}} \\ \midrule
& \multicolumn{2}{c}{DiffEdit} & 0.2277 $\pm$ 0.0018 & 0.5910 $\pm$ 0.0001 & 0.8338 $\pm$ 0.0018 & \textbf{0.9608 $\pm$ 0.0002} & 0.7477 $\pm$ 0.0024 & 0.6552 $\pm$ 0.0006
 \\ \cmidrule(lr){2-3} \cmidrule(lr){4-8} \cmidrule(lr){9-9}
& \multicolumn{2}{c}{DiffuseIT} & 0.3202 $\pm$ 0.0019 & 0.6045 $\pm$ 0.0001 & 0.8616 $\pm$ 0.0012 & 0.8958 $\pm$ 0.0002 & 0.5569 $\pm$ 0.0021 & 0.6682 $\pm$ 0.0006
 \\ \cmidrule(lr){2-3} \cmidrule(lr){4-8} \cmidrule(lr){9-9}
& & $s_T = 2.5$ & 0.2571 $\pm$ 0.0018 & 0.5903 $\pm$ 0.0001 & 0.8980 $\pm$ 0.0011 & 0.9413 $\pm$ 0.0002 & 0.8360 $\pm$ 0.0014 & 0.6749 $\pm$ 0.0005
 \\ 
& & $s_T = 5.0$ & 0.2954 $\pm$ 0.0020 & 0.5939 $\pm$ 0.0001 & 0.8766 $\pm$ 0.0012 & 0.9253 $\pm$ 0.0002 & 0.7770 $\pm$ 0.0016 & 0.6749 $\pm$ 0.0006
 \\ 
& FreeDiff & $s_T = 7.5$ & 0.3174 $\pm$ 0.0020 & 0.5964 $\pm$ 0.0001 & 0.8631 $\pm$ 0.0013 & 0.9150 $\pm$ 0.0002 & 0.7180 $\pm$ 0.0014 & 0.6739 $\pm$ 0.0006
 \\
& & $s_T = 10.0$ & 0.3323 $\pm$ 0.0020 & 0.5982 $\pm$ 0.0001 & 0.8545 $\pm$ 0.0013 & 0.9074 $\pm$ 0.0002 & 0.6651 $\pm$ 0.0020 & 0.6729 $\pm$ 0.0006
 \\ 
& & $s_T = 12.5$ & 0.3405 $\pm$ 0.0021 & 0.5995 $\pm$ 0.0001 & 0.8487 $\pm$ 0.0013 & 0.9014 $\pm$ 0.0002 & 0.6185 $\pm$ 0.0028 & 0.6714 $\pm$ 0.0006
 \\ \cmidrule(lr){2-3} \cmidrule(lr){4-8} \cmidrule(lr){9-9}
& & $\tau = 0.3$ & \textbf{0.3595 $\pm$ 0.0020} & \textbf{0.6073 $\pm$ 0.0002} & 0.8528 $\pm$ 0.0015 & 0.9261 $\pm$ 0.0003 & 0.6551 $\pm$ 0.0023 & 0.6858 $\pm$ 0.0006
 \\ 
& & $\tau = 0.4$ & 0.3334 $\pm$ 0.0019 & 0.6038 $\pm$ 0.0002 & 0.8705 $\pm$ 0.0014 & 0.9339 $\pm$ 0.0003 & 0.7097 $\pm$ 0.0031 & \textbf{0.6859 $\pm$ 0.0006}
 \\ 
& P2P & $\tau = 0.5$ & 0.3063 $\pm$ 0.0019 & 0.6002 $\pm$ 0.0001 & 0.8811 $\pm$ 0.0013 & 0.9397 $\pm$ 0.0003 & 0.7551 $\pm$ 0.0029 & 0.6833 $\pm$ 0.0006
 \\ 
& & $\tau = 0.6$ & 0.2823 $\pm$ 0.0018 & 0.5971 $\pm$ 0.0001 & 0.8864 $\pm$ 0.0013 & 0.9433 $\pm$ 0.0003 & 0.7830 $\pm$ 0.0027 & 0.6794 $\pm$ 0.0006
 \\ 
& & $\tau = 0.7$ & 0.2630 $\pm$ 0.0018 & 0.5946 $\pm$ 0.0001 & 0.8898 $\pm$ 0.0013 & 0.9456 $\pm$ 0.0003 & 0.7993 $\pm$ 0.0023 & 0.6758 $\pm$ 0.0005
 \\ \cmidrule(lr){2-3} \cmidrule(lr){4-8} \cmidrule(lr){9-9}
& & $\eta = 0.2$ & 0.2138 $\pm$ 0.0016 & 0.5866 $\pm$ 0.0001 & \textbf{0.9380 $\pm$ 0.0008} & 0.9421 $\pm$ 0.0003 & 0.8499 $\pm$ 0.0018 & 0.6737 $\pm$ 0.0004
 \\ 
& & $\eta = 0.4$ & 0.2179 $\pm$ 0.0017 & 0.5869 $\pm$ 0.0001 & 0.9311 $\pm$ 0.0009 & 0.9339 $\pm$ 0.0003 & \textbf{0.8519 $\pm$ 0.0019} & 0.6711 $\pm$ 0.0005
 \\ 
& Imagic & $\eta = 0.6$ & 0.2271 $\pm$ 0.0017 & 0.5878 $\pm$ 0.0001 & 0.9237 $\pm$ 0.0009 & 0.9201 $\pm$ 0.0003 & 0.8406 $\pm$ 0.0014 & 0.6682 $\pm$ 0.0005
 \\ 
& & $\eta = 0.8$ & 0.2534 $\pm$ 0.0018 & 0.5902 $\pm$ 0.0001 & 0.9134 $\pm$ 0.0010 & 0.9038 $\pm$ 0.0004 & 0.8184 $\pm$ 0.0020 & 0.6682 $\pm$ 0.0005
 \\ 
& & $\eta = 1.0$ & 0.2927 $\pm$ 0.0019 & 0.5970 $\pm$ 0.0001 & 0.8942 $\pm$ 0.0012 & 0.8841 $\pm$ 0.0003 & 0.7678 $\pm$ 0.0034 & 0.6690 $\pm$ 0.0006
 \\ \cmidrule(lr){2-3} \cmidrule(lr){4-8} \cmidrule(lr){9-9}
& & $s_T = 2.5$ & 0.1833 $\pm$ 0.0017 & 0.5932 $\pm$ 0.0001 & 0.6312 $\pm$ 0.0026 & 0.9042 $\pm$ 0.0003 & 0.2580 $\pm$ 0.0032 & 0.5716 $\pm$ 0.0008
 \\ 
& & $s_T = 5.0$ & 0.2111 $\pm$ 0.0018 & 0.5949 $\pm$ 0.0001 & 0.6586 $\pm$ 0.0024 & 0.9003 $\pm$ 0.0003 & 0.2598 $\pm$ 0.0030 & 0.5846 $\pm$ 0.0007
 \\ 
& MasaCtrl & $s_T = 7.5$ & 0.2332 $\pm$ 0.0019 & 0.5967 $\pm$ 0.0001 & 0.6754 $\pm$ 0.0023 & 0.8954 $\pm$ 0.0003 & 0.2685 $\pm$ 0.0022 & 0.5936 $\pm$ 0.0007
 \\ 
& & $s_T = 10.0$ & 0.2496 $\pm$ 0.0019 & 0.5983 $\pm$ 0.0001 & 0.6868 $\pm$ 0.0023 & 0.8908 $\pm$ 0.0003 & 0.2834 $\pm$ 0.0029 & 0.5999 $\pm$ 0.0007
 \\ 
\multirow{-25}{*}{\parbox{2cm}{\centering Description\\-based}} & & $s_T = 12.5$ & 0.2632 $\pm$ 0.0020 & 0.5995 $\pm$ 0.0001 & 0.6982 $\pm$ 0.0022 & 0.8867 $\pm$ 0.0002 & 0.2968 $\pm$ 0.0025 & 0.6056 $\pm$ 0.0007
 \\ \midrule\midrule
& \multicolumn{2}{c}{MagicBrush} & 0.5378 $\pm$ 0.0020 & 0.6196 $\pm$ 0.0002 & 0.8259 $\pm$ 0.0018 & 0.9513 $\pm$ 0.0003 & 0.6977 $\pm$ 0.0032 & \textbf{0.7329 $\pm$ 0.0007}
 \\ \cmidrule(lr){2-3} \cmidrule(lr){4-8} \cmidrule(lr){9-9}
& \multicolumn{2}{c}{InstDiff} & 0.4596 $\pm$ 0.0022 & 0.6205 $\pm$ 0.0001 & 0.6870 $\pm$ 0.0022 & 0.9090 $\pm$ 0.0005 & 0.4148 $\pm$ 0.0039 & 0.6639 $\pm$ 0.0008
 \\ \cmidrule(lr){2-3} \cmidrule(lr){4-8} \cmidrule(lr){9-9}
& & $s_T = 2.5$ & 0.2141 $\pm$ 0.0016 & 0.5914 $\pm$ 0.0001 & \textbf{0.9407 $\pm$ 0.0008} & \textbf{0.9678 $\pm$ 0.0002} & \textbf{0.8983 $\pm$ 0.0014} & 0.6829 $\pm$ 0.0005
 \\ 
& & $s_T = 5.0$ & 0.3373 $\pm$ 0.0019 & 0.6055 $\pm$ 0.0002 & 0.8573 $\pm$ 0.0017 & 0.9506 $\pm$ 0.0003 & 0.8049 $\pm$ 0.0024 & 0.6900 $\pm$ 0.0006
 \\ 
& & $s_T = 7.5$ & 0.4474 $\pm$ 0.0021 & 0.6169 $\pm$ 0.0002 & 0.7903 $\pm$ 0.0021 & 0.9319 $\pm$ 0.0003 & 0.6658 $\pm$ 0.0031 & 0.6960 $\pm$ 0.0007
 \\ 
& & $s_T = 10.0$ & 0.5064 $\pm$ 0.0020 & 0.6237 $\pm$ 0.0001 & 0.7531 $\pm$ 0.0023 & 0.9164 $\pm$ 0.0004 & 0.5394 $\pm$ 0.0035 & 0.6967 $\pm$ 0.0008
 \\ 
& & $s_T = 12.5$ & 0.5403 $\pm$ 0.0020 & 0.6277 $\pm$ 0.0001 & 0.7258 $\pm$ 0.0023 & 0.9041 $\pm$ 0.0004 & 0.4393 $\pm$ 0.0034 & 0.6943 $\pm$ 0.0008
 \\ \cmidrule(lr){3-3} \cmidrule(lr){4-8} \cmidrule(lr){9-9}
& & $s_I = 1.0$ & \textbf{0.6002 $\pm$ 0.0020} & \textbf{0.6342 $\pm$ 0.0001} & 0.6640 $\pm$ 0.0025 & 0.8762 $\pm$ 0.0004 & 0.2839 $\pm$ 0.0026 & 0.6855 $\pm$ 0.0008
 \\ 
& & $s_I = 1.25$ & 0.5385 $\pm$ 0.0020 & 0.6273 $\pm$ 0.0001 & 0.7213 $\pm$ 0.0023 & 0.9092 $\pm$ 0.0004 & 0.4858 $\pm$ 0.0029 & 0.6948 $\pm$ 0.0008 
 \\ 
& & $s_I = 1.5$ & 0.4474 $\pm$ 0.0021 & 0.6169 $\pm$ 0.0002 & 0.7903 $\pm$ 0.0021 & 0.9319 $\pm$ 0.0003 & 0.6658 $\pm$ 0.0031 & 0.6960 $\pm$ 0.0007
 \\ 
& & $s_I = 1.75$ & 0.3547 $\pm$ 0.0020 & 0.6060 $\pm$ 0.0002 & 0.8521 $\pm$ 0.0018 & 0.9467 $\pm$ 0.0003 & 0.7831 $\pm$ 0.0032 & 0.6918 $\pm$ 0.0007 
 \\ 
\multirow{-13}{*}{\parbox{2cm}{\centering Instruction\\-based}} & \multirow{-11}{*}{IP2P} & $s_I = 2.0$ & 0.2862 $\pm$ 0.0018 & 0.5983 $\pm$ 0.0001 & 0.8943 $\pm$ 0.0014 & 0.9547 $\pm$ 0.0003 & 0.8367 $\pm$ 0.0021 & 0.6865 $\pm$ 0.0006
 \\ \bottomrule[1.5pt]
\end{tabular}
}
}
\caption{\textbf{\ours\ results across all score criteria.} Our \ours\ evaluation results are shown for the six description-based models and three instruction-based models, across five score criteria. We also show the differences between various hyper-parameter values in each model.}
\vspace{5mm}
\label{tab:res}
\end{table*}

\subsection{Detailed Results}
\label{sec:app:add:detail_res}

\cref{fig:modelcomp_qual} plots a similar figure to \cref{fig:ip2pcomp} but across various models, where we see that our \ours\ gives accurate assessments aligned with real images.
Both fidelity scores are low for failure cases, while they get higher on successful ones.
Consistency scores for those that excellently retains the original details, that partially retains the original details, and that completely loses the original details are clearly distinguished throughout the model scores.
Total score shows a balance between these criteria, where the highest score is given to the image that not only changes the object correctly, but also keeps the background perfectly.

\cref{fig:param_comp_ip2p} plots the same graph with \cref{fig:param_comp} but for result of one of the instruction-based model, InstructPix2Pix. We can also observe identical trends found in \cref{fig:param_comp}. \cref{fig:polygon_results} plots the same data listed in \cref{tab:per_task_scores} and \ref{tab:per_class_scores}. Here we can visually see strong points and weak points of each model together with inter-model comparison. 
\cref{tab:res} gives full list of every model benchmark scores on every tested parameter settings for all 5 subscores and total scores.

\subsection{Experiment on human perception alignment with additional dataset}
\label{sec:app:add:imagenet}

\begin{figure}
    \centering
    \includegraphics[width=0.4\linewidth]{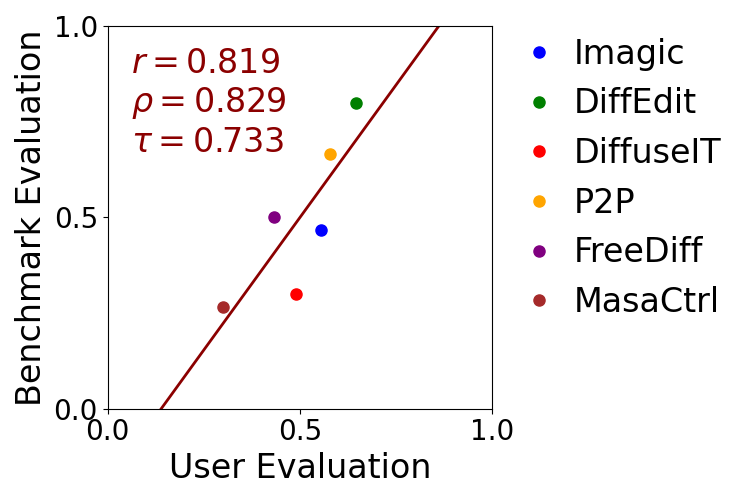}
    \caption{Alignment between human preference and \ours\ on unseen new dataset.}
    \label{fig:reb_imagenet_alignment}
\end{figure}

The alignment results of the unseen dataset in Figure \ref{fig:reb_imagenet_alignment} is presented in the same format as Figure \ref{fig:user_alignment_test} in the main paper. We tested our metric and human preference alignment on out-of-domain dataset, ImageNet\cite{deng2009imagenet}. This result demonstrates that \ours\ maintains high correlation with human evaluations even on a unseen dataset. This suggests that \ours\ can be extended to various types of data and applied to a wide range of editing tasks in the future.

\section{Limitations}
\label{sec:limit}

\begin{figure}
\centering
\begin{subfigure}{0.51\linewidth}
    \centering
    \includegraphics[width=\textwidth]{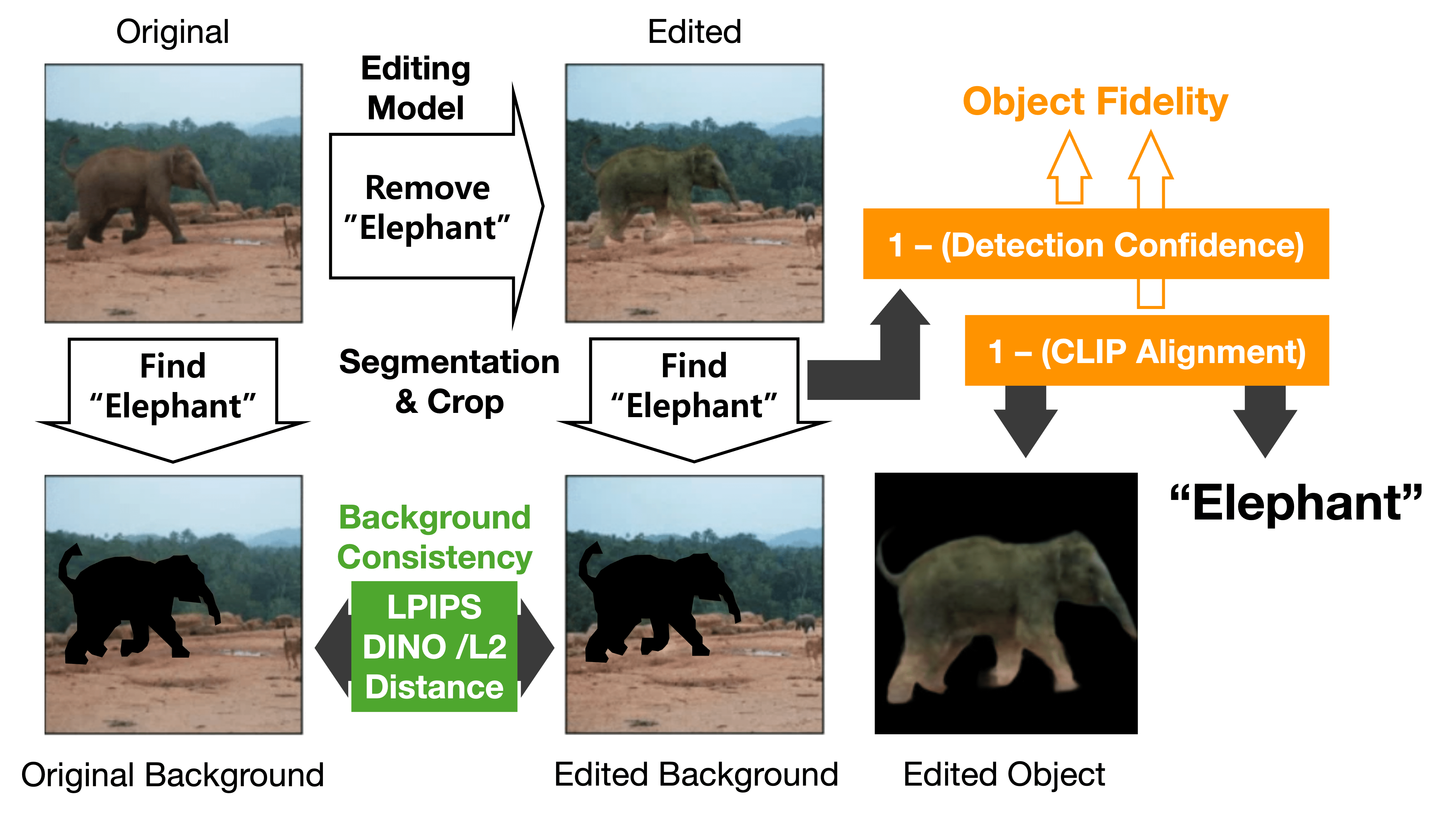}
    \caption{Target object found}
    \label{fig:eval_rm:found}
\end{subfigure}
\hfill
\begin{subfigure}{0.46\linewidth}
    \centering
    \includegraphics[width=\textwidth]{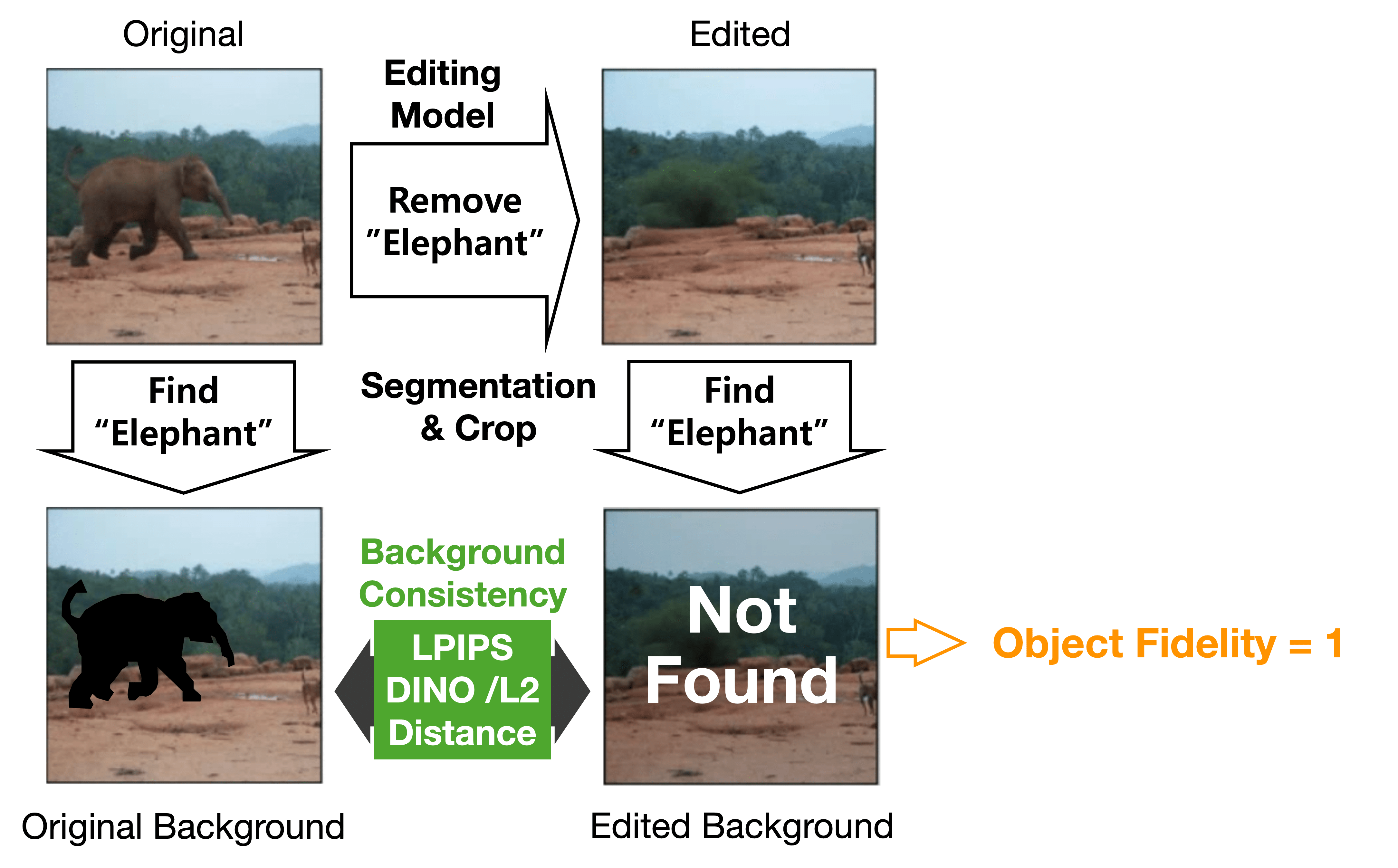}
    \caption{Target object not found}
    \label{fig:eval_rm:notfound}
\end{subfigure}
\caption{\textbf{Evaluation workflow of \textsc{Object Removal} query result.} (a) shows the case when target object is found in edited image, and (b) shows the case when target object is not found in edited image.}
\label{fig:eval_rm}
\end{figure}

Our benchmark aims to encompass every feasible editing task, but some tasks have been inevitably excluded.
First, \textsc{Object Removal} has been limited only for the instruction-based models, due to the difficulty generating captions for description-based models, as explained in \cref{sec:pre}.
Specifically, 8,315 \textsc{Object Removal} queries are carried out with the three instruction-based models.
The evaluation workflow of \textsc{Object Removal} task for the instruction-based models is shown in \cref{fig:eval_rm}, and the benchmark results are listed in \cref{tab:rm_scores}.

Second, moving or rotating an object has not been considered.
In order to clearly define where or which direction to move or rotate an object, not only the target object to move or rotate but also another object to become a reference point of a new location or direction are needed.
However, our base dataset contain few images containing two or more editable objects, making it challenging to create a large number of movement or rotation queries.
We leave it as a future work which should be addressed with an additional base dataset.

Lastly, for the \textsc{Object Attribute Change} queries, GQA annotations have many other attribute words beyond our four classes (color, state, material, action), such as attributes about emotion, hair, or fashion.
We decide to exclude these attributes mainly because of the ambiguity to localize the exact area to be affected by changing these attributes, resulting in difficulty to measure fidelity and consistency.
To overcome this, we may divide the human figure into more manageable regions (such as the face, hair, body, arms, and legs) using human pose estimation.
This strategy may widen the scope of our benchmark, but we leave it as a potential future work.

\begin{table}[]
    \centering
    \begin{tabular}{ll p{3cm}<{\centering}}
    \toprule[1.5pt]
    \multicolumn{2}{c}{\textbf{Models}} & \textbf{Object Removal} \\ 
    \midrule
    \multicolumn{2}{c}{MagicBrush} & 0.7230 $\pm$ 0.0025 \\ 
    \cmidrule(lr){1-2} \cmidrule(lr){3-3} 
    \multicolumn{2}{c}{InstructDiffusion} & \textbf{0.9082 $\pm$ 0.0016} \\ 
    \cmidrule(lr){1-2} \cmidrule(lr){3-3} 
    & $s_T = 2.5$ & 0.5272 $\pm$ 0.0010 \\
    & $s_T = 5.0$ & 0.5834 $\pm$ 0.0019 \\
    & $s_T = 7.5$ & 0.6485 $\pm$ 0.0024 \\
    & $s_T = 10.0$ & 0.6817 $\pm$ 0.0025 \\
    & $s_T = 12.5$ & 0.6925 $\pm$ 0.0025 \\
    \cmidrule(lr){2-2} \cmidrule(lr){3-3} 
    & $s_I = 1.0$ & 0.6972 $\pm$ 0.0025 \\
    & $s_I = 1.25$ & 0.7159 $\pm$ 0.0026 \\
    & $s_I = 1.5$ & 0.6485 $\pm$ 0.0024 \\
    & $s_I = 1.75$ & 0.5694 $\pm$ 0.0018 \\
    \multirow{-11}{*}{Instruct-Pix2Pix} & $s_I = 2.0$ & 0.5369 $\pm$ 0.0013 \\
    \bottomrule[1.5pt]
    \end{tabular}
    \caption{\textbf{Result of \textsc{Object Removal} task} in instruction-based models on \ours\ benchmark.}
    \label{tab:rm_scores}
\end{table}

\end{document}